%% file: main.tex
\documentclass[journal]{IEEEtran}
\usepackage{amsmath,amsopn,amssymb}
\usepackage{graphicx,xspace,color,soul}
\usepackage{epsfig,subfigure}
\usepackage{longtable,multirow}
\usepackage{array,float}
\usepackage{cite}
\usepackage{algorithm,algorithmicx,algpseudocode}
\usepackage{bm,epstopdf}

\renewcommand{\vec}[1]{\mathbf{#1}}

\providecommand{\norm}[1]{\lVert#1\rVert}

\begin{document}

\title{Local Frequency Interpretation and Non-Local Self-Similarity on Graph for Point Cloud Inpainting}

\author{
	Zeqing~Fu,~\IEEEmembership{Student Member,~IEEE,}
	Wei~Hu{\small $~^{\ast}$},~\IEEEmembership{Member,~IEEE,}
	and~Zongming~Guo,~\IEEEmembership{Member,~IEEE}
\thanks{Z. Fu (e-mail: zeqing\_fu@pku.edu.cn), W. Hu (e-mail: forhuwei@pku. edu.cn), Z. Guo (e-mail: guozongming@pku.edu.cn) is with Institute of Computer Science \& Technology, Peking University, No. 128 Zhongguancun North Street, Beijing, China.}
}

\markboth{Submission of IEEE Transactions on Image Processing}
{Shell \MakeLowercase{\textit{et al.}}: Bare Demo of IEEEtran.cls for IEEE Journals}

\maketitle
\begin{abstract}
As 3D scanning devices and depth sensors mature, point clouds have attracted increasing attention as a format for 3D object representation, with applications in various fields such as tele-presence, navigation and heritage reconstruction. However, point clouds usually exhibit holes of missing data, mainly due to the limitation of acquisition techniques and complicated structure. Further, point clouds are defined on irregular non-Euclidean domains, which is challenging to address especially with conventional signal processing tools. Hence, leveraging on recent advances in graph signal processing, we propose an efficient point cloud inpainting method, exploiting both the local smoothness and the non-local self-similarity in point clouds. Specifically, we first propose a frequency interpretation in graph nodal domain, based on which we introduce the local graph-signal smoothness prior in order to describe the local smoothness of point clouds. Secondly, we explore the characteristics of non-local self-similarity, by globally searching for the most similar area to the missing region. The similarity metric between two areas is defined based on the direct component and the anisotropic graph total variation of normals in each area. Finally, we formulate the hole-filling step as an optimization problem based on the selected most similar area and regularized by the graph-signal smoothness prior. Besides, we propose voxelization and automatic hole detection methods for the point cloud prior to inpainting. Experimental results show that the proposed approach outperforms four competing methods significantly, both in objective and subjective quality.
\end{abstract}

\begin{IEEEkeywords}
Graph signal processing, frequency interpretation, non-local self-similarity, point cloud inpainting.
\end{IEEEkeywords}

\IEEEpeerreviewmaketitle

\section{Introduction}
\label{sec:intro}
\input{intro}

\section{Related Work}
\label{sec:related}
\input{related}

\section{Frequency Interpretation in Graph Nodal Domain}
\label{sec:graph}
\input{graph}

\section{Overview of the Proposed Inpainting Framework}
\label{sec:overview}
\input{overview}

\section{Voxelization}
\label{sec:voxelize}
\input{voxelize}

\section{Automatic Hole Detection}
\label{sec:detect}
\input{detect}

\section{The Proposed Inpainting Method}
\label{sec:method}
\input{method}

\section{Experiments}
\label{sec:results}
\input{results}

\section{Conclusion}
\label{sec:conclude}
\input{conclude}

\ifCLASSOPTIONcaptionsoff
  \newpage
\fi

\bibliographystyle{IEEEtran}
\bibliography{ref}



\end{document}

%% file: intro.tex
\IEEEPARstart{G}{raphs} are generic data representation forms for the description of pairwise relations in the data, which consist of vertices and connecting edges.  The weight associated with each edge in the graph often represents the similarity between the two vertices it connects. Nowadays, graphs have been used to model various types of relations in physical, biological, social and information systems because of advantages in consuming \textit{irregular} discrete signals. The connectivities and edge weights are either dictated by the physics (e.g. physical distance between nodes ) or inferred from the data. The data residing on these graphs is a finite collection of samples, with one sample at each vertex in the graph, which is referred to as graph signal~\cite{Shuman13}. Compared with conventional signals (e.g., images and videos) defined on regular grids, graph signals are often defined on non-Euclidean domains where the neighborhood is unclear and the order might be unimportant. Statistical properties such as stationarity and compositionality are even not well defined for graph signals. Such kinds of data arise in diverse engineering and science fields, such as logistics data in transportation networks, functionalities in neural networks, temperatures on sensor networks, pixel intensities in images, etc.

For any undirected graph, there is an orthogonal basis corresponding to the eigenvectors of a graph matrix\footnote{In spectral graph theory~\cite{Chung96}, a graph matrix is a matrix describing the connectivities in the graph, such as the commonly used graph Laplacian matrix, which will be introduced in Section~\ref{sec:graph}.}. Thus once a graph has been defined, a definition of frequency is readily available, which allows us to extend conventional signal processing to graph signals. Multiple choices are possible, such as the cut size of the constructed orthogonal basis set~\cite{Sardellitti17} and the total variation of the associated spectral component~\cite{Ortega17}. These methods are able to accurately reflect the relationship among the frequencies of different eigenvectors in certain kinds of graph, but they are inapplicable to other graphs. Hence, it still remains an open question to choose a local frequency interpretation appropriately for a given application, such as inpainting in our context, which is actively being investigated.

As the local information is often not enough for applications like inpainting, we also resort to the wisdom of  \textit{non-local self-similarity}~\cite{Buades05}. This idea is under the assumption of non-local stationarity, and has been widely used in image processing such as image denoising, inpainting, texture segmentation and authentication~\cite{Glew11,Bertalmio03,Criminisi04,Daribo12,Yang15,El02}. However, non-local self-similarity on graphs is challenging because it is difficult to define stationarity on \textit{irregular} graphs. Qi et al.~\cite{Qi11} introduce a weighted graph model to investigate the self-similarity characteristics of eubacteria genomes, but their graph signals are defined on regular Euclidean domains. Pang et al.~\cite{Pang14} define self-similarity in images via regular graphs, but there is no such definition for high-dimensional and irregular signal yet.

\begin{figure}[h]
	\centering
	\includegraphics[width=0.49\textwidth]{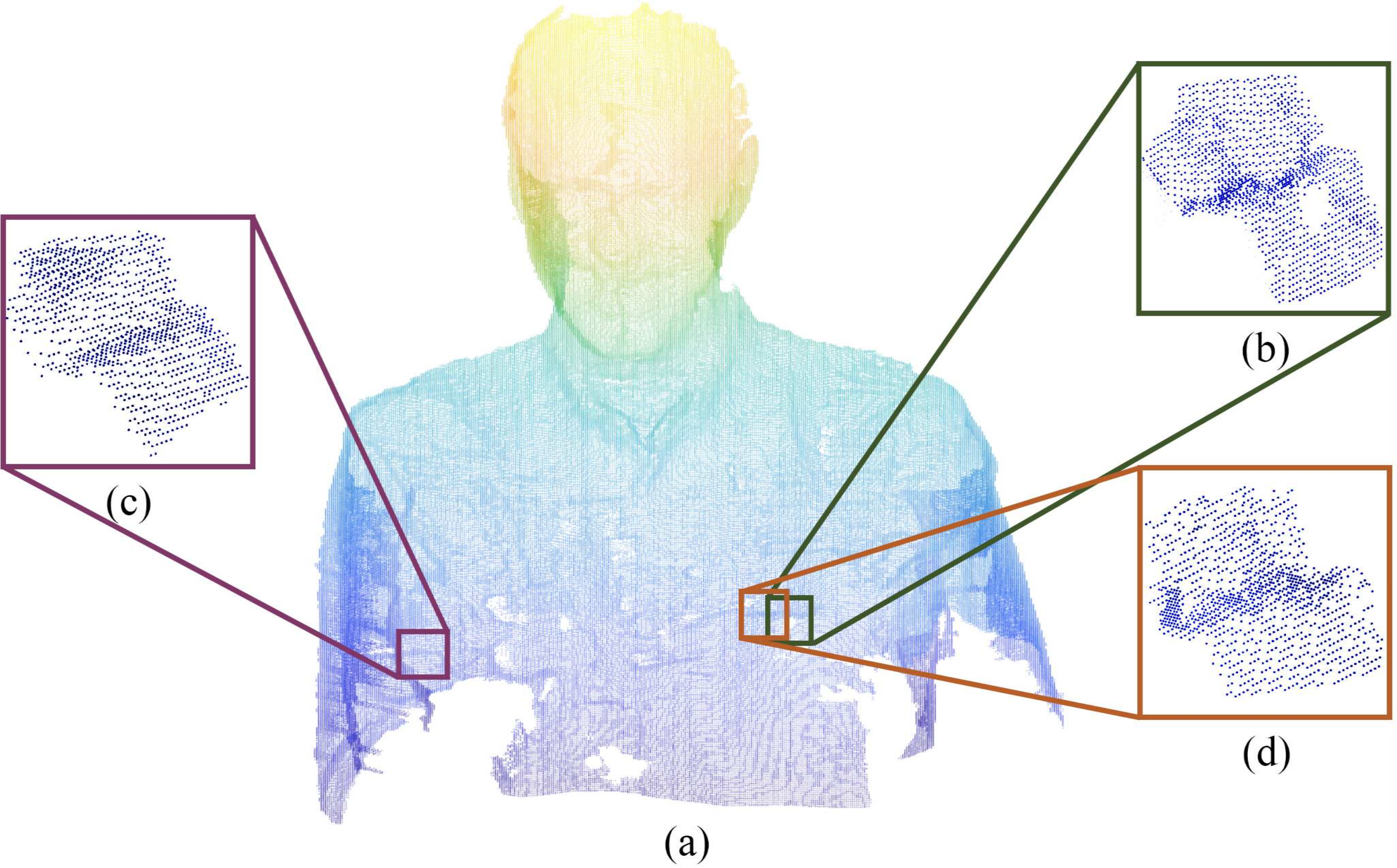}
	\caption{Point cloud \textit{Phili} (a) with three self-similarity examples (b) (c) (d). (b) is an example with a hole due to the inherent limitations of the acquisition equipments. (c) (d) are non-locally self-similar regions to (b).}
	\vspace{-0.1in}
	\label{fig:intro}
\end{figure}

In particular, we propose to exploit both local frequency interpretation and non-local self-similarity on graph for point cloud inpainting. Point clouds have received increasing attention as an efficient representation of 3D objects. It consists of a set of points, each of which corresponds to a measurement point with the 3D coordinates representing the geometric information and possibly attribute information such as color and normal.  The development of depth sensing and 3D laser scanning\footnote{Examples include Microsoft Kinect, Intel RealSense, LiDAR, etc. } techniques enables the convenient acquisition of point clouds, thus catalyzing its applications in various fields, such as 3D immersive tele-presence, navigation in autonomous driving, and heritage reconstruction~\cite{Tulvan16}. However, point clouds often exhibit several holes of missing data inevitably, as shown in Fig.~\ref{fig:intro} (b). This is mainly due to incomplete scanning views and inherent limitations of the acquisition equipments\footnote{For example, 3D laser scanning is less sensitive to the objects in dark colors. This is because the darker it is, the more wavelengths of light the surface absorbs and the less light it reflects. Thus the laser scanning devices are unable to receive enough reflected light for dark objects to recognize.}. Besides, there may lack some regions in the data itself (\textit{e.g.}, dilapidated heritage). Therefore, it is necessary to inpaint the incomplete point clouds prior to subsequent applications.

Our proposed point cloud inpainting method has three main contributions. Firstly, we present a frequency interpretaion in graph nodal domain, in which we prove the positive correlation between eigenvalues and the frequencies of eigenvectors for the graph Laplacian. Based on this correlation, we discuss a graph-signal smoothness prior from the perspective of suppressing high-frequency components so as to smooth and denoise the inpainting area at the same time.

Secondly, we propose a point cloud inpainting method by exploiting the non-local self-similarity in the geometry of point clouds, leveraging on our previous work~\cite{Fu18}. We globally search for a source region which is the most similar to the target region (\textit{i.e.}, the region containing the hole), and then fill the hole by formulating an optimization problem with the source region and the aforementioned graph-signal smoothness prior. In particular, we define the similarity metric between two regions on graph based on 1) the \textit{direct component} (DC) of the normals of points and 2) the \textit{anistropic graph total variation} (AGTV) of normals, which is an extension of the graph total variation~\cite{Shahid16,Berger17,Bai17}. Further, we improve our previous work by 1) mirroring the filtered cubes (the processing unit in our method) so as to augment non-local self-similar candidates; and 2) registering the geometric structure in the source cube and the target cube in order to strengthen the similarity between them for better inpainting, which includes the translation obtained by the difference in location between the source cube and the target cube and the rolation obtained by the proposed simplified Iterative Closest Points (ICP) method~\cite{Besl02,Chetverikov02}.

Thirdly, we propose voxelization and automatic hole detection prior to point cloud inpainting. Specifically, we perturb each point in the point cloud to regular grids so as to facilitate the subsequent processing. Thereafter, we design a hole detection method in the voxelized point cloud based on Breadth-First Search (BFS). We project the 3D point cloud to a 2D depth map along the principle  projection direction, search for holes in the 2D depth map via BFS, and finally obtain the 3D holes by calculating the coordinate range of the hole along the projection direction. Experimental results show that our approach outperforms four competing methods significantly, both in objective and subjective quality.

The outline of the paper is as follows. We first discuss previous methods in Section~\ref{sec:related}. Then we propose our frequency interpretation in Section~\ref{sec:graph} and overview the proposed point cloud inpainting framework in Section~\ref{sec:overview}. Next, we present our voxelization and hole detection methods in Section~\ref{sec:voxelize} and Section~\ref{sec:detect}, respectively. In Section~\ref{sec:method}, we elaborate on the proposed inpainting approach, including preprocessing, non-local similar cube searching, structure matching and problem formulation. Finally, experimental results and conclusion are presented in Section~\ref{sec:results} and \ref{sec:conclude}, respectively.

%% file: related.tex
We review the previous work on hole detection and inpainting respectively.

\subsection{Hole Detection}
Existing hole detection methods for point clouds include mesh-based methods and point-based methods. The idea of mesh-based methods is to construct a mesh over the point cloud, extract the hole boundary by some greedy strategy and treat the point not surrounded by the triangle chains as the boundary feature point~\cite{Emelyanov02,Orriols03,Floater01}. Nevertheless, the computational complexity is high, and the extracted boundaries are not accurate enough when the point distribution is uneven. Point-based methods are thus proposed to perform hole detection on the point cloud directly. \cite{Nguyen15,Aldeeb17,Chen17} extract the exterior boundary of the surface based on the neighborhood relationship among the 3D points, and then detect the hole boundary by a growth function on the interior boundary of the surface. However, these methods are unable to distinguish the hole from closed edges. Bendels et al.~\cite{Bendels06} compute a boundary probability for each point, and construct closed loops circumscribing the hole in a shortest-cost-path manner. \cite{He14} and \cite{Wu15} detect the K-nearest neighbors of each point and calculate the connection between the neighbors. When the angle between two adjacent edges reaches a threshold, the point is regarded as a boundary point. Nevertheless, these methods (\cite{Bendels06,He14,Wu15}) often treat points in sparse areas as the hole boundary, and the detected boundary points are likely to be incomplete and incoherent.

\subsection{Point Clouds Inpainting}
Few point clouds inpainting methods have been proposed, which include two main categories according to the cause of holes: (1) restore holes in the object itself such as heritage and sculptures~\cite{Sahay12,Sahay15,Setty15,Lozes15,Dinesh17}, and (2) inpaint holes caused by the limitation of scanning devices~\cite{Wang07,Lozes14,Lin17}. For the first class of methods, the main hole-filling data source is online database, as the holes are often large. Sahay et al.~\cite{Sahay12} attempt to fill big damage regions using neighbourhood surface geometry and geometric prior derived from registered similar examples in a library. Then they propose another method~\cite{Sahay15}, which projects the point cloud to a depth map, searches a similar one from an online depth database via dictionary learning, and then adaptively propagates the local 3D surface smoothness. However, the projection process inevitably introduces geometric loss. Instead, Dinesh et al.~\cite{Dinesh17} fill this kind of holes just with the data in the object itself. In particular, they give priority to the points along the hole boundary so as to determine the inpainting order, and find best matching regions based on an exemplar, which are used to fill the missing area. The results still suffer from geometric distortion due to the simple data source.

The other class of methods focus on holes generated due to the limitations of scanning devices. This kind of holes is smaller than the aforementioned ones in general, thus the information of the data itself is often enough for inpainting. Wang et al.~\cite{Wang07} create a triangle mesh from the input point cloud, identify the vicinity of the hole and interpolate the missing area based on a moving least squares approach. Lozes et al.~\cite{Lozes14} deploy partial difference operaters to solve an optimization equation on the point cloud, which only refers to the neighborhood of the hole to compute the geometric structure. Lin et al.~\cite{Lin17} segment a hole into several sub-holes, and then fill the sub-holes by extracting geometric features separately with tensor voting. Due to the reference information from the local neighborhood only, the results of these methods tend to be more planar than the ground truth. Also, artifacts are likely to occur around the boundary when the geometric structure is complicated.

In addition, some works deal with particular point cloud data. For instance, flattened bar-shaped holes in the human body data are studied in~\cite{Wu15}, where the holes are filled by calculating the extension direction of the boundary points from their neighboring points iteratively. \cite{Friedman12} mainly focuses on geometrically regular and planar point clouds of buildings from obstructions such as trees to fill the holes, which searches non-local similar blocks by performing Fourier analysis on each scanline in real-time scanning to detect the periodicity of the plane with scene elements. These methods have higher data requirements, thus they are unsuitable for general cases.

%% file: graph.tex
We first provide a review on basic concepts in spectral graph theory, including graph, graph Laplacian and graph-signal smoothness prior, and then propose the frequency interpretation of the prior in graph nodal domain, which will be utilized in the proposed point cloud inpainting.

\subsection{Preliminary}
\label{subsec:knn}

\begin{figure}[h]
	\centering
	\includegraphics[width=0.45\textwidth]{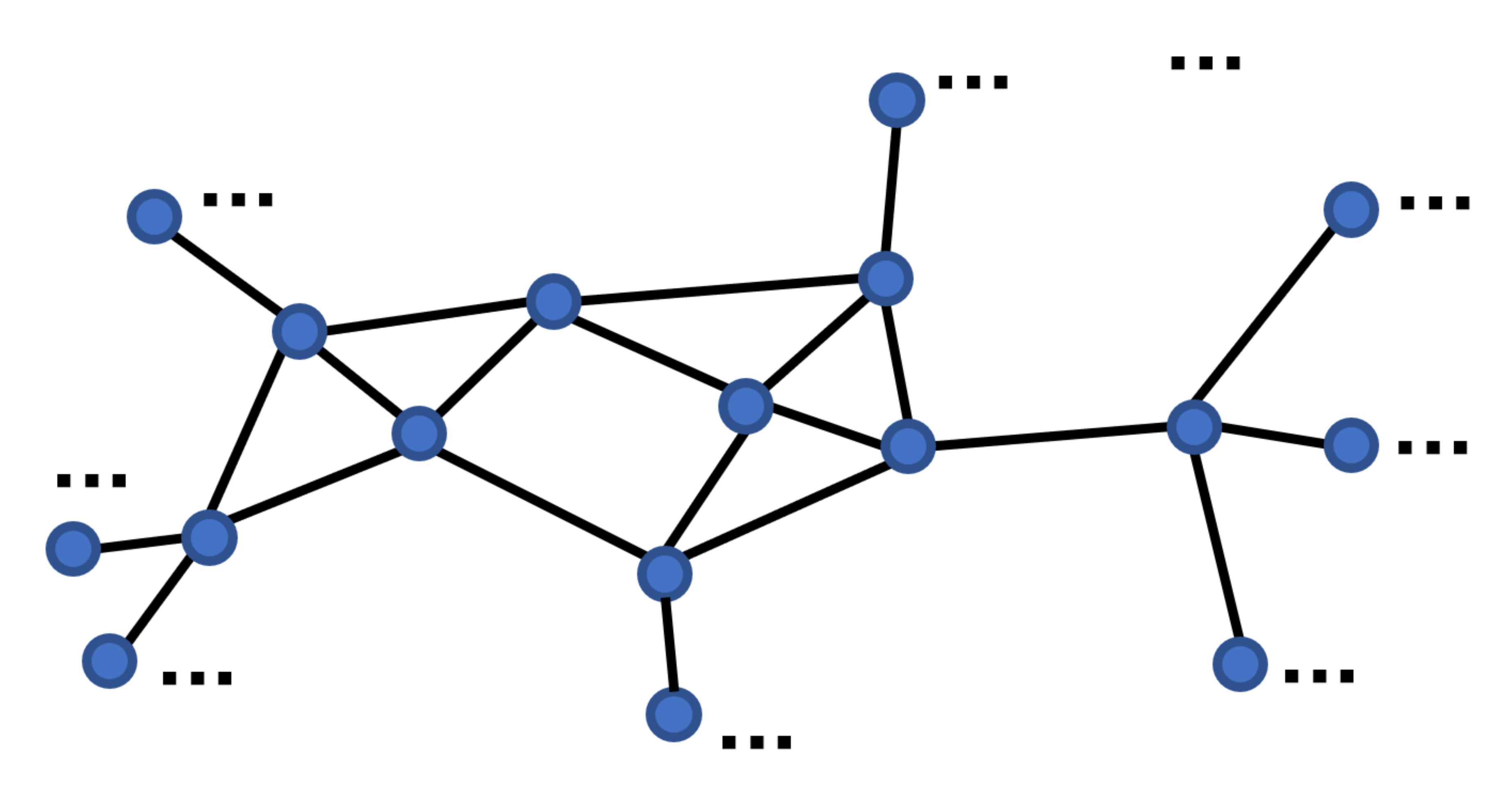}
	\vspace{-0.1in}
	\caption{A $ K $-NN graph constructed when $ K $= 4. The connections of boundary vertices are omitted.}
	\label{fig:knn}
\end{figure}

We consider an undirected graph $ \mathcal{G}=\{\mathcal{V},\mathcal{E},\mathbf{W}\} $ composed of a vertex set $ \mathcal{V} $ of cardinality $|\mathcal{V}|=N$, an edge set $ \mathcal{E} $ connecting vertices, and a weighted \textit{adjacency matrix} $ \mathbf{W} $. $ \mathbf{W} $ is a real symmetric $ N \times N $ matrix, where $ w_{i,j} $ is the weight assigned to the edge $ (i,j) $ connecting vertices $ i $ and $ j $. We assume non-negative weights, \textit{i.e.}, $w_{i,j} \geq 0$. For example, $ K $-Nearest Neighbor ($ K $-NN) graph is a common undirected graph, which is constructed by connecting each point with its nearest $ K $ neighbors, as shown in Fig.~\ref{fig:knn}.

The Laplacian matrix, defined from the adjacency matrix, can be used to uncover many useful properties of a graph~\cite{Chung96}. Among different variants of Laplacian matrices, the \textit{combinatorial graph Laplacian} used in \cite{Shen10,Hu12,Hu15} is defined as $ \mathcal{L}:=\mathbf{D}-\mathbf{W} $, where $ \mathbf{D} $ is the \textit{degree matrix}---a diagonal matrix where $ d_{i,i}=\sum_{j=1}^N w_{i,j} $. We choose to adopt the combinatorial graph Laplacian in our problem formulation, because it provides energy compaction in low-frequency components when integrated into the graph-signal smoothness prior, as discussed next. This property enables the structure-adaptive smoothing of the transition from the known region to the inpainted region.

\subsection{Graph-Signal Smoothness Prior and Its Frequency Interpretation}
\label{subsec:prior}
Graph signal refers to data residing on the vertices of a graph. For example, if we construct a $ K $-NN graph on the point cloud, then the normal of each point can be treated as graph signal defined on the $ K $-NN graph. This will be discussed further in the proposed cubic matching approach in Section~\ref{subsec:Search}.

A graph signal $ \vec{z} $ defined on a graph $ \mathcal{G} $ is smooth with respect to the topology of $ \mathcal{G} $ if
\begin{equation}
	\sum\limits_{i \sim j}w_{i,j}(z_i - z_j)^2 < \epsilon, ~~\forall i,j,
	\label{eq:prior}
\end{equation}
where $ \epsilon $ is a small positive scalar, and $ i \sim j $ denotes two vertices $ i $ and $ j $ are one-hop neighbors in the graph. In order to satisfy (\ref{eq:prior}), $ z_i $ and $ z_j $ have to be similar for a large edge weight $ w_{i,j} $, and could be quite different for a small $ w_{i,j} $. Hence, (\ref{eq:prior}) enforces $ \vec{z} $ to adapt to the topology of $ \mathcal{G} $, which is thus coined \textit{graph-signal smoothness prior}. As $ \vec{z}^T \mathcal{L} \vec{z} = \sum\limits_{i \sim j}w_{i,j}(z_i - z_j)^2 $ \cite{Spielman04}, (\ref{eq:prior}) is concisely written as $ \vec{z}^T \mathcal{L} \vec{z} < \epsilon $ in the sequel.

Despite the smoothing property, this prior also has denoising functionality. By definition of eigenvectors we have
\begin{equation}
	\mathcal{L} \mathbf{X} = \mathbf{X} \mathbf{\Lambda},
	\label{eq:eigendef}
\end{equation}
where $ \mathbf{X} = [ \vec{x}_1, \vec{x}_2, ..., \vec{x}_Q ] $ is the matrix of the $ Q $ eigenvectors $ \{\vec{x}_1, \vec{x}_2, ..., \vec{x}_Q\} $ of $ \mathcal{L} $, $ \mathbf{\Lambda} = diag [ \lambda_1, \lambda_2, ..., \lambda_Q ] $ is the matrix of eigenvalues of $ \mathcal{L} $ with $ 0 \leq \lambda_1 \leq \lambda_2 \leq ... \leq \lambda_Q $. Because $ \mathcal{L} $ is a real symmetric matrix, $ \mathbf{X}^{-1} = \mathbf{X}^T $. Then we have
\begin{equation}
	\mathcal{L} = \mathbf{X} \mathbf{\Lambda} \mathbf{X}^{-1} = \mathbf{X} \mathbf{\Lambda} \mathbf{X}^T.
	\label{eq:eigenbian}
\end{equation}

The eigenvectors are then used to define the graph Fourier transform (GFT). Formally, for any signal $ \vec{z} \in \mathbb{R}^N $ residing on the vertices of $ \mathcal{G} $, its GFT $ \bm{\eta} \in \mathbb{R}^N $ is defined in\cite{Hammond09} as
\begin{equation}
	\bm{\eta} = \mathbf{X}^T \vec{z},
	\label{eq:fourier}
\end{equation}
and the inverse GFT follows as
\begin{equation}
\vec{z} = \mathbf{X} \bm{\eta} = \sum\limits_{i} \eta_i \vec{x}_i.
\label{eq:infourier}
\end{equation}

Therefore, with (\ref{eq:fourier}) we have
\begin{equation}
	\vec{z}^T \mathcal{L} \vec{z} = \vec{z}^T ( \mathbf{X} \mathbf{\Lambda} \mathbf{X}^T ) \vec{z}
								  = \bm{\eta}^T \mathbf{\Lambda} \bm{\eta}
								  = \sum\limits_{i} \lambda_i \eta_i^2.
	\label{eq:fenjie}
\end{equation}

Hence, (\ref{eq:prior}) can be also written as $ \sum \lambda_i \eta_i^2 < \epsilon $. In order to satisfy this, $ \eta_i $ with larger $ \lambda_i $ will be smaller when minimizing (\ref{eq:prior}). Hence the term $ \eta_i \vec{x}_i $ in (\ref{eq:infourier}) with smaller $ \eta_i $ will be suppressed. Since the frequency of $ \vec{x}_i $, $ f(\vec{x}_i) $, is positively correlated with the eigenvalue $ \lambda_i $, which is denoted as $ f(\vec{x}_i) \propto \lambda_i $ and will be proved in Section~\ref{subsec:nodal}, the graph-signal smoothness prior suppresses high-frequency components, thus leading to a compact representation of $ \vec{z} $.

This prior will be deployed in our problem formulation of point cloud inpainting as a regularization term, as discussed in Section~\ref{sec:method}.

\subsection{Proof of Frequency Interpretation in Graph Nodal Domain}
\label{subsec:nodal}
While it is known that a larger eigenvalue of the graph Laplacian corresponds to an eigenvector with higher frequency in general, there is no rigorous mathematical proof yet to the best of our knowledge. Hence, we prove it from the perspective of nodal domains~\cite{Briandavies01} and have the following theorem:

\textbf{Theorem}~~~ Regarding the graph Laplacian, there is a positive correlation between eigenvalues and the frequencies of eigenvectors, by defining frequency as the number of zero crossings in graph nodal domains in particular.

\begin{figure}[h]
	\centering
	\includegraphics[width=0.5\textwidth]{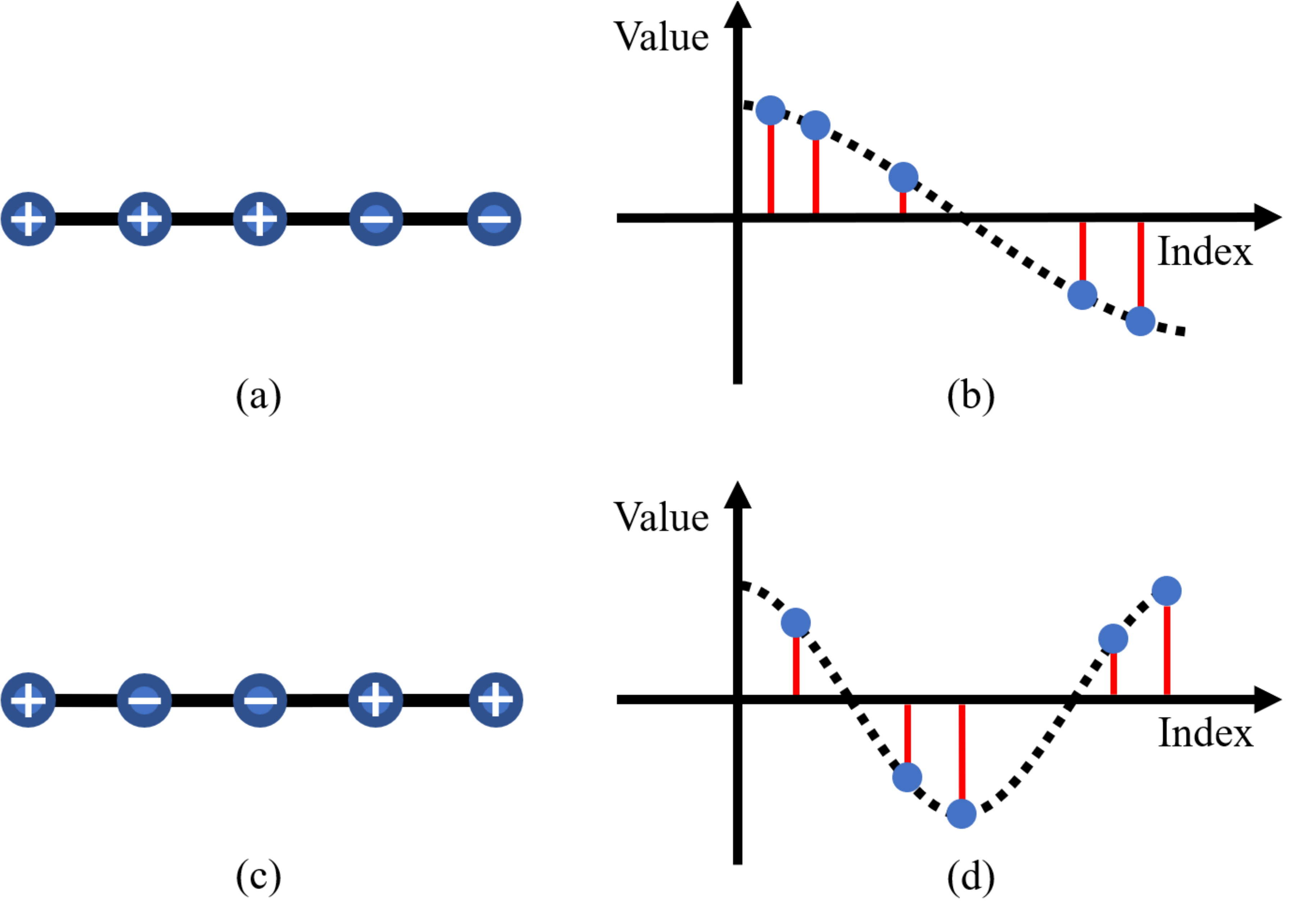}
	\vspace{-0.1in}
	\caption{Two 1D examples (row 1 and row 2) of the relationship between the number of zero crossings in graph nodal domains of the eigenvectors (a) (c) and their corresponding frequencies (b) (d), in which the horizontal axis is the index in the eigenvector and the vertical axis is the eigenvector value on each index. (a) has 1 zero crossing while (c) has 2, thus (a) corresponds to lower frequency than (c), as demonstrated in (b) and (d).}
	\label{fig:nodal}
\end{figure}

\textbf{Proof}~~~ A positive (negative) strong nodal domain of the eigenvector $ \vec{x}_i = \{x_{i,1},x_{i,2}, ...\} $ corresponding to $ \lambda_i $ on $ \mathcal{G} $ is a maximally connected subgraph such that $ x_{i,k} > 0~(x_{i,k} < 0) $ for all vertices $ k $ in the subgraph. A positive (negative) weak nodal domain of $ \vec{x}_i $ on $ \mathcal{G} $ is a maximally connected subgraph such that $ x_{i,k} \geq 0~(x_{i,k} \leq 0) $ for all vertices $ k $ in the subgraph, with $ x_{i,k} \neq 0 $ for at least one vertex $ k $ in the subgraph. As we know, the number of zero crossings in the signal can represent its frequency. Similarly, the number of zero crossings in nodal domains of the eigenvectors is able to interpret the frequency, as shown in Fig.~\ref{fig:nodal}.

In order to prove the positive correlation between the eigenvalue $ \lambda_i $ and the frequency of the corresponding eigenvector $ \vec{x}_i $, $ f(\vec{x}_i) $, we introduce the upper and lower bounds of the number of nodal domains of  $ \vec{x}_i $, $ \xi(\vec{x}_i) $:
\begin{itemize}
	\item Upper bound: $ \xi(\vec{x}_i) \leq i+u-1 $ ($ u $ is the multiplicity of $ \lambda_i $) for strong nodal domains and $ \xi(\vec{x}_i) \leq i $ for weak nodal domains~\cite{Briandavies01}. 
	\item Lower bound: $ \xi(\vec{x}_i) \geq i+u-1-q-z $ ($ q $ is the minimal number of edges that need to be removed from $ \mathcal{G} $ in order to turn it into a forest, $ z $ is the number of zero in $ \vec{x}_i $) for strong nodal domains~\cite{Xu13}.
\end{itemize}

Therefore, both the upper and lower bounds of $ \xi(\vec{x}_i) $ are monotonic in $ i $. This means that $ \xi(\vec{x}_i) \propto i $ is established in general, and thus $ f(\vec{x}_i) \propto \lambda_i $, because $ f(\vec{x}_i) $ is monotonic in the number of zero crossings in nodal domains $ \xi(\vec{x}_i) $ and $ \lambda_i $ is arranged monotonically with respect to $ i $.

%% file: overview.tex
\begin{figure*}
	\centering
	\includegraphics[width=\textwidth]{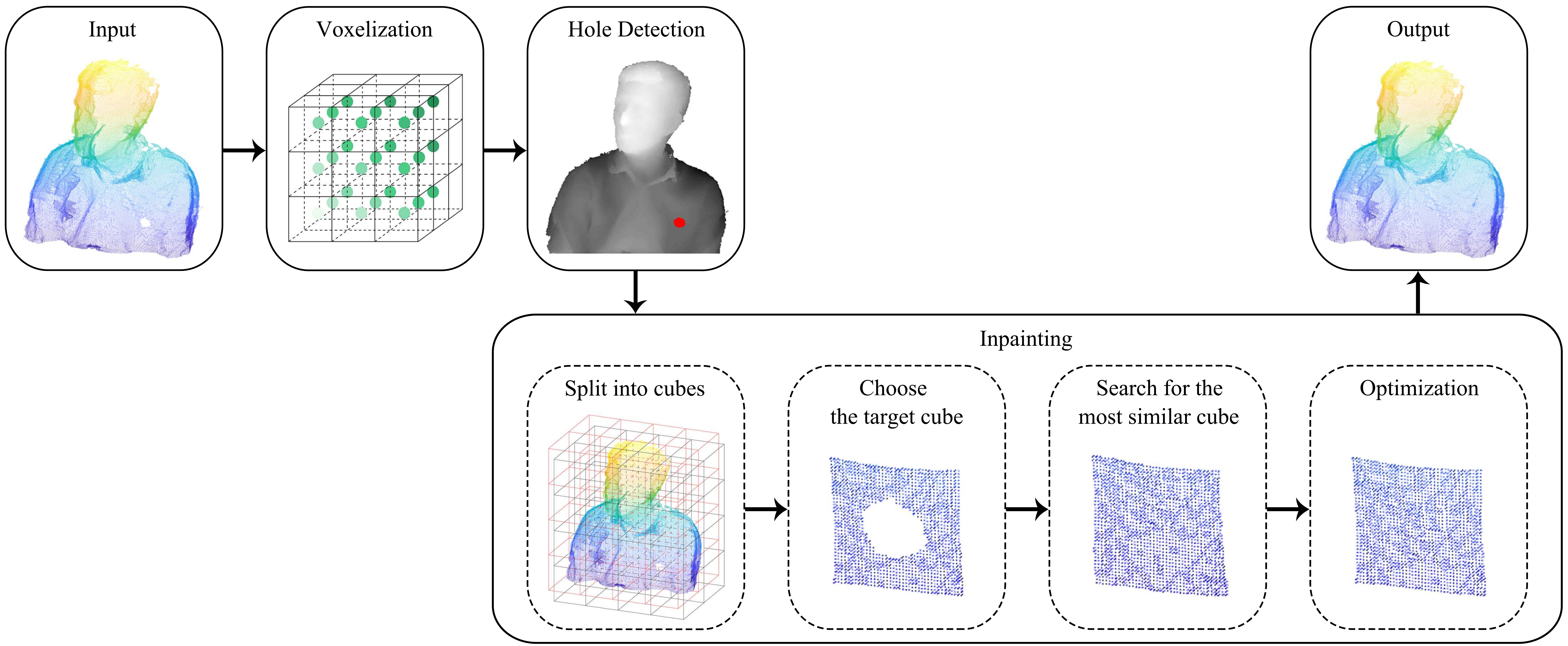}
	\caption{The framework of the proposed point cloud inpainting method.}
	\label{fig:pipe}
\end{figure*}

We now introduce the proposed point cloud inpainting method based on the spectral graph theory in Section~\ref{sec:graph}. As shown in Fig.~\ref{fig:pipe}, the method consists of the following steps:
\begin{itemize}
	\item Firstly, we \textit{voxelize} the input point clouds before inpainting to deal with the irregularity of point clouds. 
	\item Secondly, we detect the holes in the voxelized point cloud to inpaint in the next step.
	\item Thirdly, we inpaint the missing area. (a) We split it into cubes of fixed size as units to be processed in the subsequent steps. (b) We choose the cube with missing data as the target cube. (c) We search for the most similar cube to the target cube in the point cloud, which is referred to as the source cube, based on the DC and AGTV in the normals of points. (d) The inpainting problem is formulated into an optimization problem, which leverages the source cube and the graph-signal smoothness prior. The closed-form solution of the optimization problem leads to the resulting cube.
	\item Finally, we replace the target cube with the resulting cube as the output.
\end{itemize}

%% file: voxelize.tex
\begin{figure}[h]
	\centering
	\subfigure[]{
		\includegraphics[width=0.22\textwidth]{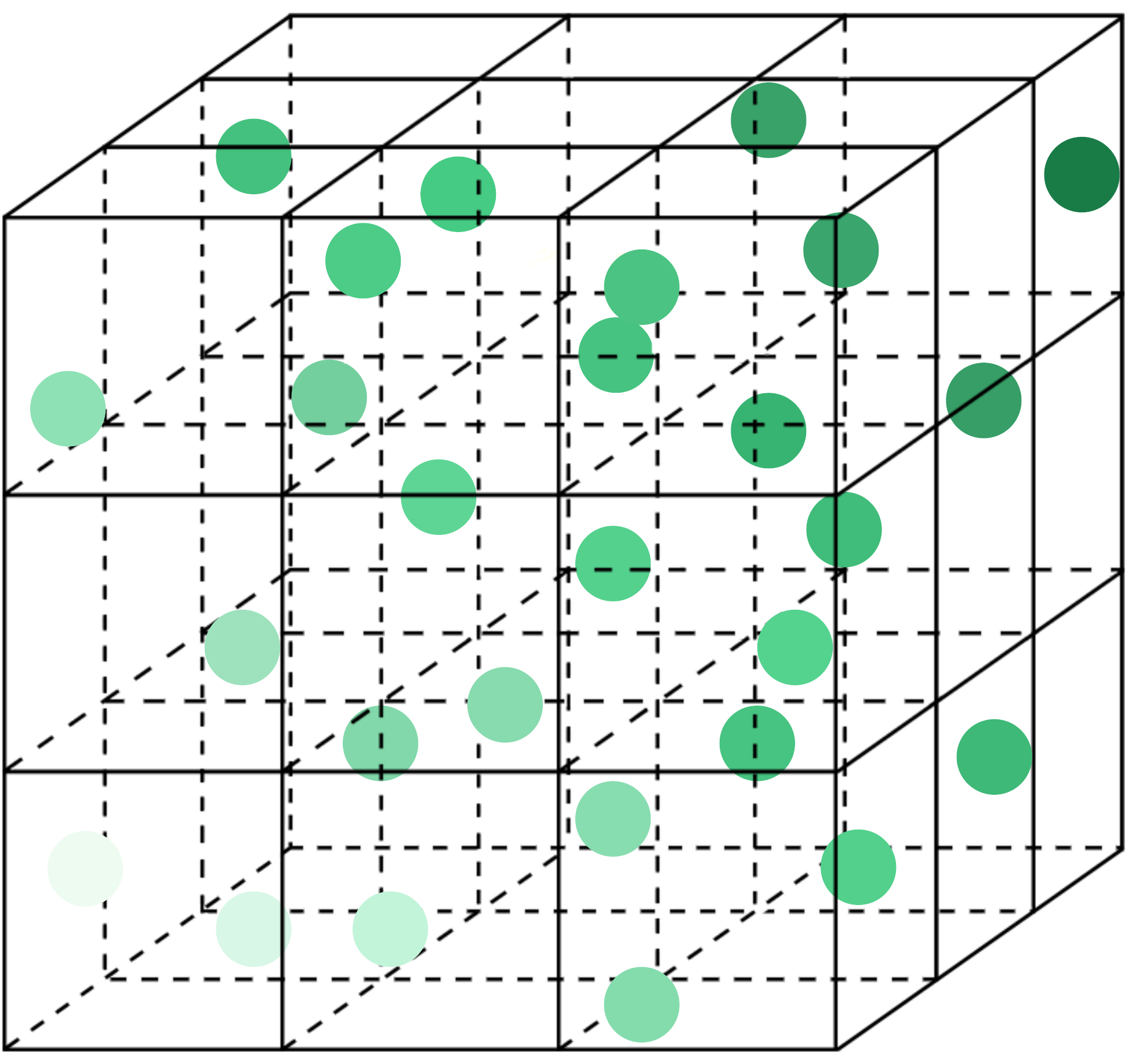}}
	\hspace{0.05in}
	\subfigure[]{
		\includegraphics[width=0.22\textwidth]{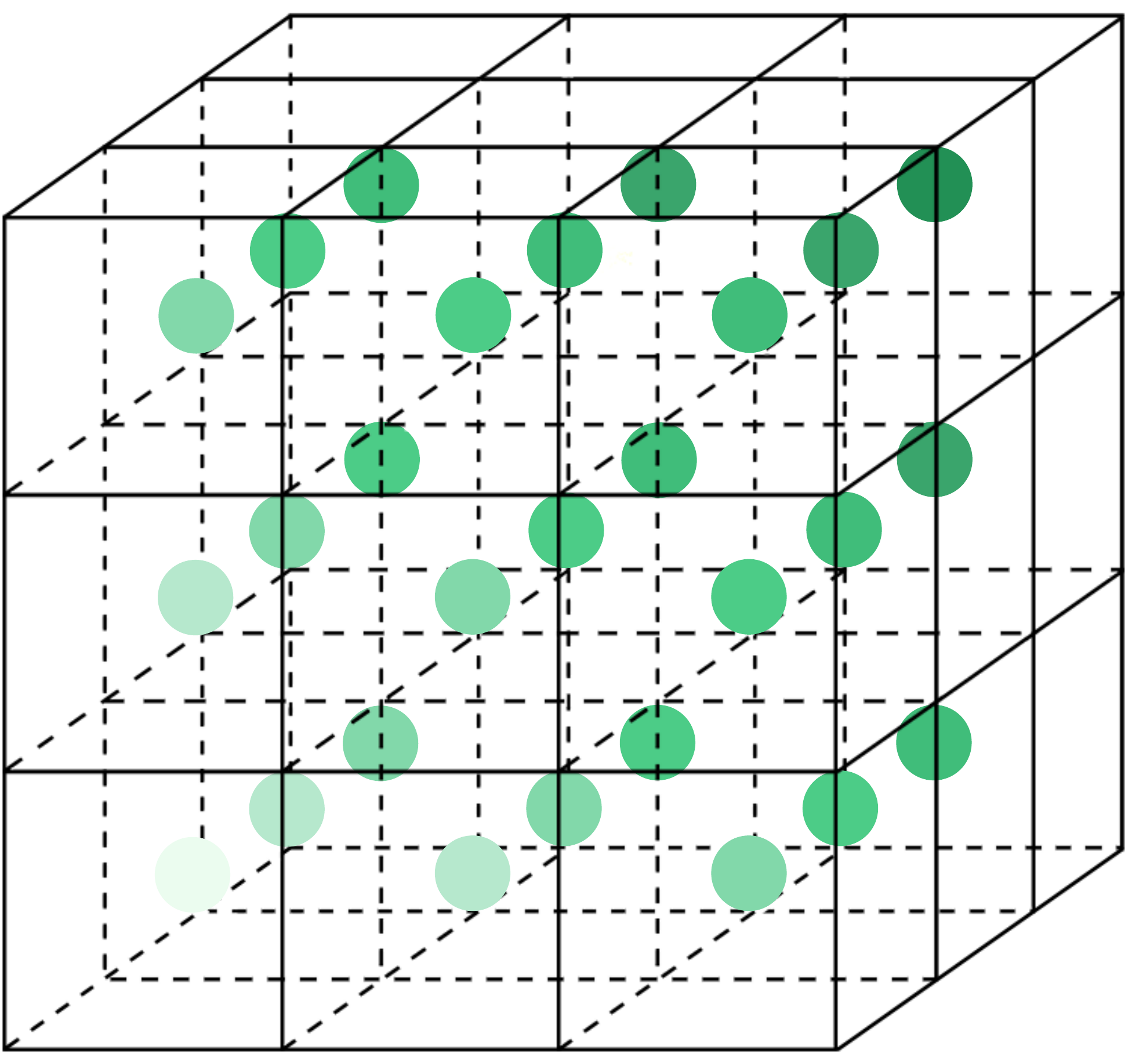}}
	\vspace{-0.05in}
	\caption{Demonstration of the proposed voxelization for point clouds. (a) The original point cloud. (b) The voxelized result.}
	\label{fig:voxelize}
\end{figure}

In order to address the irregularity of point clouds, we firstly voxelize the point clouds before inpainting, which means we perturb each point in the point cloud to regular grids, as shown in Fig.~\ref{fig:voxelize}. The voxelization step has the following advantages:
\begin{itemize}  
	\item Helps delineate missing areas accurately.
	\item Facilitates the determination of the exact locations of points to be filled in.
	\item Reduces the computational complexity caused by the irregularity of the point cloud.
\end{itemize}

The proposed voxelization consists of two steps. Firstly, in order to distribute the point uniformly in the voxels, we preprocess the coordinates to make the distance between each two neighboring points close to $ 1 $ and translate the point cloud to the origin. Then we voxelize them and compute the attribute information for each voxelized point.

\subsection{Coordinate Preprocessing}
For the convenience of the subsequent processing and data storage, we convert the coordinates into integers. We first expand or contract the original coordinates to make the distance between each two neighboring points close to $ 1 $. The input data is a point cloud denoted by $ \mathbf{P}^o\in \mathbb{R}^{n^o \times 3} $ ($ n^o $ is the number of points in the original point cloud) as the location of the points and $ \mathbf{H}^o\in \mathbb{R}^{n^o \times 3} $ as the attribute information of the points. We then construct a $ K $-NN graph mentioned in Section~\ref{subsec:knn} over the original point cloud, and compute the average of the distance between connected vertices as $ r $. The coordinates are then converted to $ \mathbf{P}^c $:
\begin{equation}
	\mathbf{P}^c=\frac{1}{r}\mathbf{P}^o.
	\label{eq:expand}
\end{equation}

Next, we translate $ \mathbf{P}^c $ to the origin of the coordinate axis for simplicity. The translation vector $ \vec{s}=(s_x,s_y,s_z) $ is computed as $ s_x=\text{min}(\mathbf{P}^c_x) $, where $ \mathbf{P}^c_x $ represents the $ x $ coordinates in $ \mathbf{P}^c $. $ s_y $ and $ s_z $ are defined in the same way. Then we obtain the $ \vec{s} $-translated location $ \mathbf{P}^t $ as
\begin{equation}
	\mathbf{P}^t=\mathbf{P}^c-\mathbf{S},
	\label{eq:pretrans}
\end{equation}
where $ \mathbf{S} $ is a $ n^o \times 3 $ translation matrix with $ \vec{s} $ repeated in each row.

\subsection{Voxel Computation}
We split $ \mathbf{P}^t $ into voxels with side length equal to $ 1 $. Thus a voxel consists of a set of points $ \{\vec{q}_1,\vec{q}_2, ... $ $ \} $ with $ \vec{q}_i\in \mathbb{R}^3 $ meaning the coordinates of the $ i $-th point in the voxel, each of which corresponds to the attribute information denoted as $ \{\vec{g}_1,\vec{g}_2, ... $ $ \} $. In each voxel, we replace the set of points with the centering point of the voxel as long as there are points within, thus obtaining the voxelized point cloud $ \mathbf{P}^v $. Then we compute the attribute value $ \vec{h} $ for each voxel with the centering location $ \vec{p} $ as
\begin{equation}
	\vec{h}=\frac{1}{\sum \mu_i}\sum \mu_i\vec{g}_i,
	\label{eq:voxelcolor}
\end{equation}
where $ \mu_i $ is the weight for the attribute of each point $ \vec{q}_i $ based on the geometric distance to the centering point $ \vec{p} $:
\begin{equation}
	\mu_i=\text{exp}\{-\frac{\norm{\vec{q}_i-\vec{p}}_2^2}{2\sigma^2}\},
	\label{eq:voxelweight}
\end{equation}
where $\sigma$ is a weighting parameter (empirically $\sigma=1$ in our experiments). This is based on the assumption that geometrically closer points have more similar attributes in general.

The voxelized point cloud, with $ \mathbf{P}^v $ as its locations and $ \mathbf{H}^v $ as its attributes (normal at each point in our case), is the final input data for the subsequent processing.

%% file: detect.tex
\begin{figure}[h]
	\centering
	\subfigure[]{
		\includegraphics[width=0.30\textwidth]{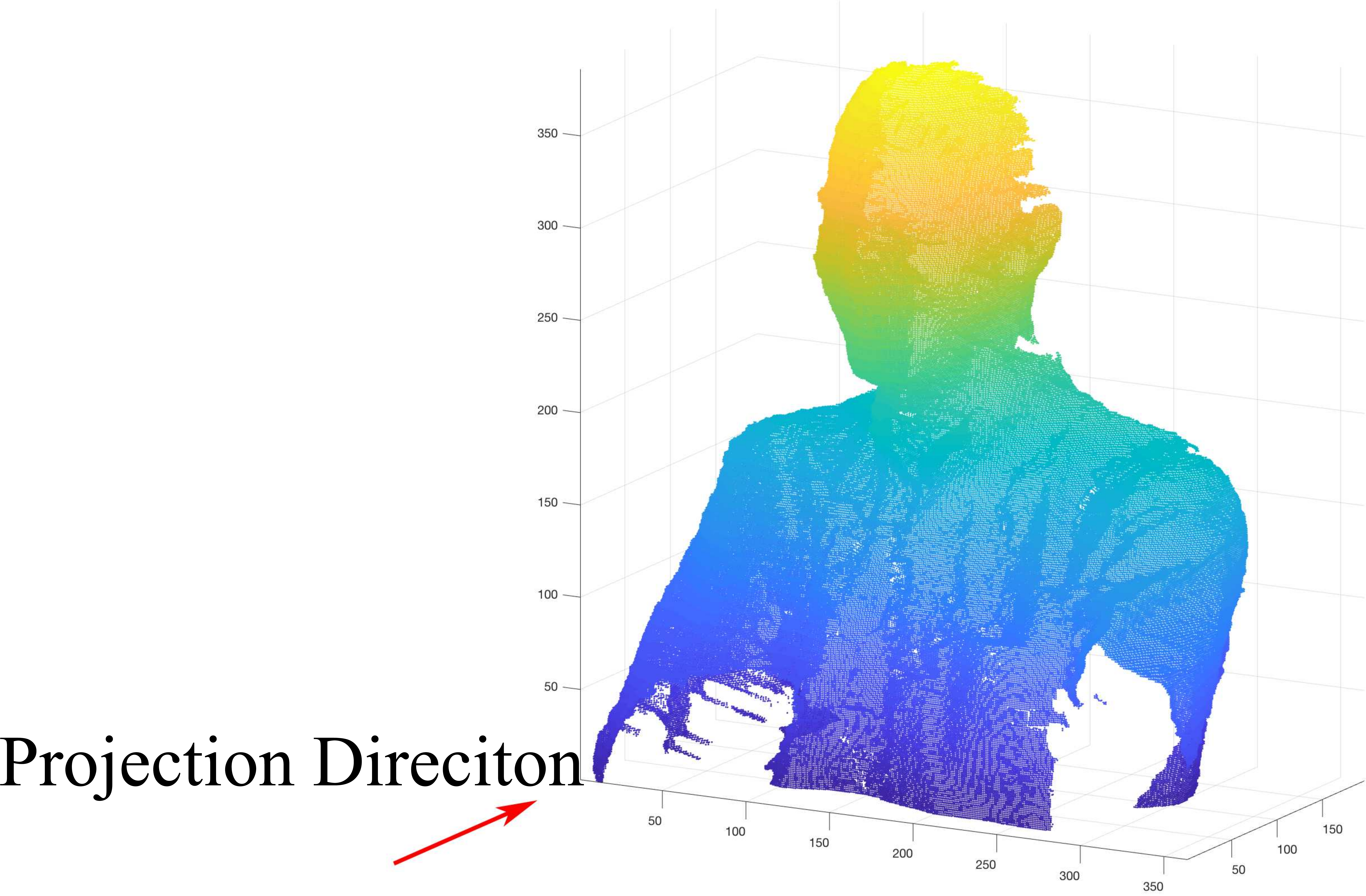}}
	\hspace{0.02in}
	\subfigure[]{
		\includegraphics[width=0.16\textwidth]{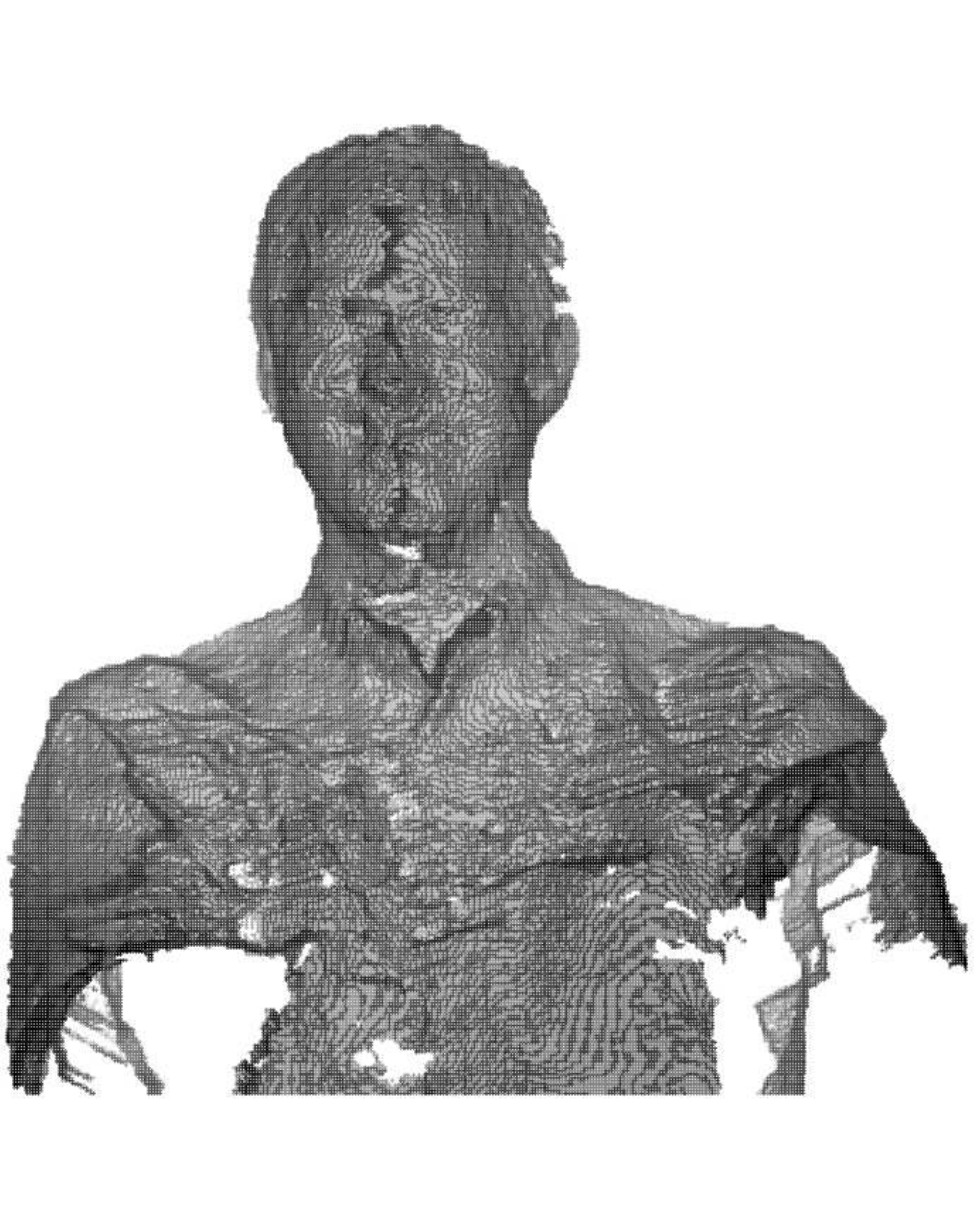}}
	\caption{The projection step for point clouds for automatic hole detection. (a) The original point cloud. (b) The depth map obtained from the principle projection direction.}
	\label{fig:project}
\end{figure}

We propose hole detection mainly based on BFS~\cite{Bundy84}. As shown in Fig.~\ref{fig:project}, we first project the 3D point cloud to a 2D depth map along a projection direction denoted by $ \vec{r}_p $. While $ \vec{r}_p $ can be all the six projection directions along $ \{x,y,z\} $ axes, we deploy Principal Component Analysis (PCA)~\cite{Abdi10} on the normals to choose the best projection direction as $ \vec{r}_p $ in order to save the computational complexity. PCA statistically transforms a set of variables that may be related to each other into a set of linearly uncorrelated variables by orthogonal transformation. The converted set of variables is the principal component, which is the projection direction in our method. Then we search for holes using BFS, and differentiate holes by marking points in each hole with different numbers. Finally, we determine the length of each detected hole along $ \vec{r}_p $ by calculating the coordinate range of each hole along $ \vec{r}_p $, which gives the hole region in the 3D point cloud. We elaborate on the BFS-based detection method as follows.

\begin{figure}[h]
	\centering
	\subfigure[]{
		\includegraphics[width=0.23\textwidth]{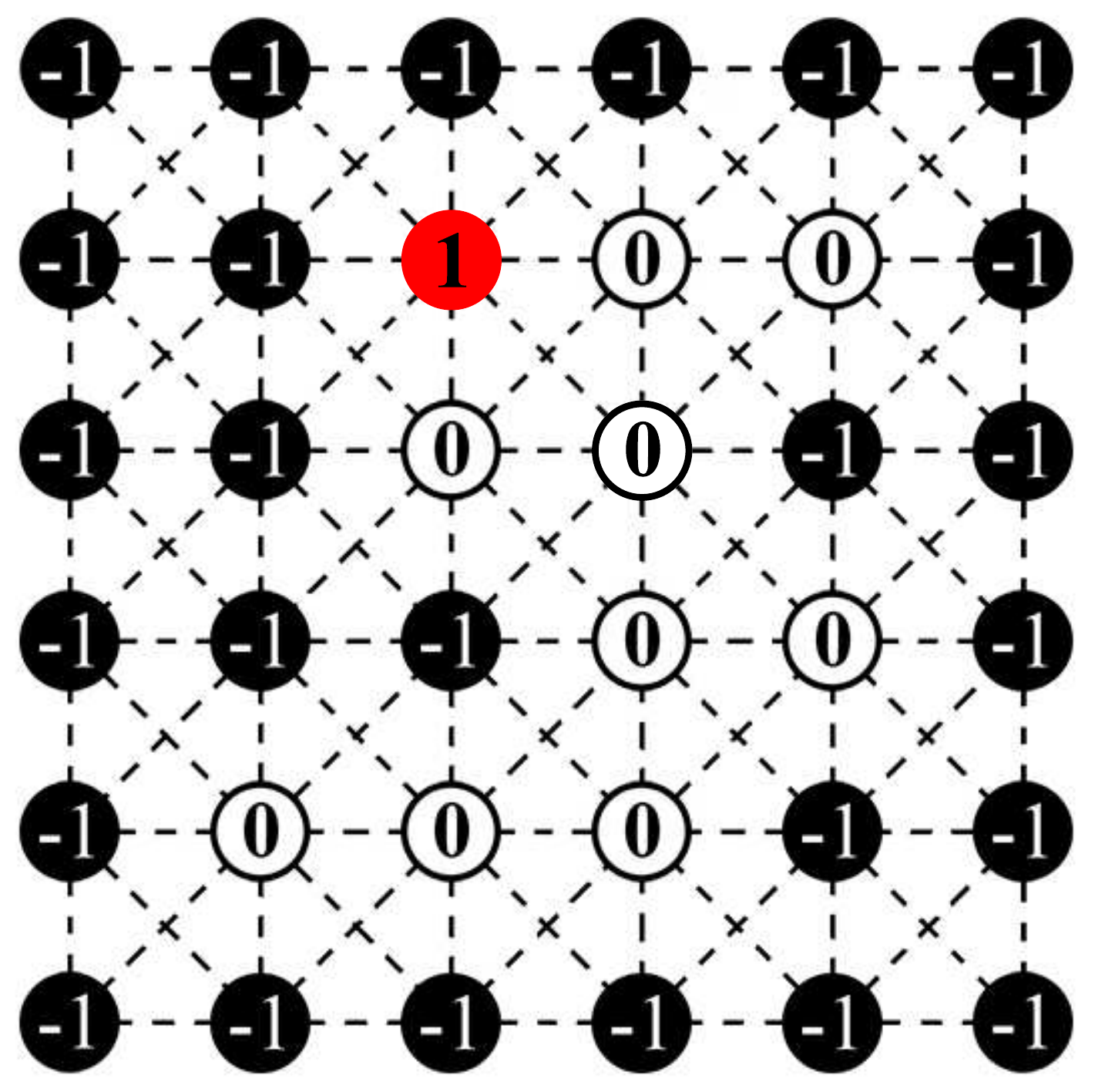}}
	\hspace{0.02in}
	\subfigure[]{
		\includegraphics[width=0.23\textwidth]{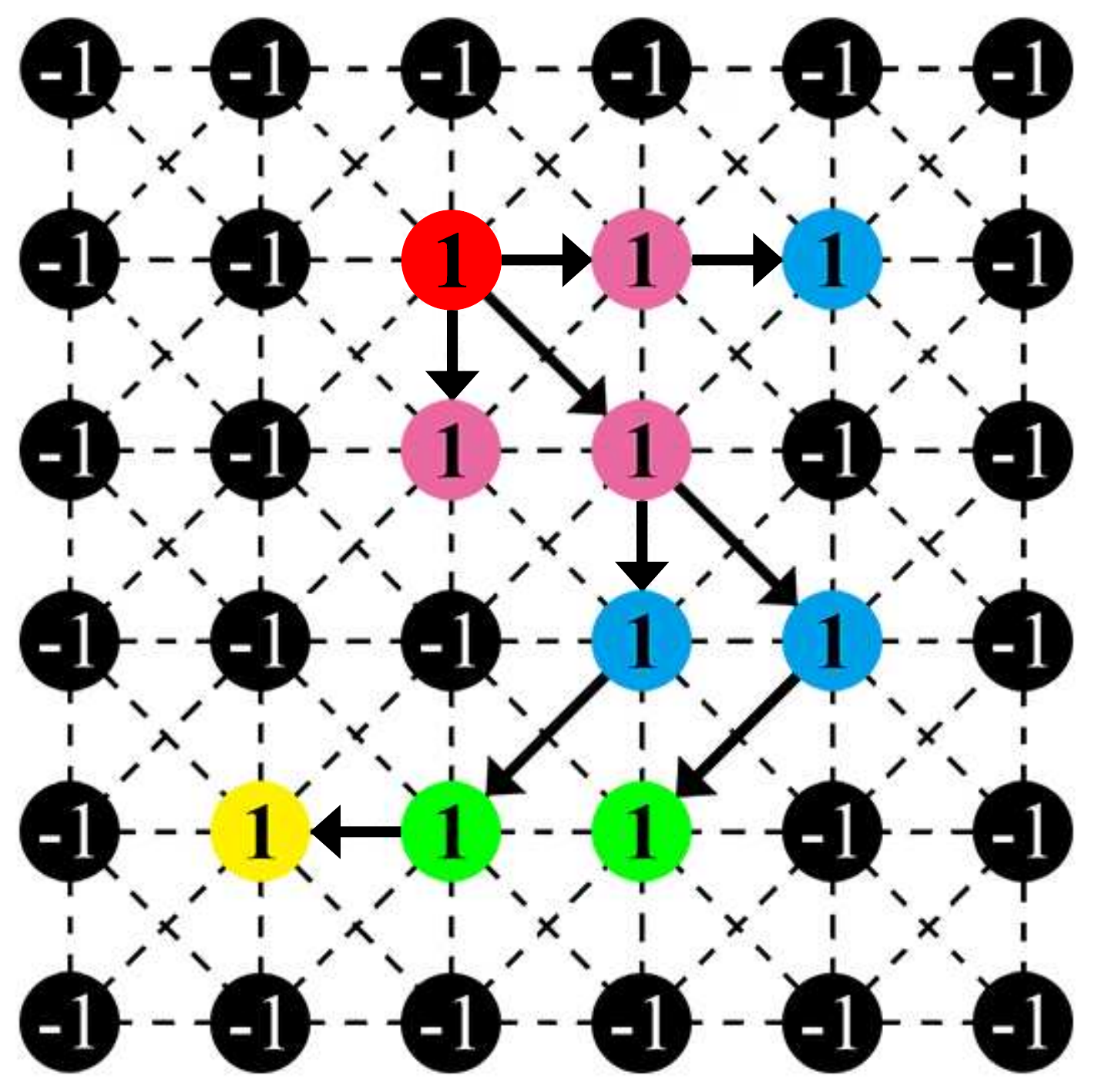}}
	\caption{The marking step for hole detection using BFS-based method. (a) The initial marking situation around the hole at the beginning of the detection, where the first $ \mathbf{1} $-marked pixel is marked in red. (b) The final marking situation around the hole at the end of the detection, where the first $ \mathbf{1} $-marked pixel is marked in red, the second in pink, the third in blue, the fourth in green and the fifth in yellow.}
	\label{fig:bfs}
\end{figure}

There are two kinds of pixels in the obtained depth map: pixels corresponding to points in the point cloud and pixels with no correspondences (i.e., corresponding to holes). Hence, we propose to detect a hole in the depth map by searching a connected set of pixels with no correspondences, where two pixels are considered connected if they have similar values and locate in one-hop neighboring locations in horizontal, vertical or diagonal directions.
We solve this problem by BFS as shown in Fig.~\ref{fig:bfs}. We first mark pixels with correspondence as $ \mathbf{-1} $ and the other pixels as $ \mathbf{0} $. Then we traverse pixels in the depth map until we find the first $ \mathbf{0} $-marked pixel, and change the mark to $ \mathbf{1} $ for the first hole (similarly, $ \mathbf{2} $ for the second hole, $ \mathbf{3} $ for the third hole, etc.) so as to distinguish different holes. Next, we search for the connected $ \mathbf{0} $-marked pixels among its one-hop neighbors, and update their marks to $ \mathbf{1} $. We repeat this operation until there are no $ \mathbf{0} $-marked neighbors of the $ \mathbf{1} $-marked pixels. Thus the first hole is detected. Similarly, we continue to detect the second hole by searching the next $ \mathbf{0} $-marked pixel in the depth map, and change its mark to $ \mathbf{2} $. All the $ \mathbf{0} $-marked pixels in the neighborhood are then updated to $ \mathbf{2} $. The same goes for the rest of the holes.

Now we detect all the holes based on BFS. However, some of them only have several pixels in the depth map, which might be detected due to the low density of the point cloud. Therefore, we filter out the holes with the number of missing pixels less than a density-dependent threshold (which is empirically set as 4 because the distance between two adjacent points after voxelization is less than 2 generally). Besides, some detected holes are actually the pixels corresponding to the outside of the object. These holes are distinguished from actual holes by checking if they include any point at the boundary of the depth map and then filtered out. The rest of holes will be inpainted.

%% file: method.tex
We now introduce the proposed point cloud inpainting method based on the spectral graph theory in Section~\ref{sec:graph}. As shown in Fig.~\ref{fig:pipe}, the input data is a point cloud denoted by $ \mathbf{P}^v=\{\vec{p}_1,\vec{p}_2, ...,\vec{p}_n\} $ with $ \vec{p}_i\in \mathbb{R}^3 $ meaning the coordinates of the $ i $-th point in the point cloud. Firstly, we split it into fixed-sized cubes as units to be processed in the subsequent steps. Secondly, we choose the target cube with missing area detected in Section~\ref{sec:detect}. Thirdly, we search for the most similar cube to the target cube in $ \mathbf{P}^v $, which is referred to as the source cube, based on the DC and AGTV in the normals of points. Fourthly, the inpainting problem is formulated into an optimization problem, which leverages the source cube and the graph-signal smoothness prior. The closed-form solution of the optimization problem gives the resulting cube. Finally, we replace the target cube with the resulting cube as the output.

\subsection{Preprocessing}
\label{subsec:Cutcube}
We first split the input point cloud into overlapping cubes $ \{\vec{c}_1,\vec{c}_2, ... $ $ \} $ with $ \vec{c}_i\in \mathbb{R}^{M^3 \times 3} $ ($ M $ is the size of the cube),  as the processing unit of the proposed inpainting algorithm. $ M $ is empirically set according to the coordinate range of $ \mathbf{P}^v $ ($ M=20 $ in our experiments), while the overlapping step is empirically set as $ \frac{M}{4} $. This is a trade-off between the computational complexity and ensuring enough geometry information available to search for the most similar cube.

Having obtained the cubes, we choose the cube with missing data as the target cube $ \vec{c}_t $. Further, in order to save the computation complexity and increase the accuracy of the subsequent cube matching, we choose candidate cubes $ \vec{c}_c $ by filtering out cubes with the number of points less than 80\% of that of $ \vec{c}_t $, and \textit{augment} the candidates by mirroring these cubes with respect to the \textit{x-y} plane, which will be used in the next step.

\subsection{Non-Local Similar Cube Searching}
\label{subsec:Search}
In order to search for the most similar cube to the target, we first define the geometric similarity metric $ \delta(\vec{c}_t,\vec{c}_c) $ between the target cube $ \vec{c}_t $ and each candidate cube $\vec{c}_c $ as
\begin{equation}
	\delta(\vec{c}_t,\vec{c}_c)=
		\text{exp}\{-[
			\delta_D(\vec{c}_t,\vec{c}_c)
			+\delta_V(\vec{c}_t,\vec{c}_c)
		]\},
	\label{eq:simi}
\end{equation}
where $ \delta_D(\vec{c}_t,\vec{c}_c) $ and $ \delta_V(\vec{c}_t,\vec{c}_c) $ are the difference in DC and AGTV between $ \vec{c}_t $ and $ \vec{c}_c $ respectively. Specifically, they are defined as
\begin{equation}
	\delta_D(\vec{c}_t,\vec{c}_c)=
		~|<\vec{d}(\vec{c}_t),~\vec{d}(\vec{c}_c)>|,
\end{equation}
\begin{equation}
	\delta_V(\vec{c}_t,\vec{c}_c)=
		|v(\vec{c}_t)-v(\vec{c}_c)|,
\end{equation}
where $ \vec{d}(\vec{c}_t) $ and $ \vec{d}(\vec{c}_c) $ are the DC of cube $ \vec{c}_t $ and $ \vec{c}_c $, while $ v(\vec{c}_t) $ and $ v(\vec{c}_c) $ are the AGTV of $ \vec{c}_t $ and $ \vec{c}_c $. We explain $ \delta_D(\vec{c}_t,\vec{c}_c) $ and $ \delta_V(\vec{c}_t,\vec{c}_c) $ in detail as follows.

\vspace{0.05in}
\textbf{Direct Component}~~~This is a function of the normals of points in the cube, which presents the \textit{prominent geometry direction} of the cube. A cube $ \vec{c}_i $ consists of a set of points $ \{c_{i,1},c_{i,2}, ...,c_{i,m}\} $ ($ m $ is the number of the points), each of which corresponds to a normal, denoted as $ \{\vec{n}_{i,1}, \vec{n}_{i,2}, ..., \vec{n}_{i,m}\} $. We compute the DC of the cube  $ \vec{c}_i $ as
\begin{equation}
	\vec{d}(\vec{c}_i)=
		\frac
			{\sum_{k=1}^m\vec{n}_{i,k}}
			{\norm{\sum_{k=1}^m\vec{n}_{i,k}}_2^2}.
\end{equation}

\vspace{0.05in}
\textbf{Anisotropic Graph Total Variation}~~~Unlike traditional total variation used in image denoising~\cite{Chambolle04} and edge extraction~\cite{Khabipova12}, graph total variation (GTV) generalizes to describe the smoothness of a graph signal with respect to the underlying graph structure. As we have the local gradient $ \nabla_i \vec{z} \in \mathbb{R}^N $  of a graph signal $ \vec{z} $ at node $ i \in \mathcal{V} $ whose $ j $-th element is defined as
\begin{equation}
	(\nabla_i \vec{z})_j \triangleq (z_j-z_i) w_{i,j},
\end{equation}
the conventional isotropic GTV~\cite{Berger17} is defined as
\begin{equation}
	\norm{\vec{z}}_{TV}
		\triangleq \sum\limits_{i \in \mathcal{V}} \norm{\nabla_i \vec{z}}_2
		= \sum\limits_{i \in \mathcal{V}} \sqrt{\sum\limits_{j \in \mathcal{V}} (z_j-z_i)^2 w_{i,j}^2}.
\end{equation}

Further, in order to describe the geometric feature of point clouds more accurately, we deploy the AGTV based on the normals in a cube instead of the conventional graph total variation. The mathematical definition of the AGTV in our method for $ \vec{c}_i $ is
\begin{equation}
	v(\vec{c}_i)=
		\frac
			{\sum\limits_{k}\sum\limits_{l}|<\vec{n}_{i,k},\vec{n}_{i,l}>|w_{k,l}}
			{K(K-1)}.
\end{equation}

Here $ w_{k,l} $ denotes the weight of the edge between $ k $ and $ l $ in the graph we construct over $ \vec{c}_i $. Specifically, we choose to build a $ K $-NN graph mentioned in Section~\ref{subsec:knn}, based on the affinity of geometric distance among points in $ \vec{c}_i $. Also, we consider unweighted graphs for simplicity, which means $ w_{k,l} $ is assigned as
\begin{equation}
	w_{k,l}=
		\left\{\begin{array}{lr}  
			1, & k \sim l\\
			0, & \text{otherwise}
		\end{array}\right.
	\label{eq:graph}
\end{equation}

Note that the AGTV is the $ l_1 $ norm of the overall difference in normals in the graph. Hence, this favors the sparsity of the overall graph gradient in normals while the conventional GTV favors the sparsity of the local gradients. Therefore, the AGTV is able to describe abrupt signal changes efficiently.

Having computed the similarity metric in (\ref{eq:simi}) between the target cube and candidate cubes, we choose the candidate cube with the largest similarity as the source cube $ \vec{c}_s $. However, $ \vec{c}_s $ cannot be directly adopted for inpainting, because it is just the most similar to $ \vec{c}_t $ in the geometric structure, but not in the relative location in the cube. Hence, we further perform structure matching (i.e., coarse registration) for $ \vec{c}_s $ and $ \vec{c}_t $ so as to match the relative locations, as discussed in the following.

\subsection{Structure Matching}
\label{subsec:Rotation}
\begin{figure}[h]
	\centering
	\subfigure[]{
		\includegraphics[width=0.15\textwidth]{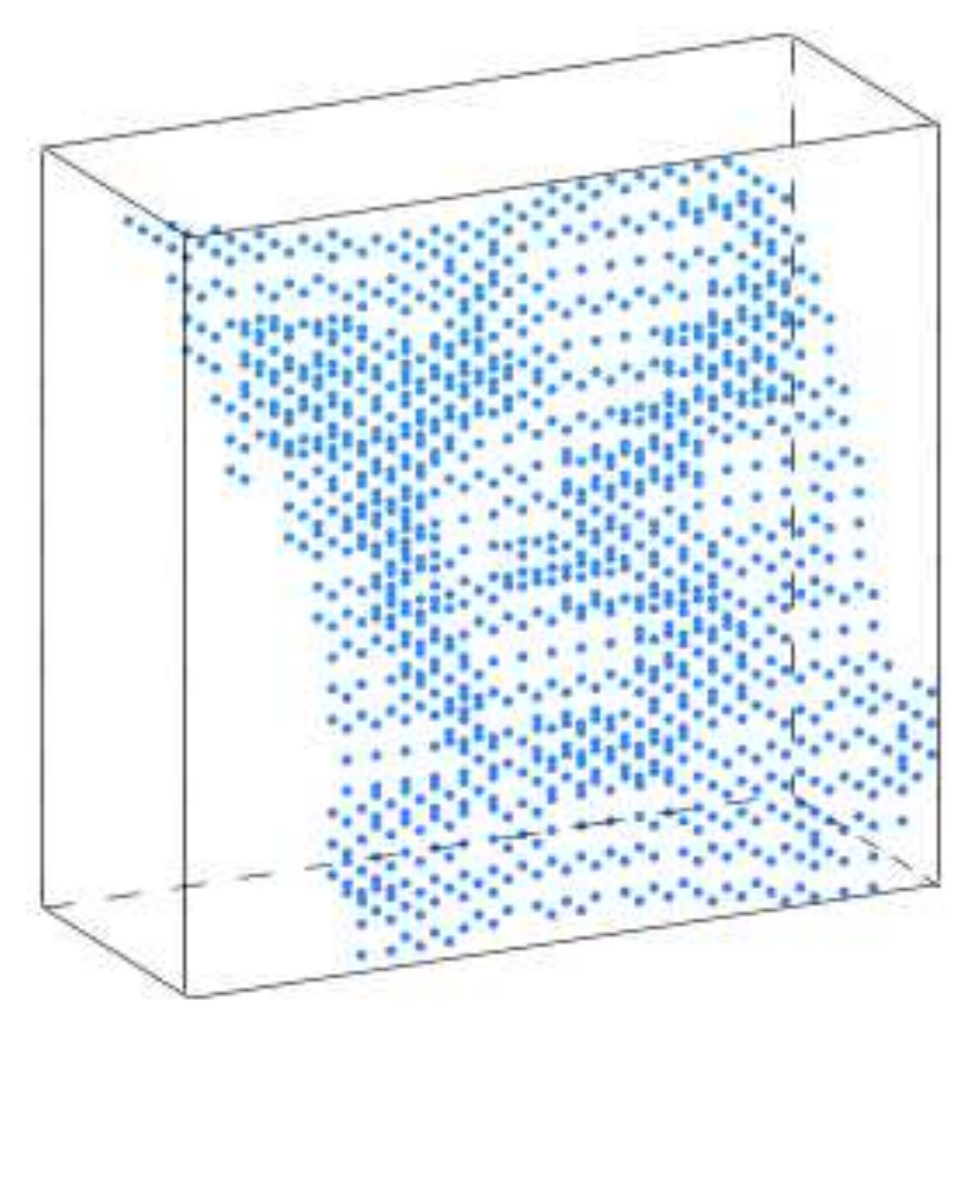}}
	\subfigure[]{
		\includegraphics[width=0.15\textwidth]{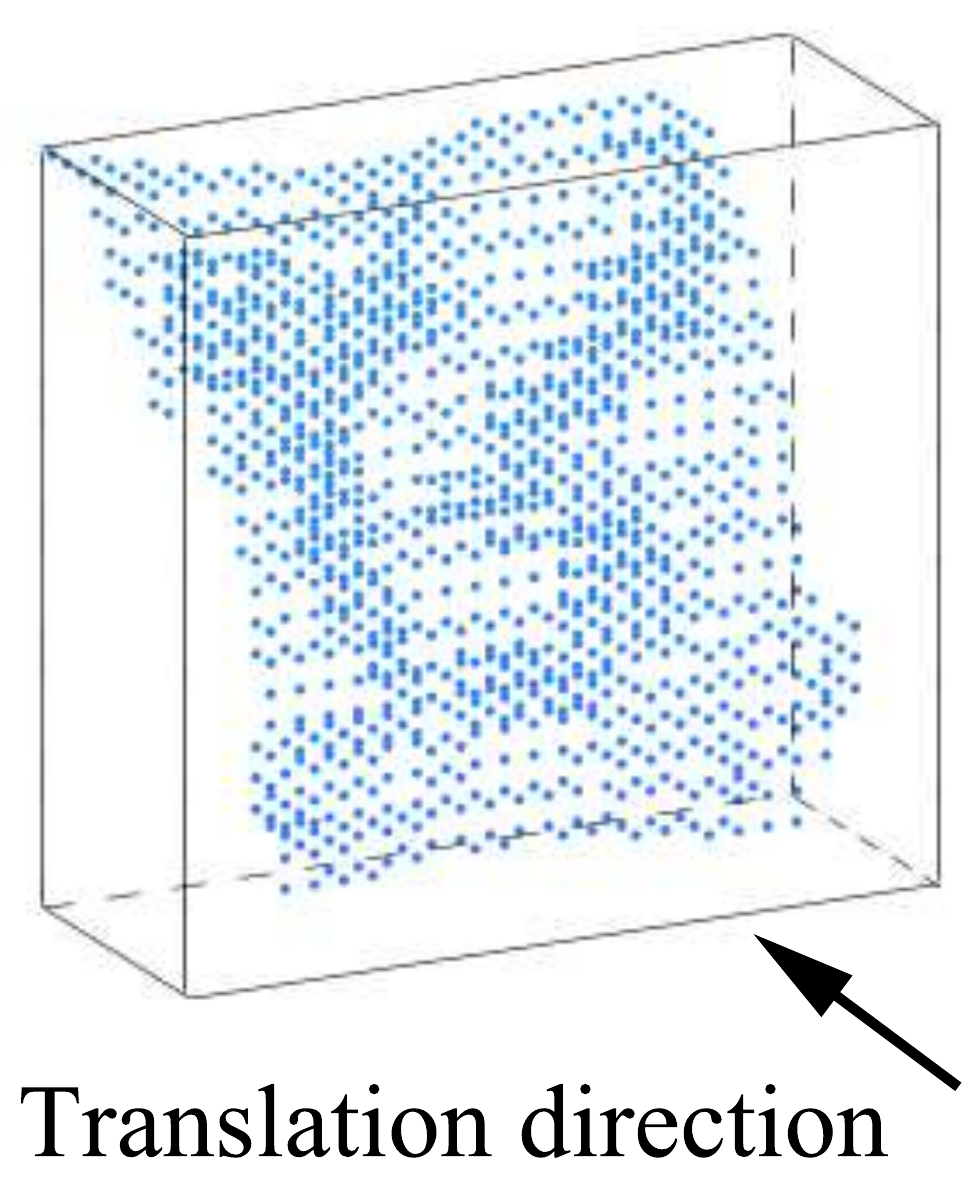}}
	\subfigure[]{
		\includegraphics[width=0.15\textwidth]{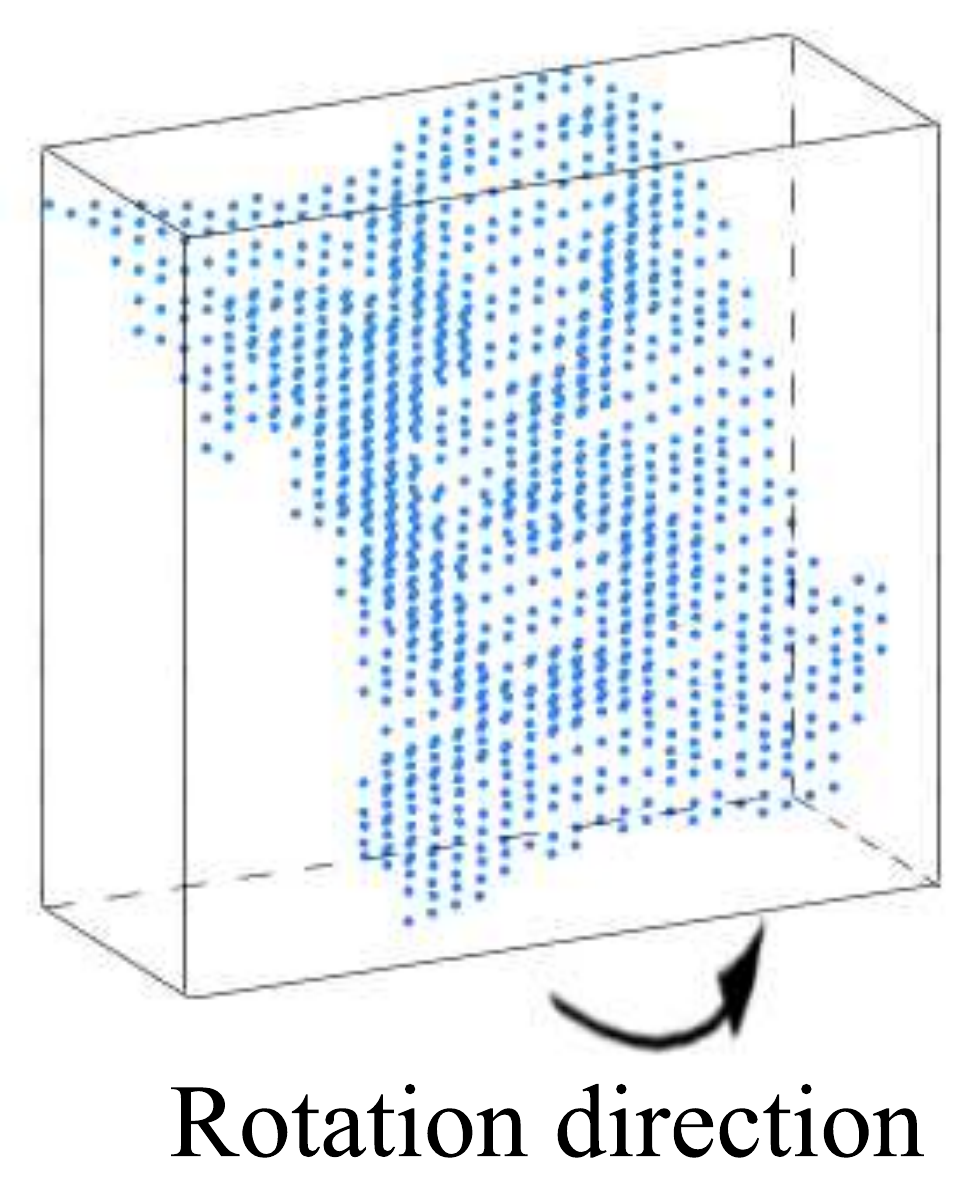}}
	\vspace{-0.05in}
	\caption{The structure matching for the source cube. (a) The original cube. (b) The translated cube. (c) The translated and rotated cube.}
	\label{fig:cubematch}
\end{figure}

We consider matching $ \vec{c}_s $ and $ \vec{c}_t $ using both translation and rotation, as shown in Fig.~\ref{fig:cubematch}. Firstly we translate $ \vec{c}_s $ by the difference in location between $ \vec{c}_s $ and $ \vec{c}_t $. The difference in each dimension, denoted as $ \vec{t}=(t_x,t_y,t_z) $, is computed as 
\begin{equation}
	t_x=
		\frac
			{\sum(\partial\mathbf{\Omega}\vec{c}_t-\partial\mathbf{\Omega}\vec{c}_s)_x}
			{|\partial\mathbf{\Omega}\vec{c}_t|},
	\label{eq:tx}
\end{equation}
where $ \partial{\Omega} $ is a $ M^3\times M^3 $ diagonal matrix, extracting the boundary within one hop to the missing region in the cube, and $ |\partial\mathbf{\Omega}\vec{c}_t| $ is the number of the boundary points. $ t_y $ and $ t_z $ are defined in the same way as (\ref{eq:tx}). Thus we obtain the $ \vec{t} $-translated cube $ \vec{c}^t_s $ as
\begin{equation}
	\vec{c}^t_s=\vec{c}_s+\mathbf{T},
	\label{eq:cubetrans}
\end{equation}
where $ \mathbf{T} $ is a $ n \times 3 $ translation matrix with $ \vec{t} $ repeated in each row.

Then we compute the rotation matrix for $ \vec{c}^t_s $ using the proposed simplified ICP method, which calculates the rotation matrix once and for all. The rotation matrix is obtained by the quaternion of the covariance matrix between two cubes. Firstly, we choose three pairs of points $ \{(\vec{u}_i,\vec{v}_i)\}_{i=1}^3 $ in $ \vec{c}_s $ and $ \vec{c}_t $ as the control points for registration. We randomly select three points in $ \vec{c}_t $ as $ \vec{v}_i $, and then find the closest point in the relative position for each of them in $ \vec{c}_t $ as $ \vec{u}_i $. Secondly, we compute the covariance matrix $ \mathbf{C} $ between $ \vec{u}_i $ and $ \vec{v}_i $ as
\begin{equation}
	\mathbf{C}=\sum_{i=1}^3\mathbf{G}_i\mathbf{G}^T_i,
	\label{eq:rotC}
\end{equation}
where $ \mathbf{G}_i $ is a $ 4 \times 4 $ matrix:
\begin{equation}
	\mathbf{G}_i={\left[
			\begin{array}{cc}
				0 & \vec{u}_i-\vec{v}_i\\
				(\vec{u}_i-\vec{v}_i)^T & \vec{u}_i+\vec{v}_i
			\end{array}
		\right]}.
	\label{eq:rotG}
\end{equation}

Then the unit eigenvector $ \vec{e}=[e_0,e_1,e_2,e_3] $ corresponding to the maximum eigenvalue of the matrix $ \mathbf{C} $ is adopted as the optimal rotation. Finally the rotation matrix $ \mathbf{R} $ corresponding to $ \vec{e} $ is computed as in~\cite{Besl02}:
\begin{equation}
\begin{smallmatrix}
\begin{scriptsize}
	\mathbf{R}={\left[
			\begin{array}{ccc}
			e^2_0+e^2_1-e^2_2-e^2_3	& 2(e_1e_2-e_0e_3)			& 2(e_1e_3+e_0e_2)\\
			2(e_1e_2+e_0e_3)		& e^2_0+e^2_2-e^2_1-e^2_3	& 2(e_2e_3-e_0e_1)\\
			2(e_1e_3-e_0e_2)		& 2(e_2e_3+e_0e_1)			& e^2_0+e^2_3-e^2_1-e^2_2
			\end{array}
		\right]}.
	\label{eq:rotmatrix}
\end{scriptsize}
\end{smallmatrix}
\end{equation}

Since $ \vec{c}^r_s,\vec{c}^t_s\in \mathbb{R}^{M^3 \times 3} $, we have $ (\vec{c}^r_s)^T=\mathbf{R}(\vec{c}^t_s)^T $. Hence, the rotated cube $ \vec{c}^r_s $ is finally computed by
\begin{equation}
	\vec{c}^r_s=\vec{c}^t_s\mathbf{R}^T.
	\label{eq:rot}
\end{equation}

Rotation might lead to points on irregular grids. In order to maintain the voxelized unity of the point cloud, we further voxelize $ \vec{c}^r_s $ as in Section~\ref{sec:voxelize}. Also, points outside the cube due to the rotation are removed. This leads to the final reference cube, denoted as $ \vec{c}^f_s $, which will be adopted in the following inpainting step. 

\subsection{Problem Formulation}
\label{subsec:Formulation}
Next, we cast the inpainting problem as an optimization problem, which is regularized with the graph-signal smoothness prior as mentioned in Section~\ref{subsec:knn}. This problem is formulated as
\begin{equation}
	\min_{\vec{c}_r}~~
		\norm{\overline{\mathbf{\Omega}}\vec{c}_r-\overline{\mathbf{\Omega}}\vec{c}_t}_2^2
		+\alpha\norm{\mathbf{\Omega}\vec{c}_r-\mathbf{\Omega}\vec{c}^f_s}_2^2
		+\beta\vec{c}_r^T\mathcal{L}\vec{c}_r, 
	\label{eq:opt}
\end{equation}
where $ \vec{c}_r\in \mathbb{R}^{M^3 \times 3} $ is the desired cube. $ \mathbf{\Omega} $ is a $ M^3\times M^3 $ diagonal matrix, extracting the missing region in $ \vec{c}_t $ and $ \vec{c}^f_s $, while $ \overline{\mathbf{\Omega}} $ is a $ M^3\times M^3 $ diagonal matrix, extracting the known region. $\alpha$ and $\beta$ are two weighting parameters (we empirically set $\alpha=0.1$ and $\beta=10$ in the experiments). Besides, $\mathcal{L}$ is the graph Laplacian matrix of $ \vec{c}^f_s $, which is computed from a $ K $-NN graph we construct in $ \vec{c}^f_s $ in the same way as in (\ref{eq:graph}).

The first term in (\ref{eq:opt}) is a fidelity term, which ensures the desired cube to be close to $ \vec{c}_t $ in the known region. The second term constraints the unknown region of $ \vec{c}_r $ to be similar to that of $ \vec{c}^f_s $. Further, the third term is the graph-signal smoothness prior, which enforces the structure of $ \vec{c}_r $ to be smooth with respect to the graph constructed over $ \vec{c}^f_s $. In other words, the third term aims to make the structure of $ \vec{c}_r $ mimic that of $ \vec{c}^f_s $. Besides, it also has the functionality of denoising the cube because the energy of the cube is compacted in low-frequency components as analyzed in Section~\ref{subsec:prior}.

(\ref{eq:opt}) is a quadratic programming problem. Taking derivative of (\ref{eq:opt}) with respect to $ \vec{c}_r $, we have the closed-form solution: 
\begin{equation}
	\hat{\vec{c}_r}=
	(\overline{\mathbf{\Omega}}^T\overline{\mathbf{\Omega}}+\alpha\mathbf{\Omega}^T\mathbf{\Omega}+\beta\mathcal{L})^{-1}
	(\overline{\mathbf{\Omega}}^T\overline{\mathbf{\Omega}}\vec{c}_t+\alpha\mathbf{\Omega}^T\mathbf{\Omega}\vec{c}^f_s).
\label{eq:solution}
\end{equation}

(\ref{eq:opt}) is thus solved optimally and efficiently. Finally, we replace the target cube with the resulting cube, which serves as the output.

%% file: results.tex
\subsection{Experimental Setup}
We evaluate the proposed method by testing on several point cloud datasets from Microsoft~\cite{Cai16}, Mitsubishi~\cite{Cohen17}, Stanford~\cite{Stanford}, Yobi~\cite{Yobi}, JPEG Pleno~\cite{Pleno} and Visual Computing Lab~\cite{VCL}. We test on two types of holes: 1) real holes generated during the capturing process, which have no ground truth; 2) synthetic holes on point clouds so as to compare with the ground truth. In particular, the number of nearest neighbors $ K $ is considered to be related to $ m $, the number of existing points in the cube. Empirically, $ K =\sqrt{m} $ in our experiments.

Further, we compare our method with four competing algorithms for 3D geometry inpainting, including Meshfix~\cite{Meshfix}, Meshlab~\cite{Meshlab},~\cite{Wang07} and~\cite{Lozes14}. Note that Meshfix and Meshlab are based on meshes, so we convert point clouds to meshes via Meshlab prior to testing the algorithms, and then convert the inpainted meshes back to point clouds as the final output.

\subsection{Experimental Results}
It is nontrivial to measure the geometry difference of point clouds objectively. We apply the geometric distortion metrics in~\cite{Tian17} and~\cite{Dinesh17}, referred to as GPSNR and NSHD respectively, as the metric for evaluation.

\begin{figure*}
	\centering
	\subfigure[Original]{
		\begin{minipage}[b]{0.148\textwidth}
			\includegraphics[width=\textwidth]{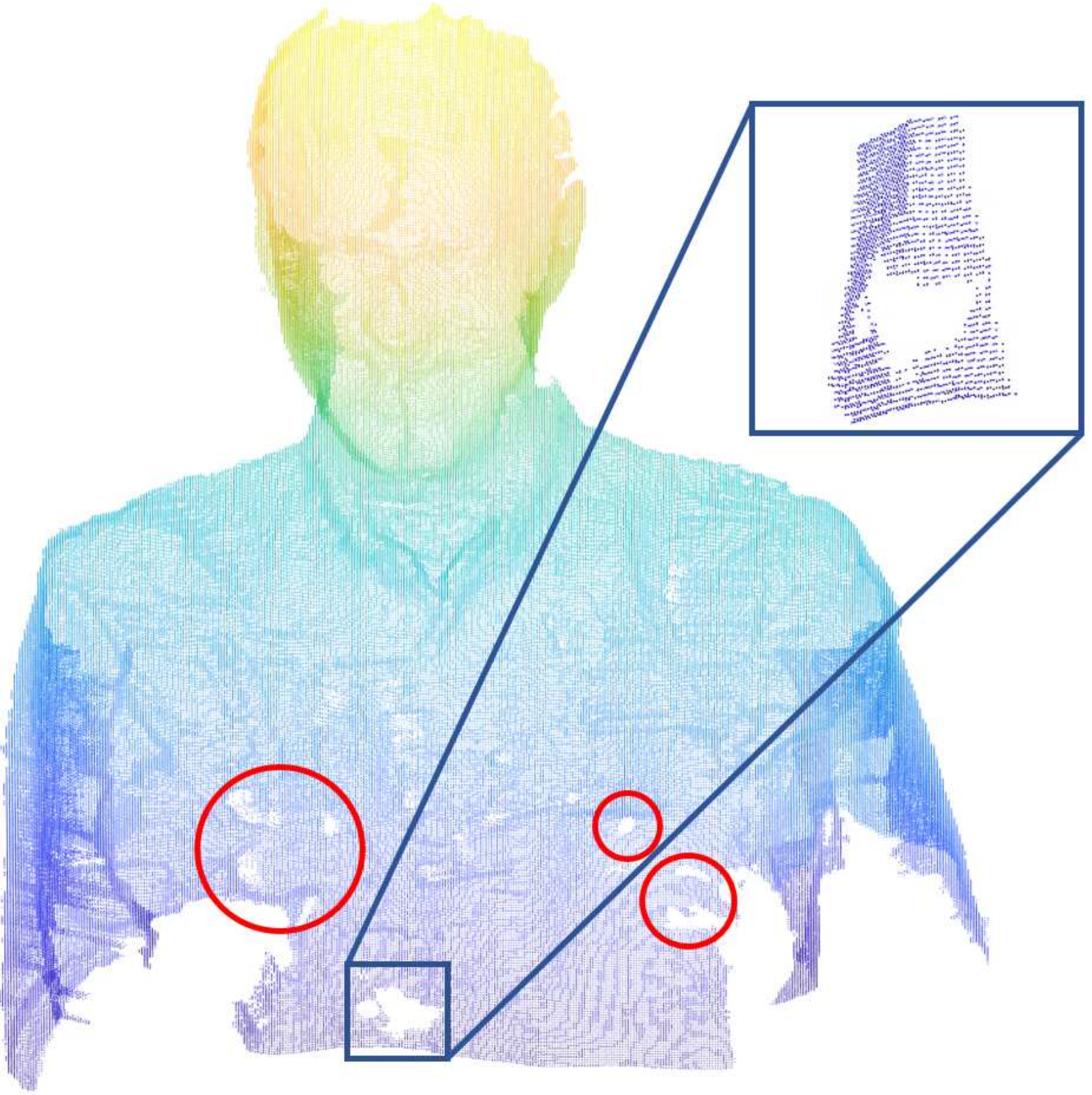} \\
			\vspace{0.01in}
			\includegraphics[width=\textwidth]{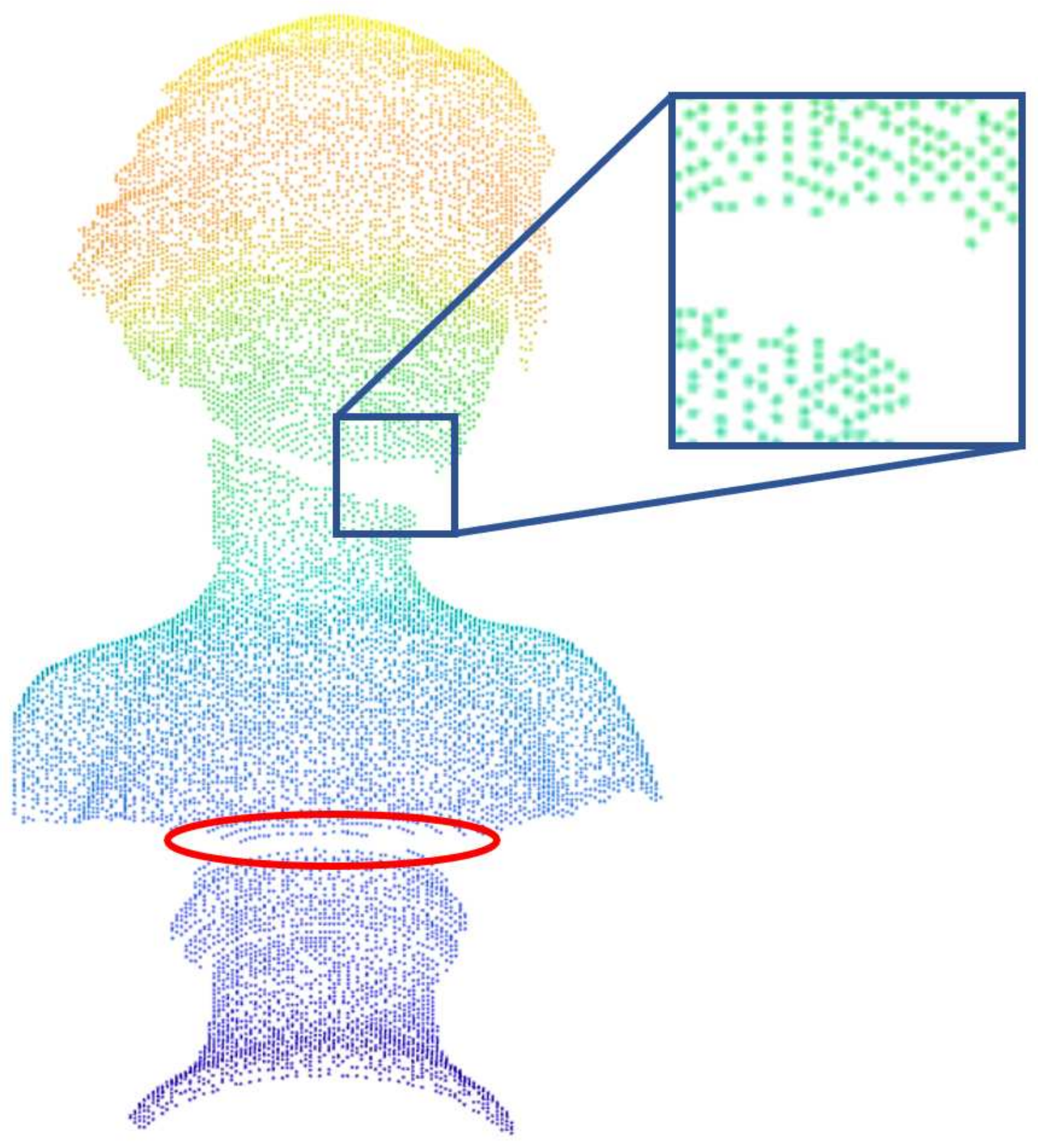} \\
			\vspace{0.01in}
			\includegraphics[width=\textwidth]{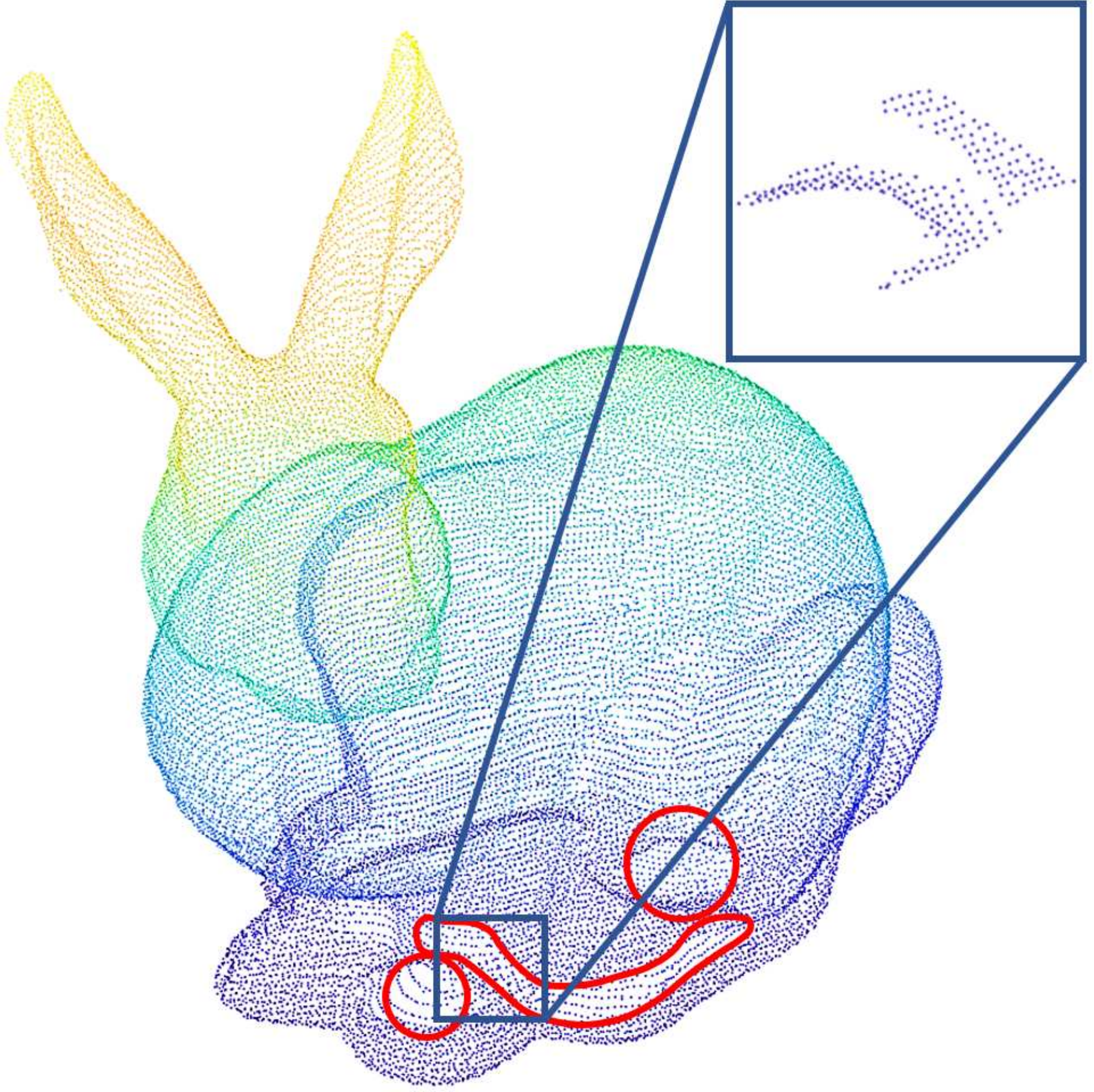}
		\end{minipage}
	}
	\subfigure[Meshfix]{
		\begin{minipage}[b]{0.148\textwidth}
			\includegraphics[width=\textwidth]{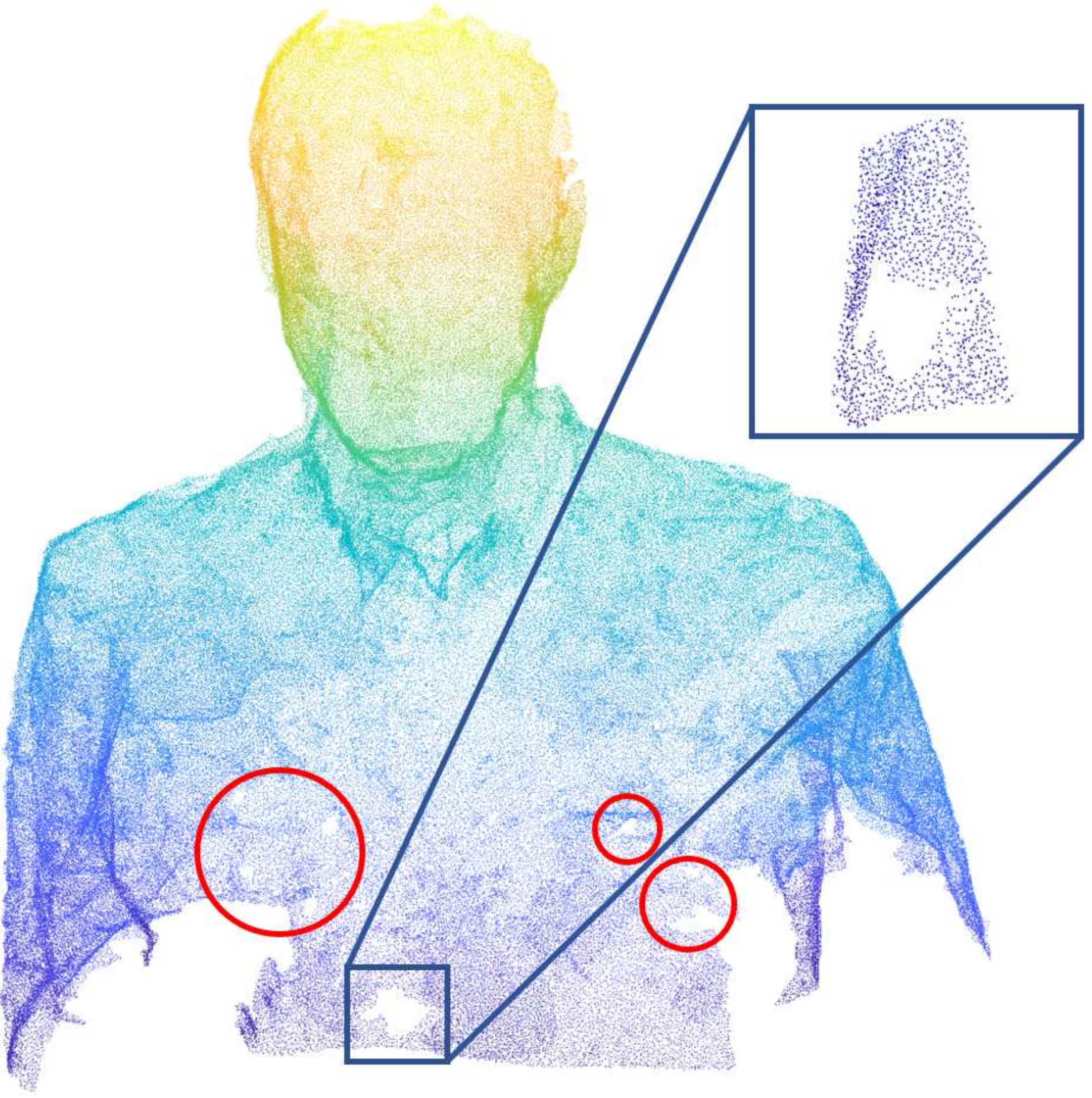} \\
			\vspace{0.01in}
			\includegraphics[width=\textwidth]{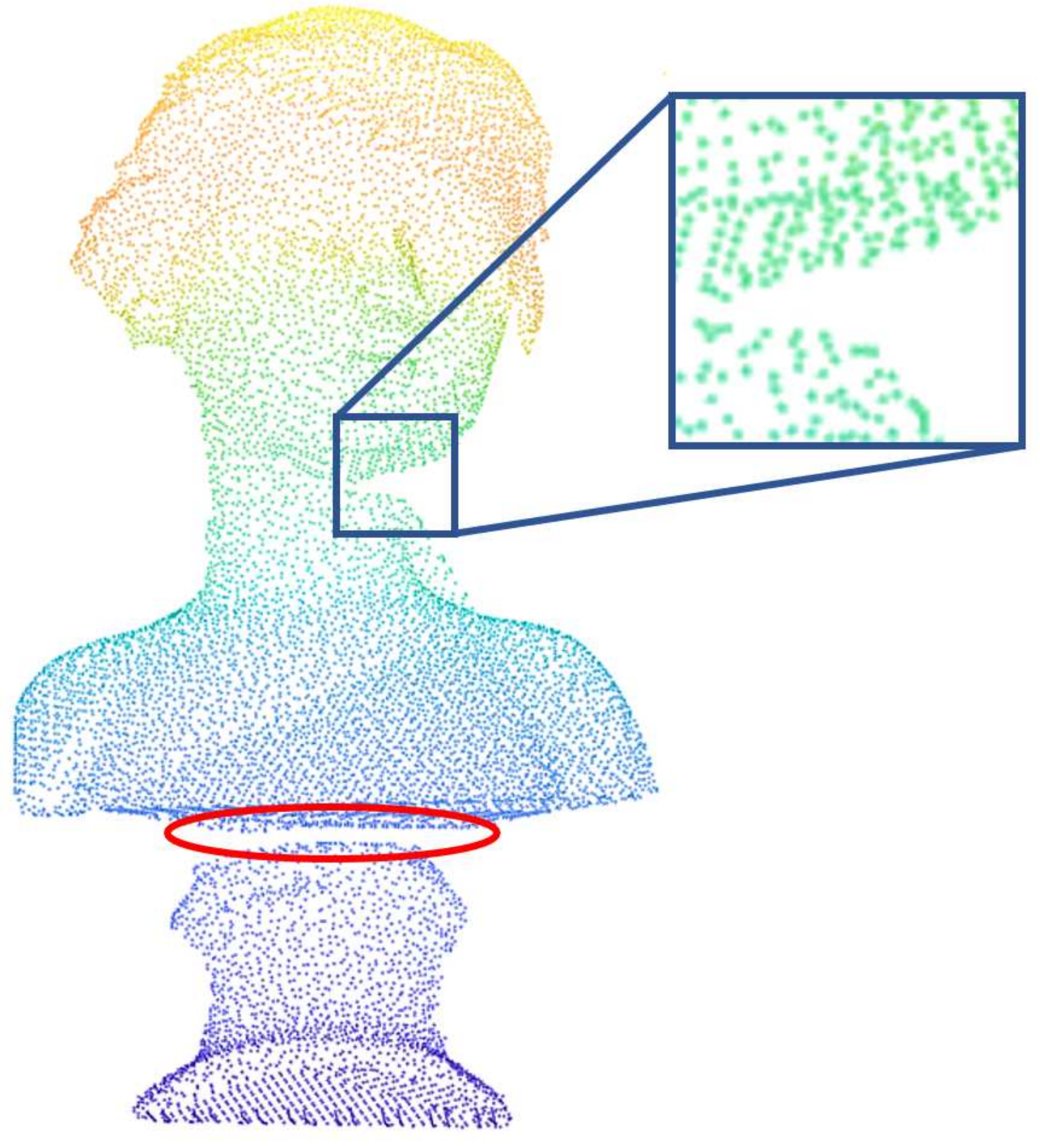} \\
			\vspace{0.01in}
			\includegraphics[width=\textwidth]{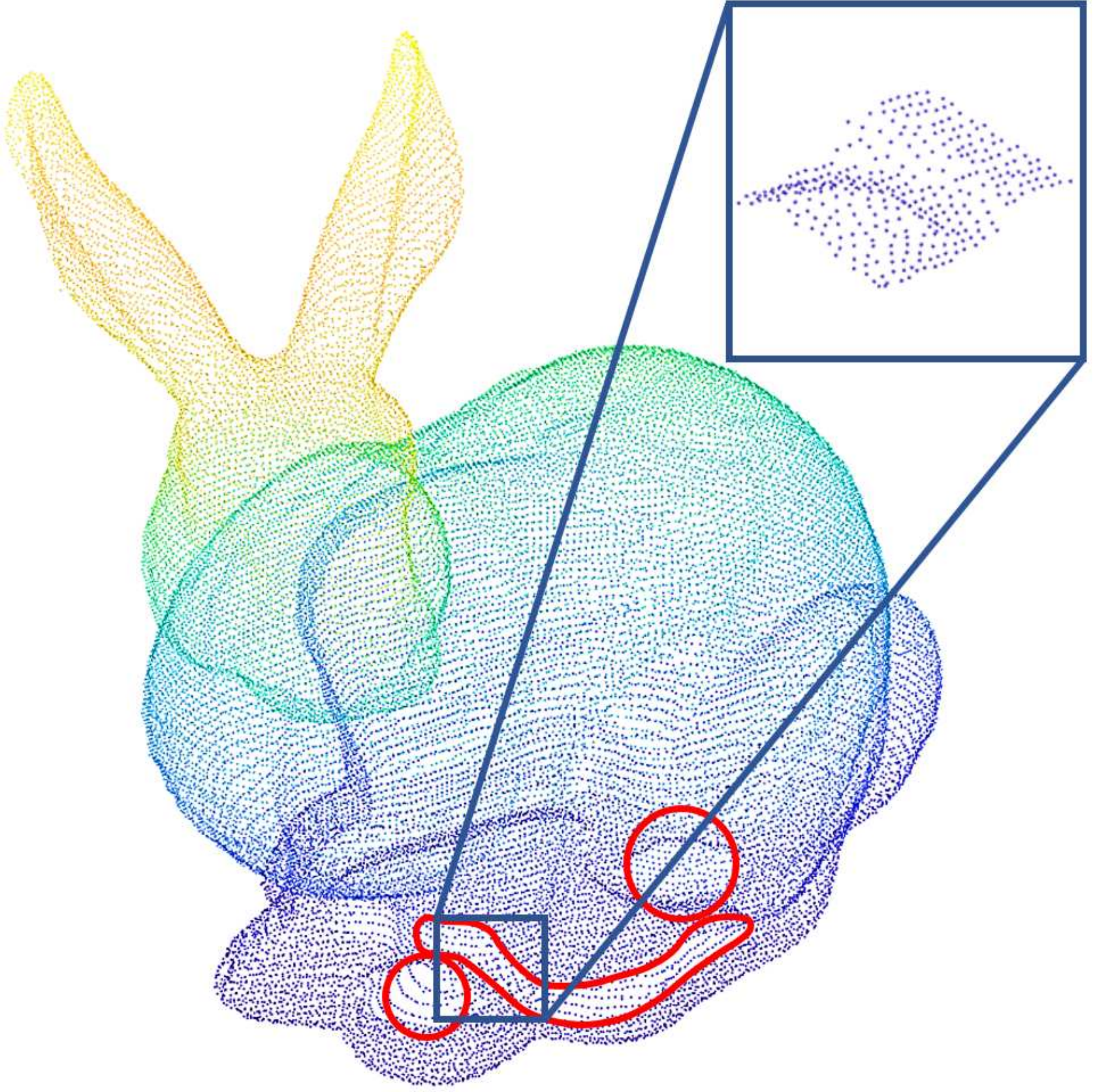}
		\end{minipage}
	}
	\subfigure[Meshlab]{
		\begin{minipage}[b]{0.148\textwidth}
			\includegraphics[width=\textwidth]{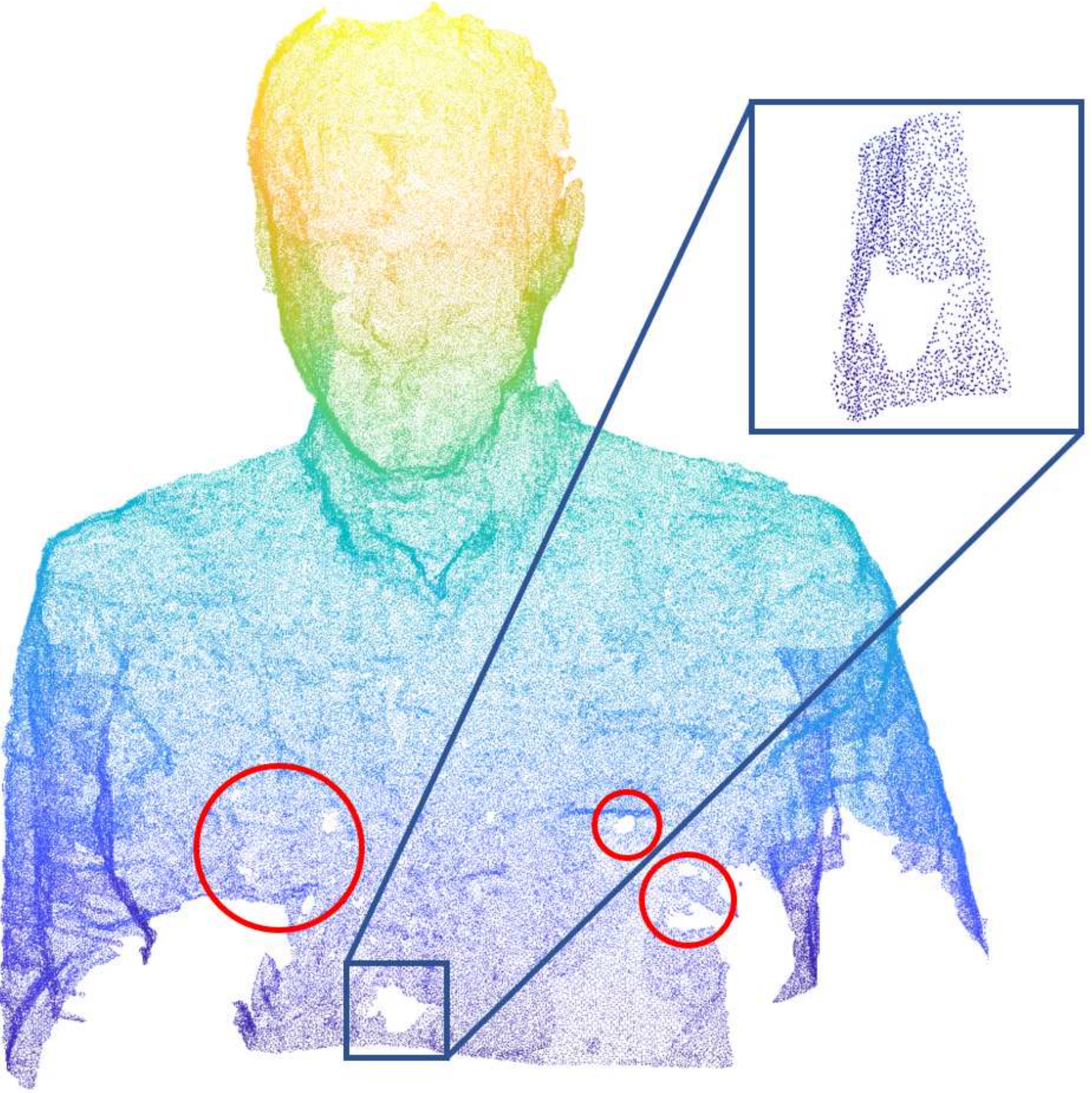} \\
			\vspace{0.01in}
			\includegraphics[width=\textwidth]{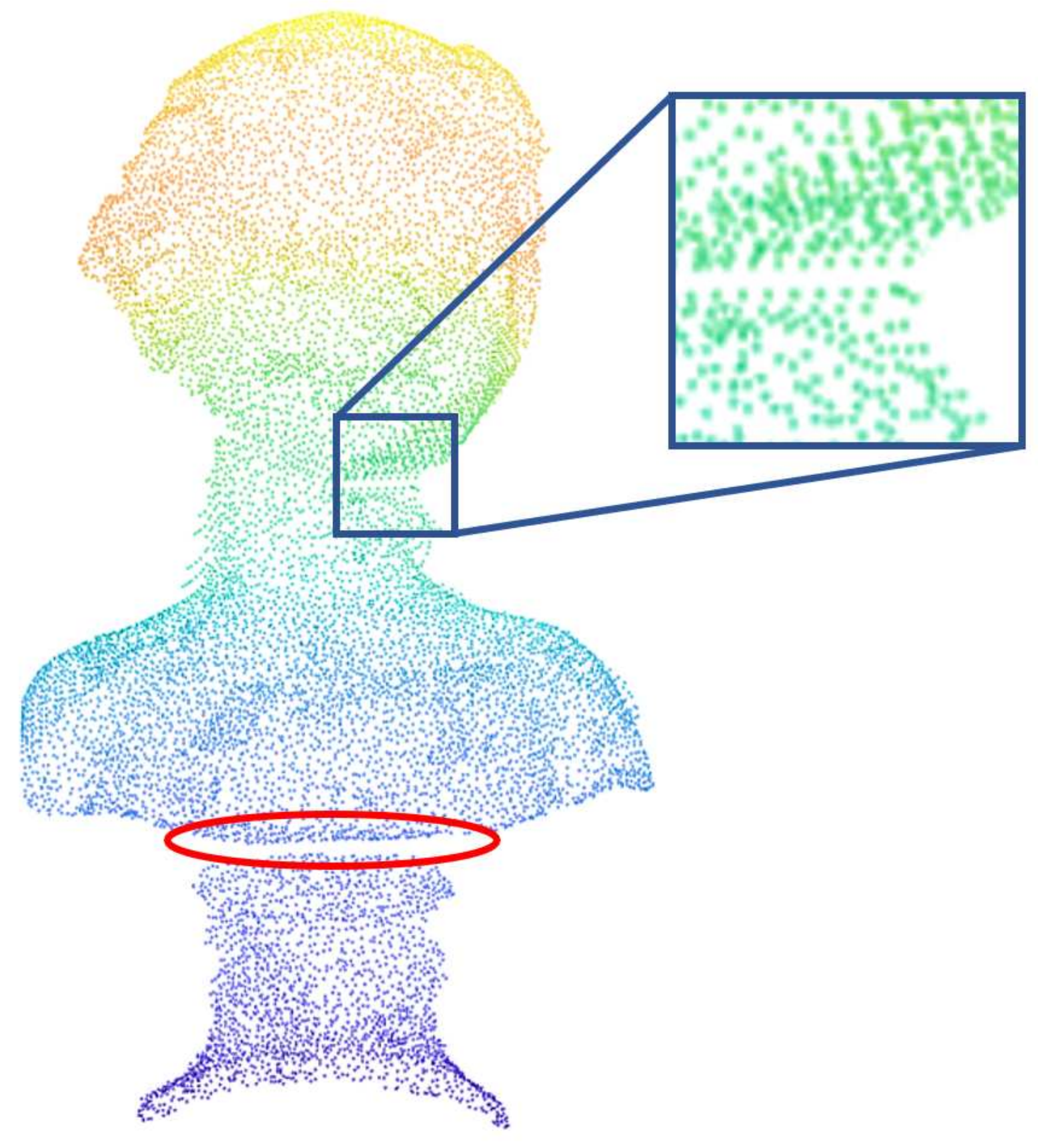} \\
			\vspace{0.01in}
			\includegraphics[width=\textwidth]{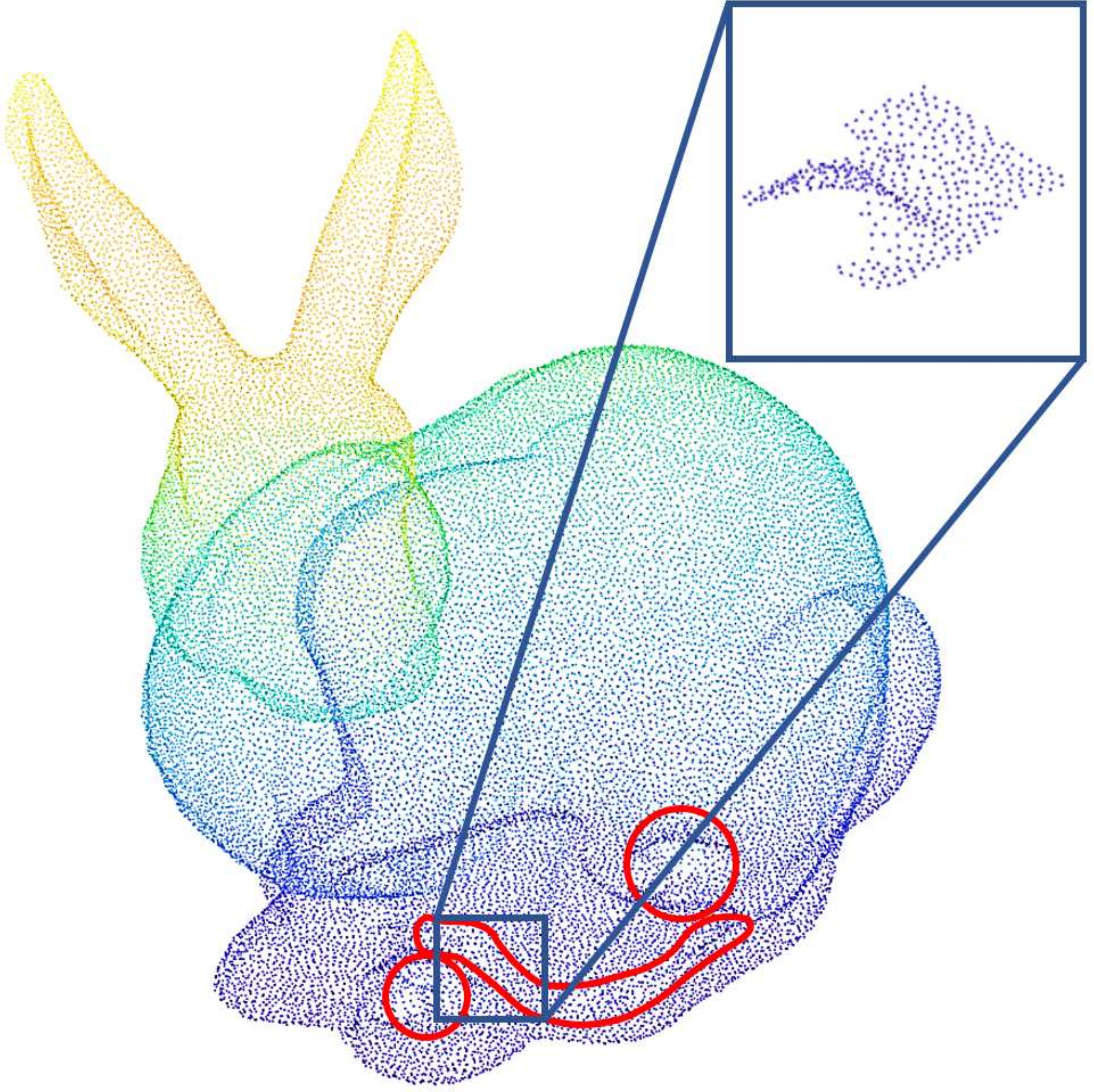}
		\end{minipage}
	}
	\subfigure[\cite{Wang07}]{
		\begin{minipage}[b]{0.148\textwidth}
			\includegraphics[width=\textwidth]{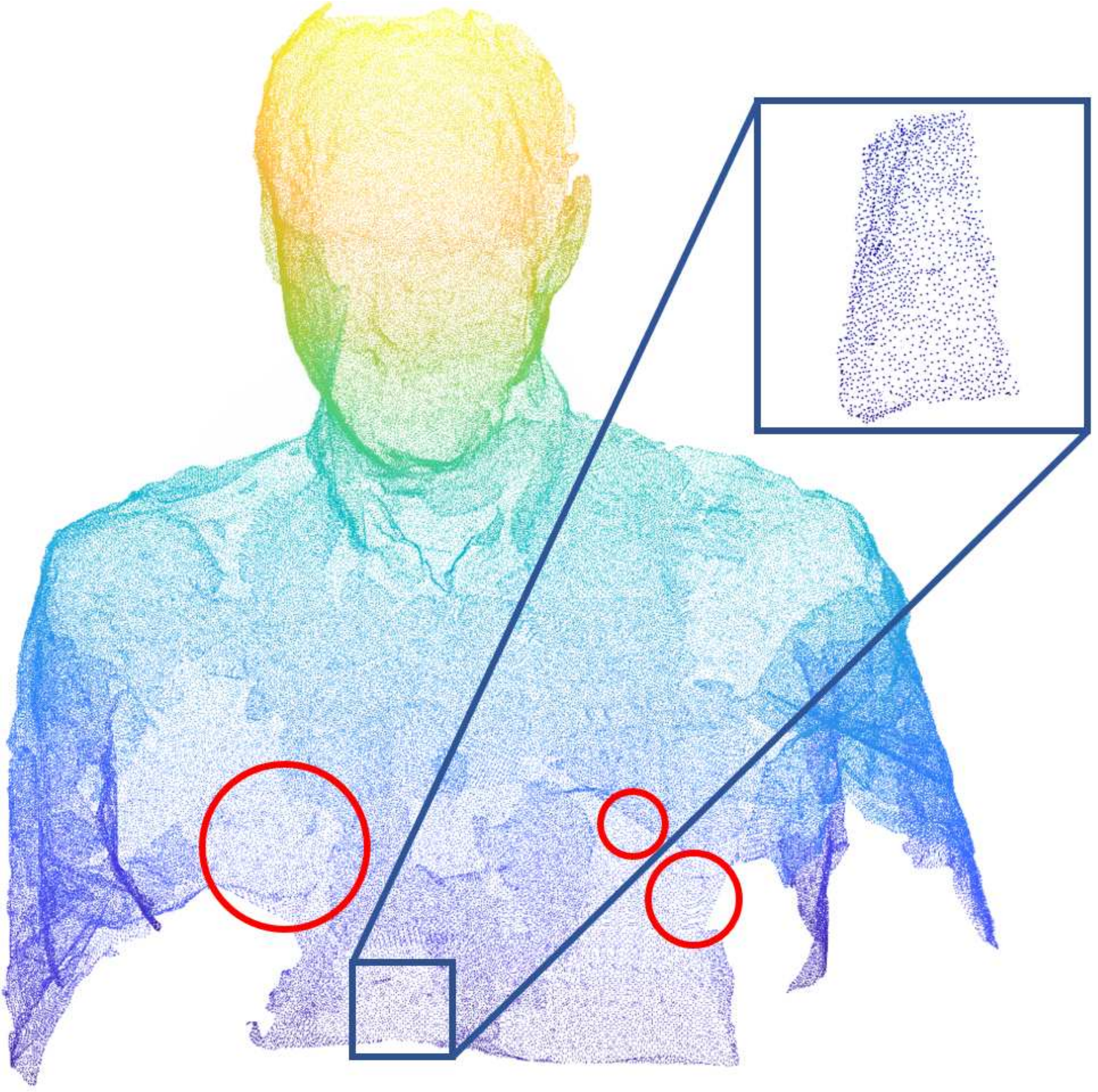} \\
			\vspace{0.01in}
			\includegraphics[width=\textwidth]{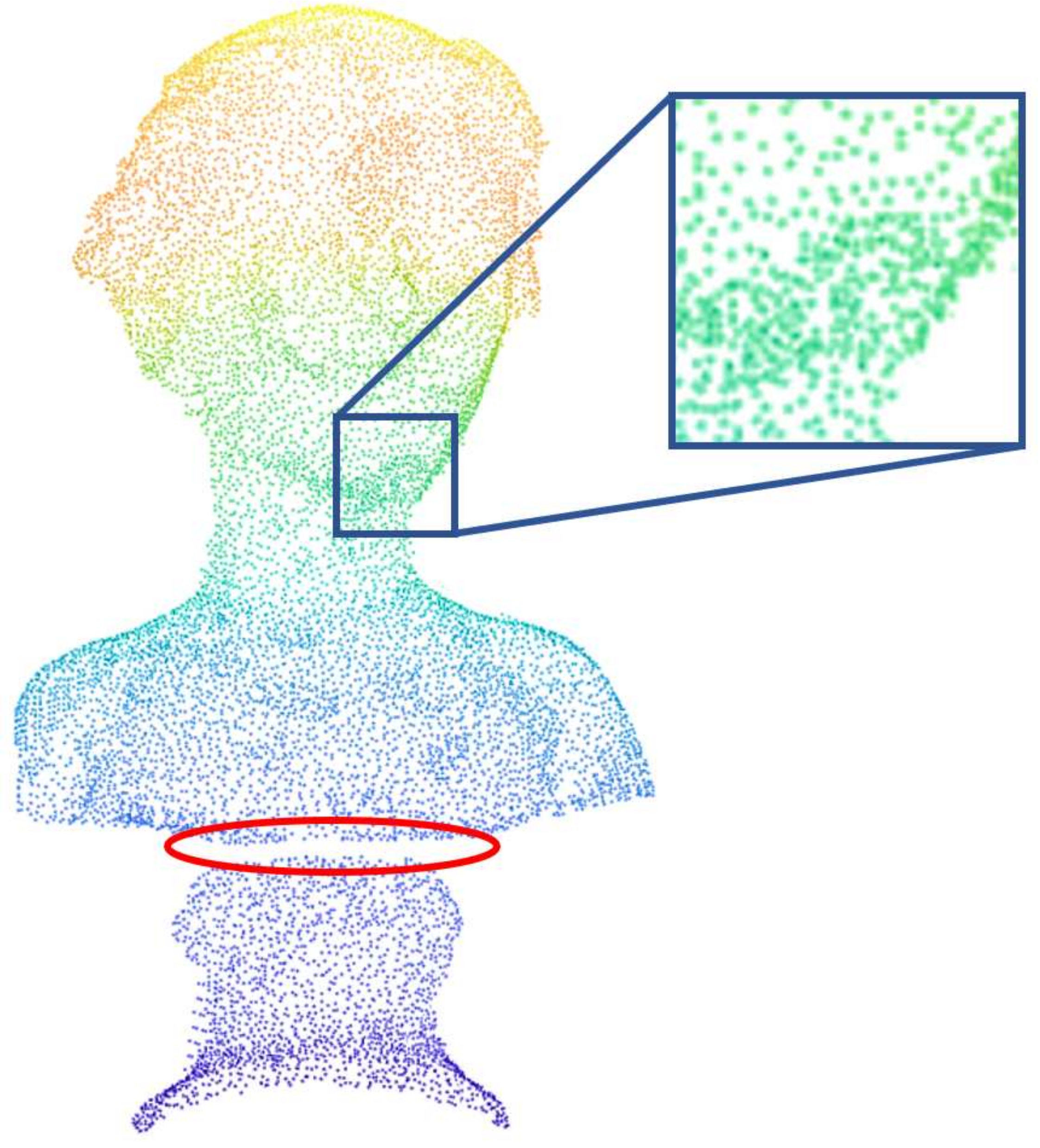} \\
			\vspace{0.01in}
			\includegraphics[width=\textwidth]{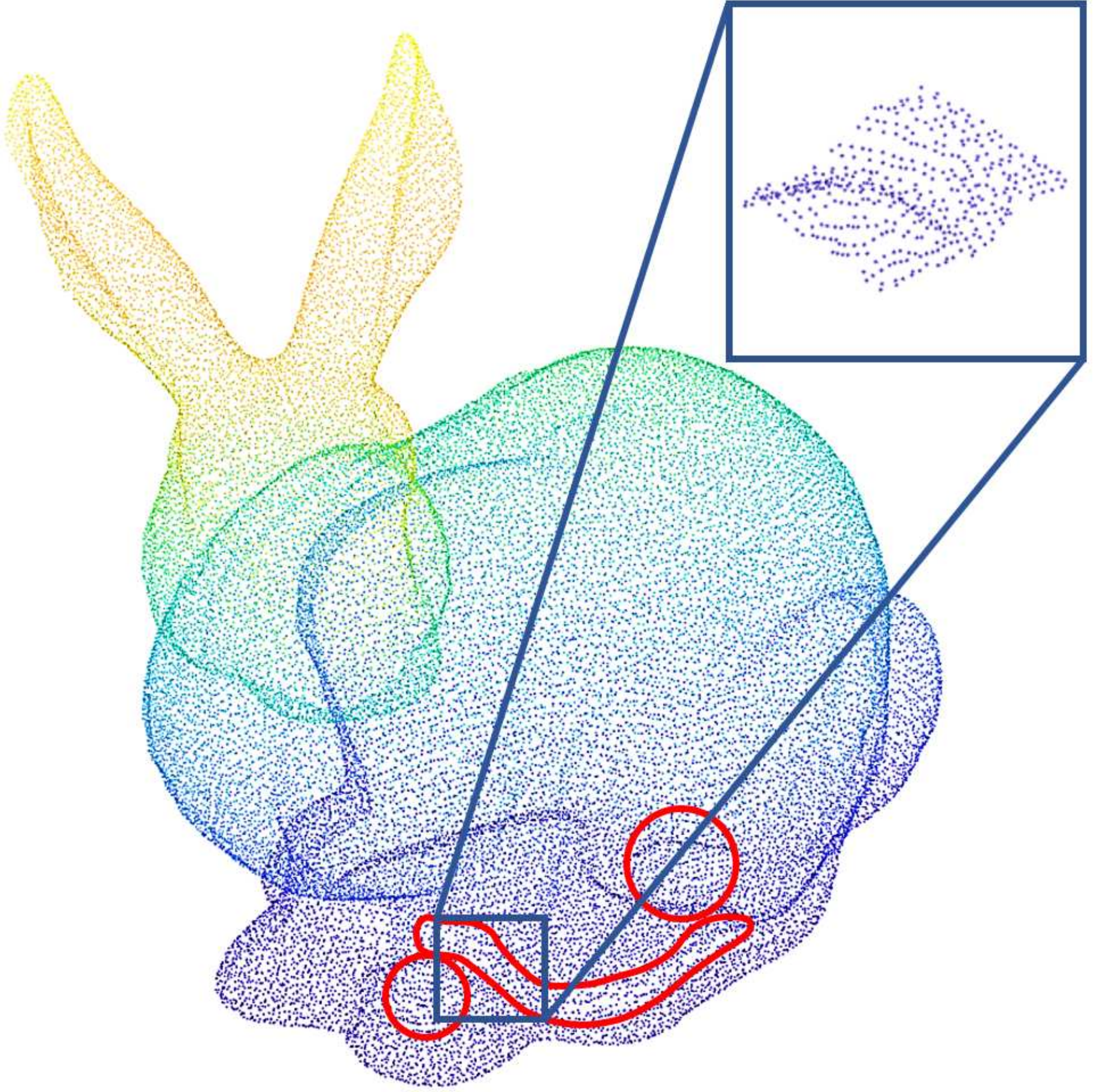}
		\end{minipage}
	}
	\subfigure[\cite{Lozes14}]{
		\begin{minipage}[b]{0.148\textwidth}
			\includegraphics[width=\textwidth]{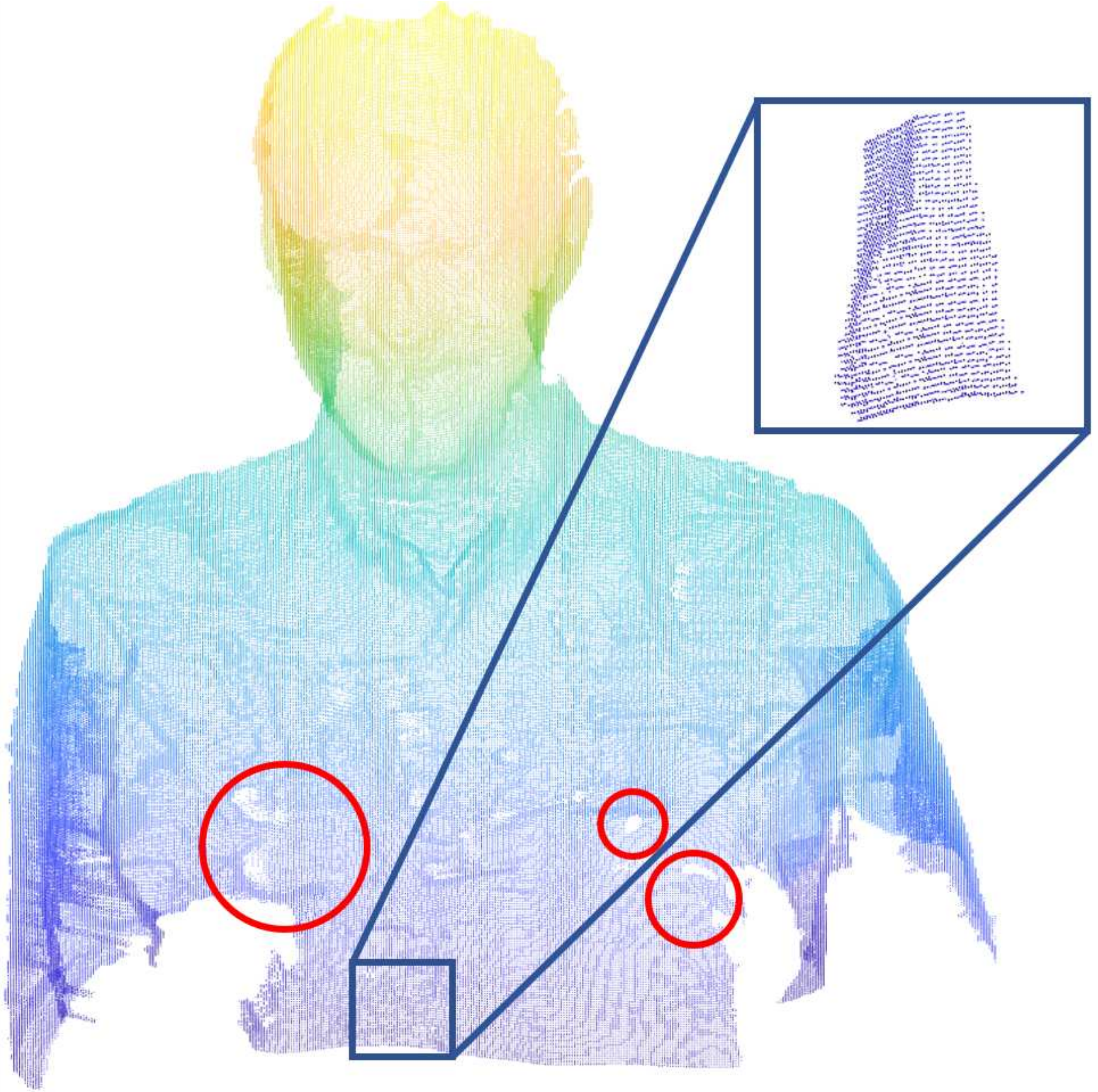} \\
			\vspace{0.01in}
			\includegraphics[width=\textwidth]{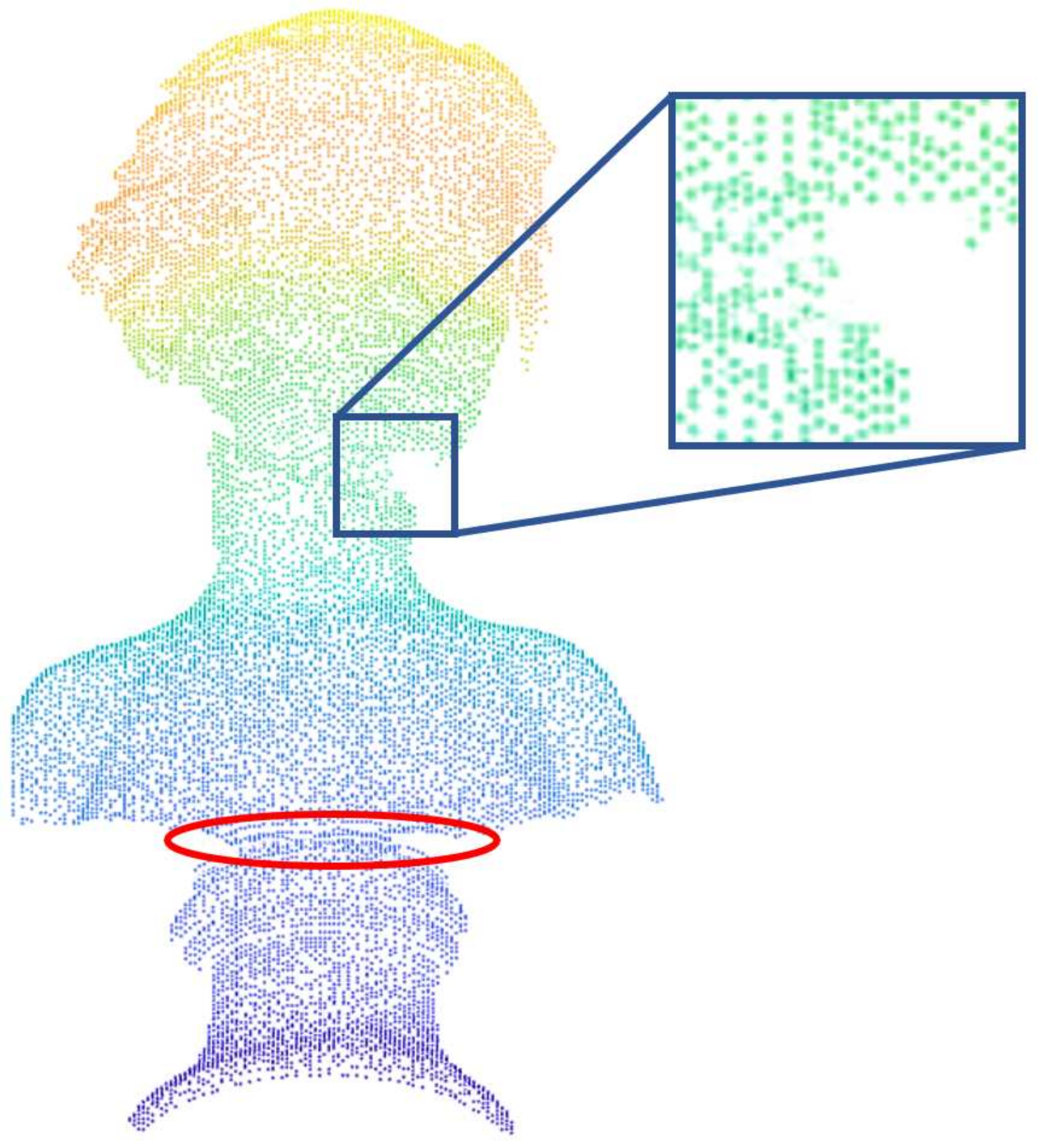} \\
			\vspace{0.01in}
			\includegraphics[width=\textwidth]{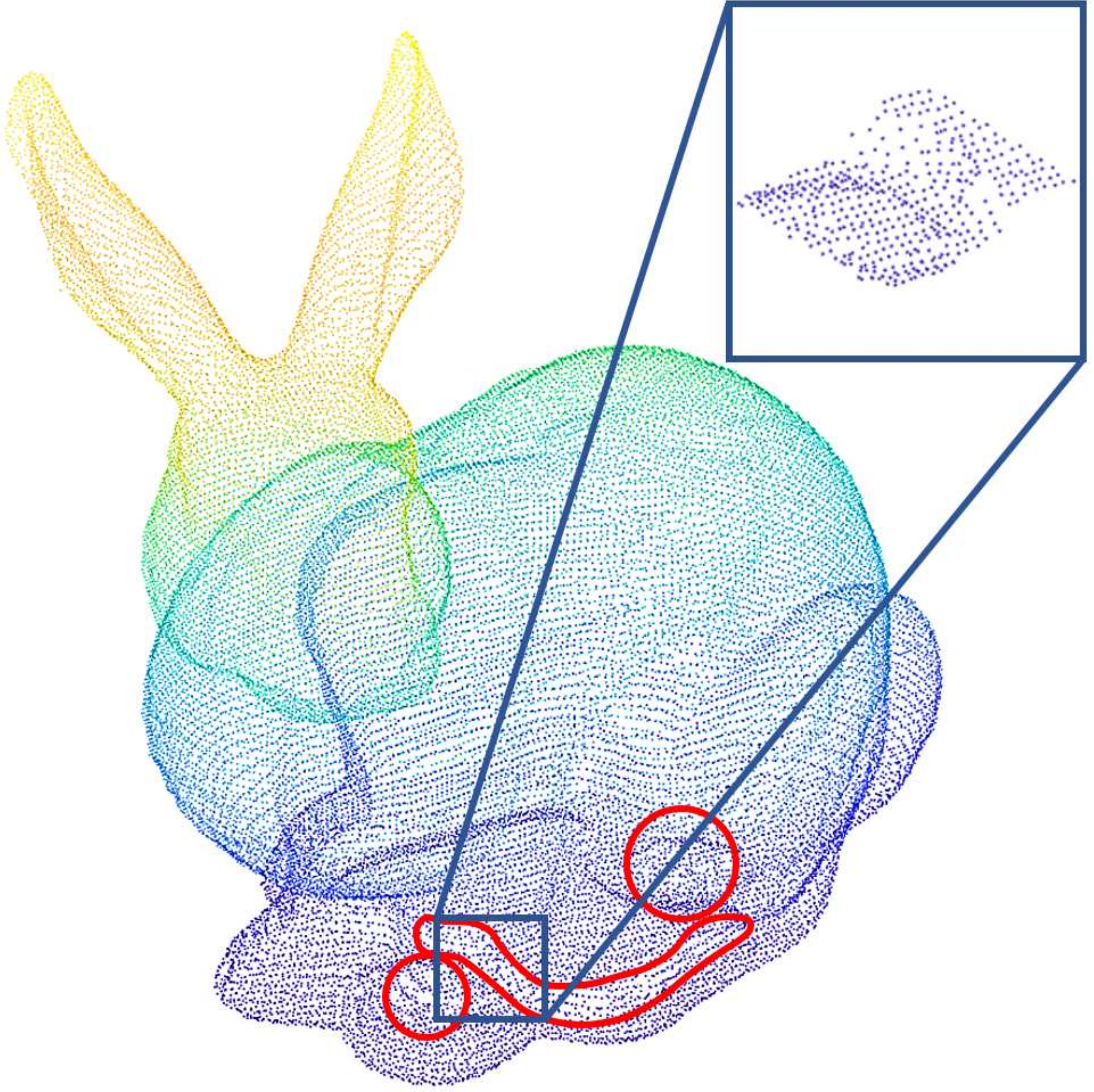}
		\end{minipage}
	}
	\subfigure[Proposed]{
		\begin{minipage}[b]{0.148\textwidth}
			\includegraphics[width=\textwidth]{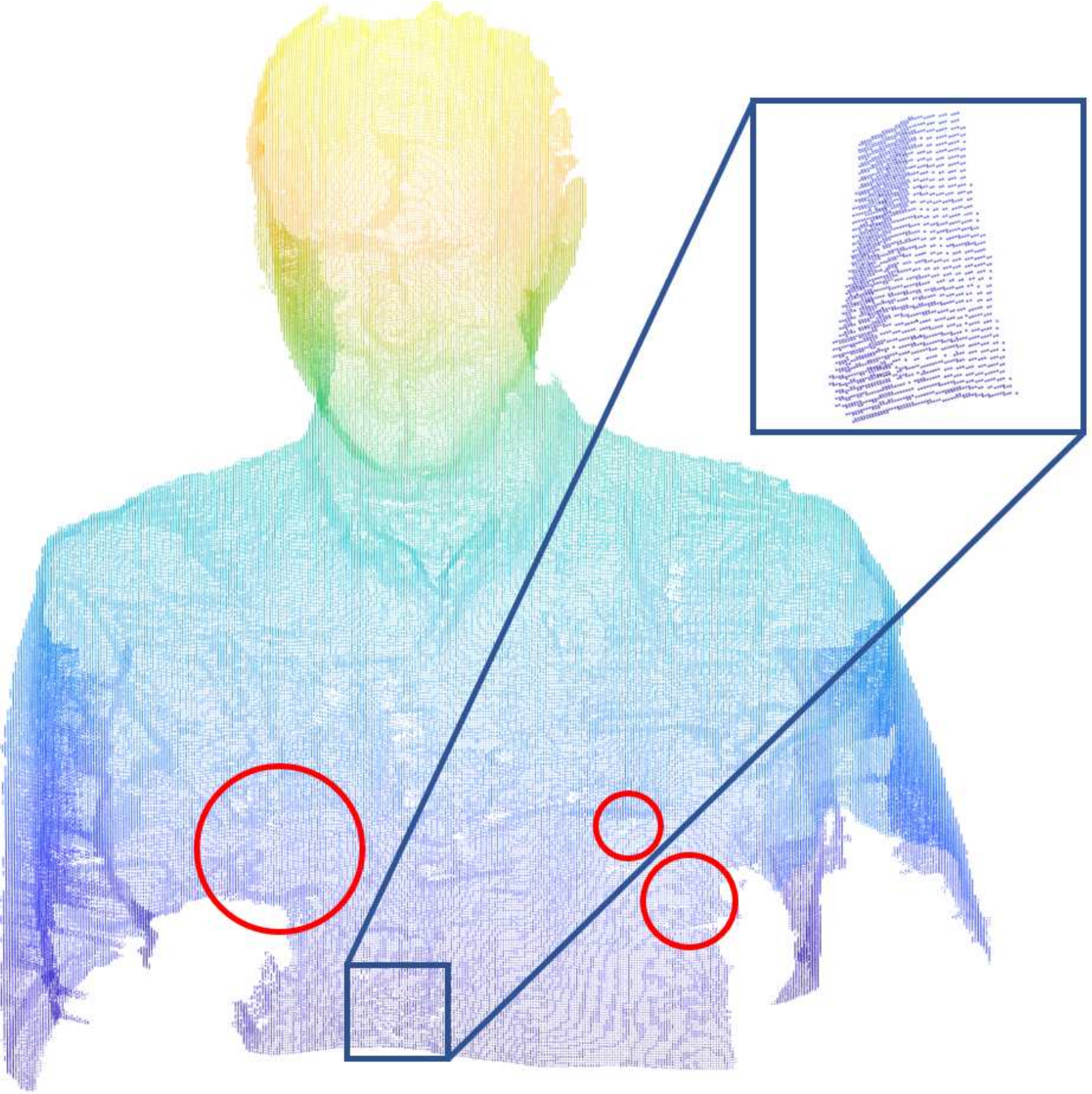} \\
			\vspace{0.01in}
			\includegraphics[width=\textwidth]{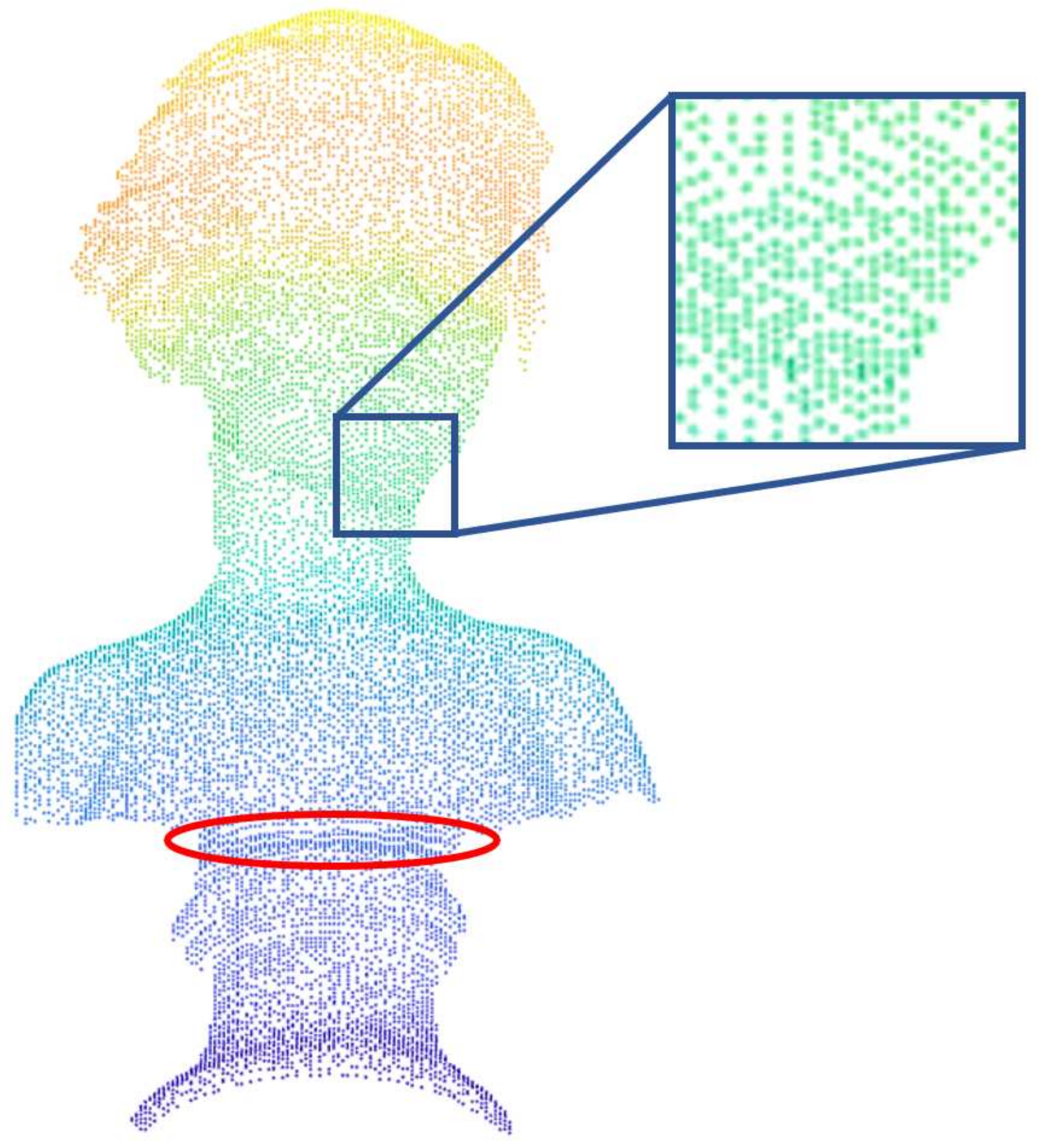} \\
			\vspace{0.01in}
			\includegraphics[width=\textwidth]{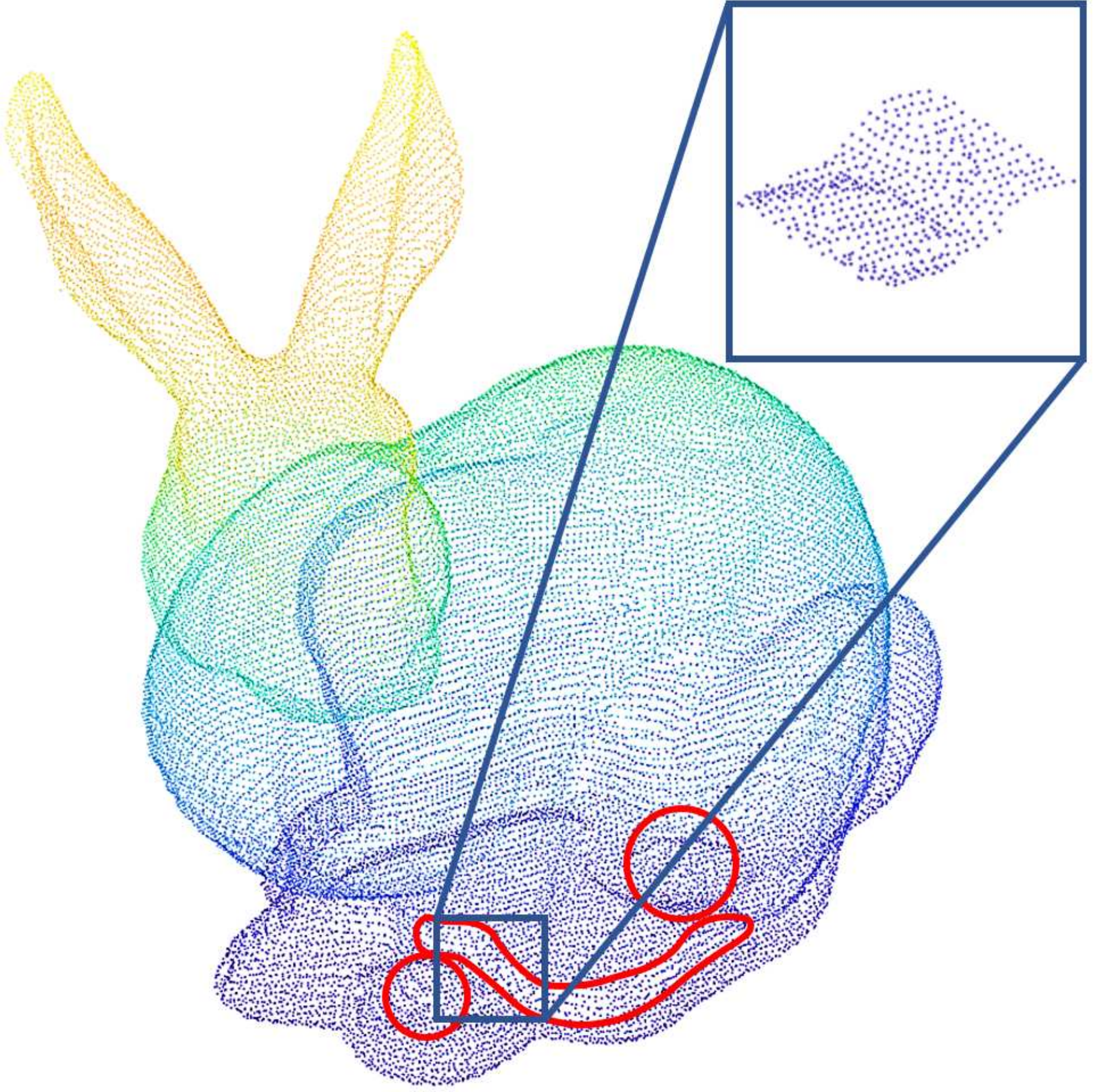}
		\end{minipage}
	}
	\caption{Inpainting from different methods for \textit{Phili} (row 1), \textit{Bimba} (row 2), \textit{Bunny} (row 3) with real holes marked in red circles, with one representative cube magnified.}
	\label{fig:result1}
\end{figure*}

\begin{figure*}
	\centering
	\subfigure[Hole]{
		\begin{minipage}[b]{0.148\textwidth}
			\includegraphics[width=\textwidth]{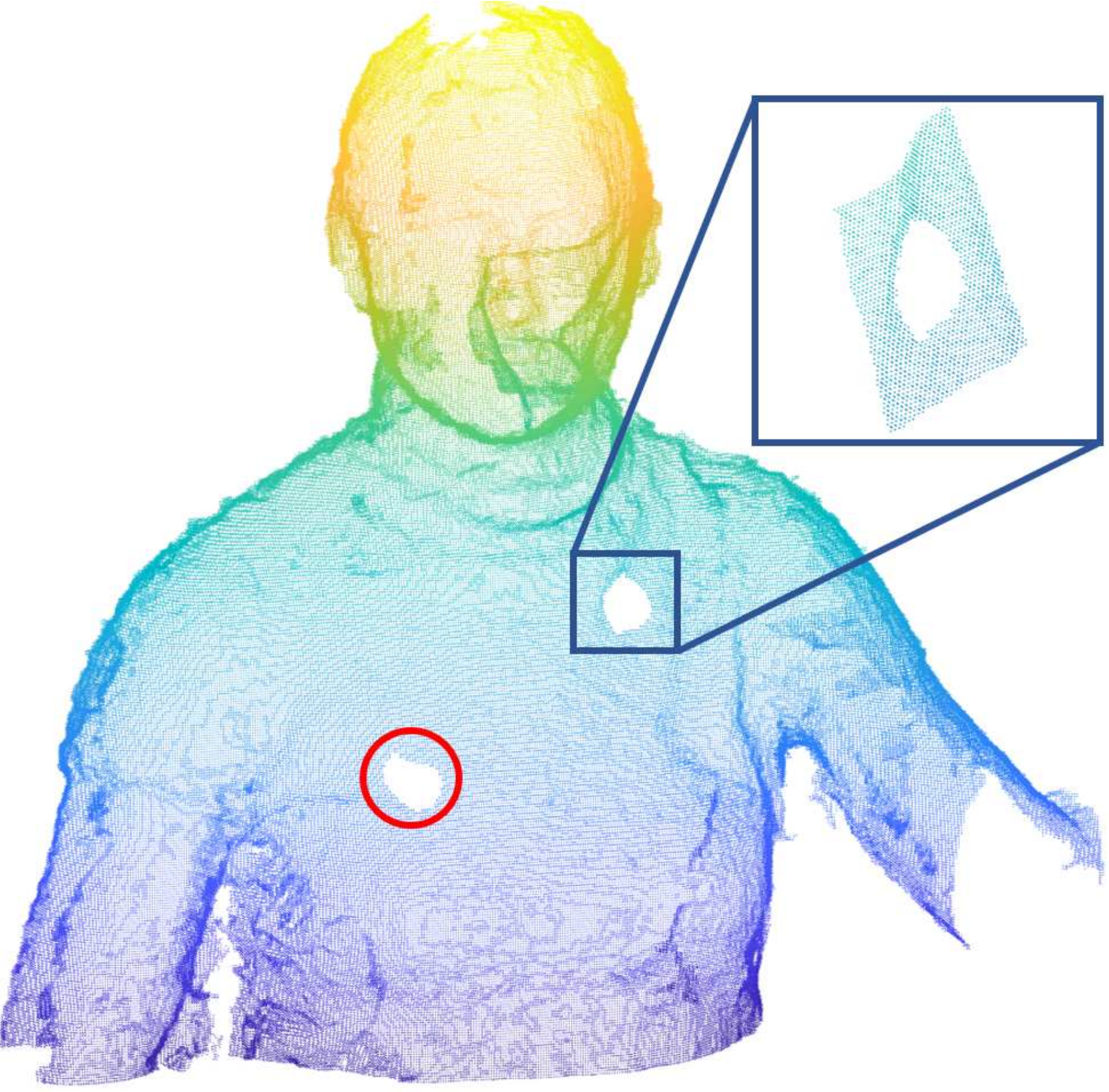} \\
			\vspace{0.01in}
			\includegraphics[width=\textwidth]{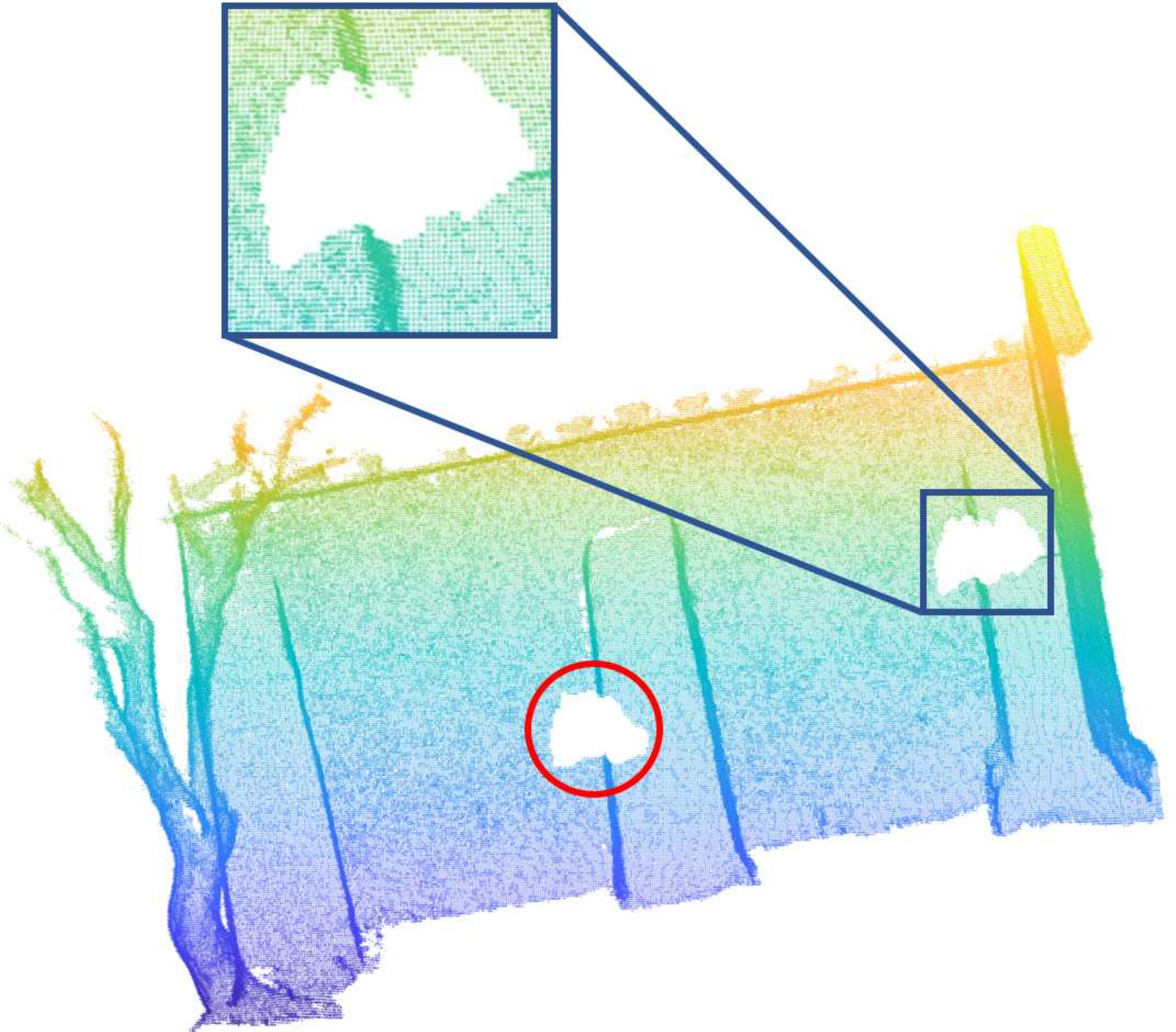}
		\end{minipage}
	}
	\subfigure[Meshlab]{
		\begin{minipage}[b]{0.148\textwidth}
			\includegraphics[width=\textwidth]{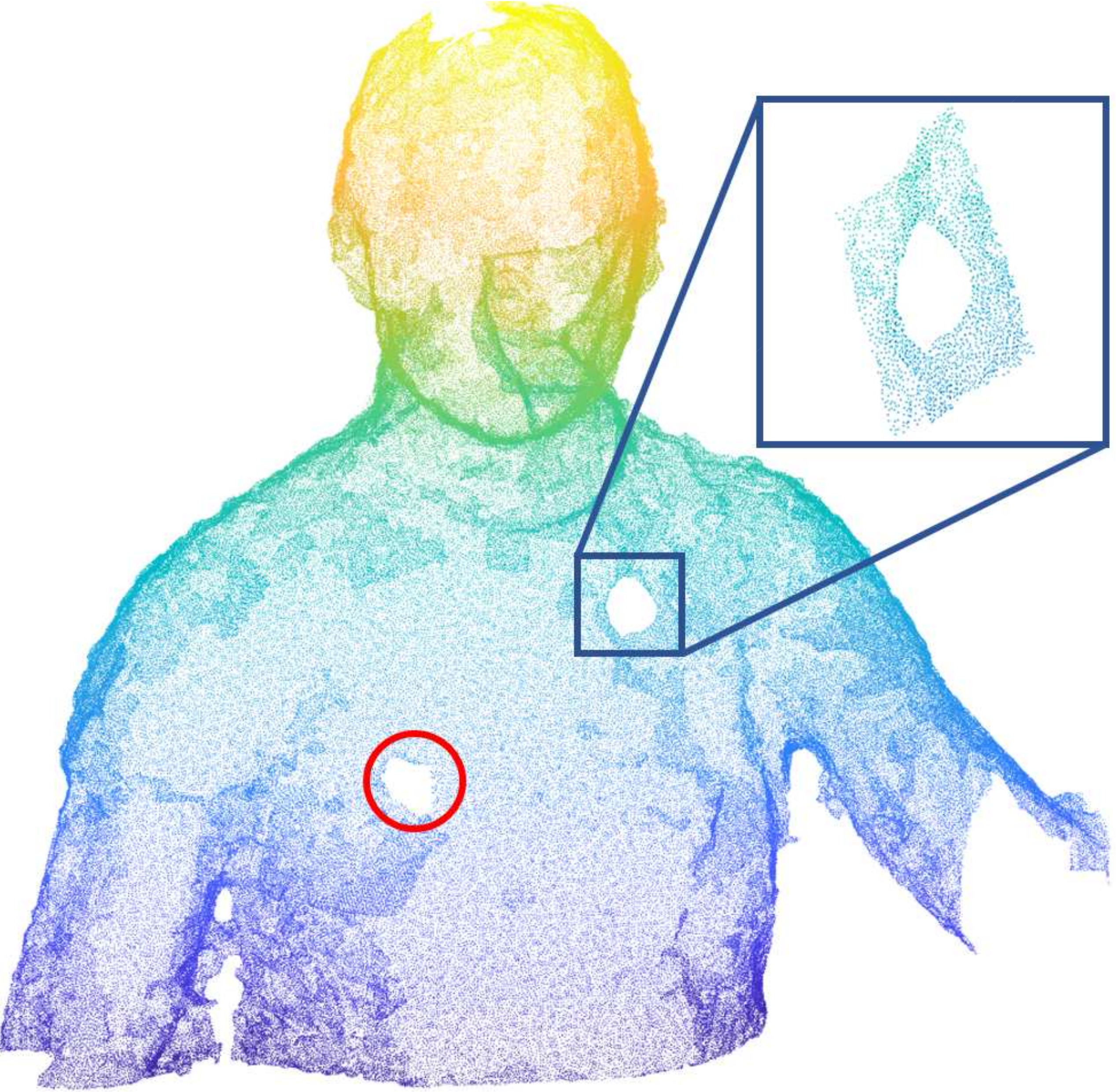} \\
			\vspace{0.01in}
			\includegraphics[width=\textwidth]{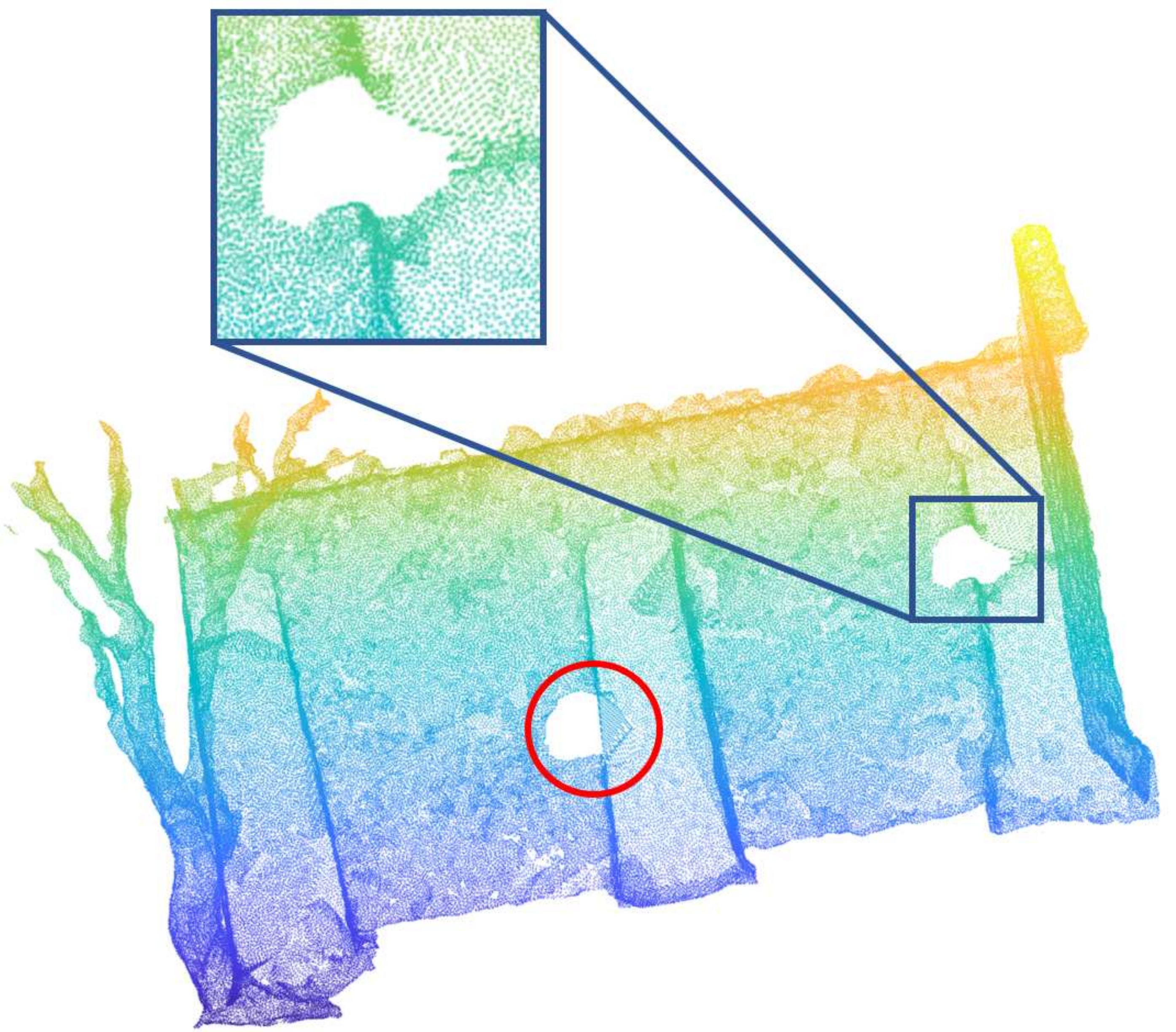}
		\end{minipage}
	}
	\subfigure[\cite{Wang07}]{
		\begin{minipage}[b]{0.148\textwidth}
			\includegraphics[width=\textwidth]{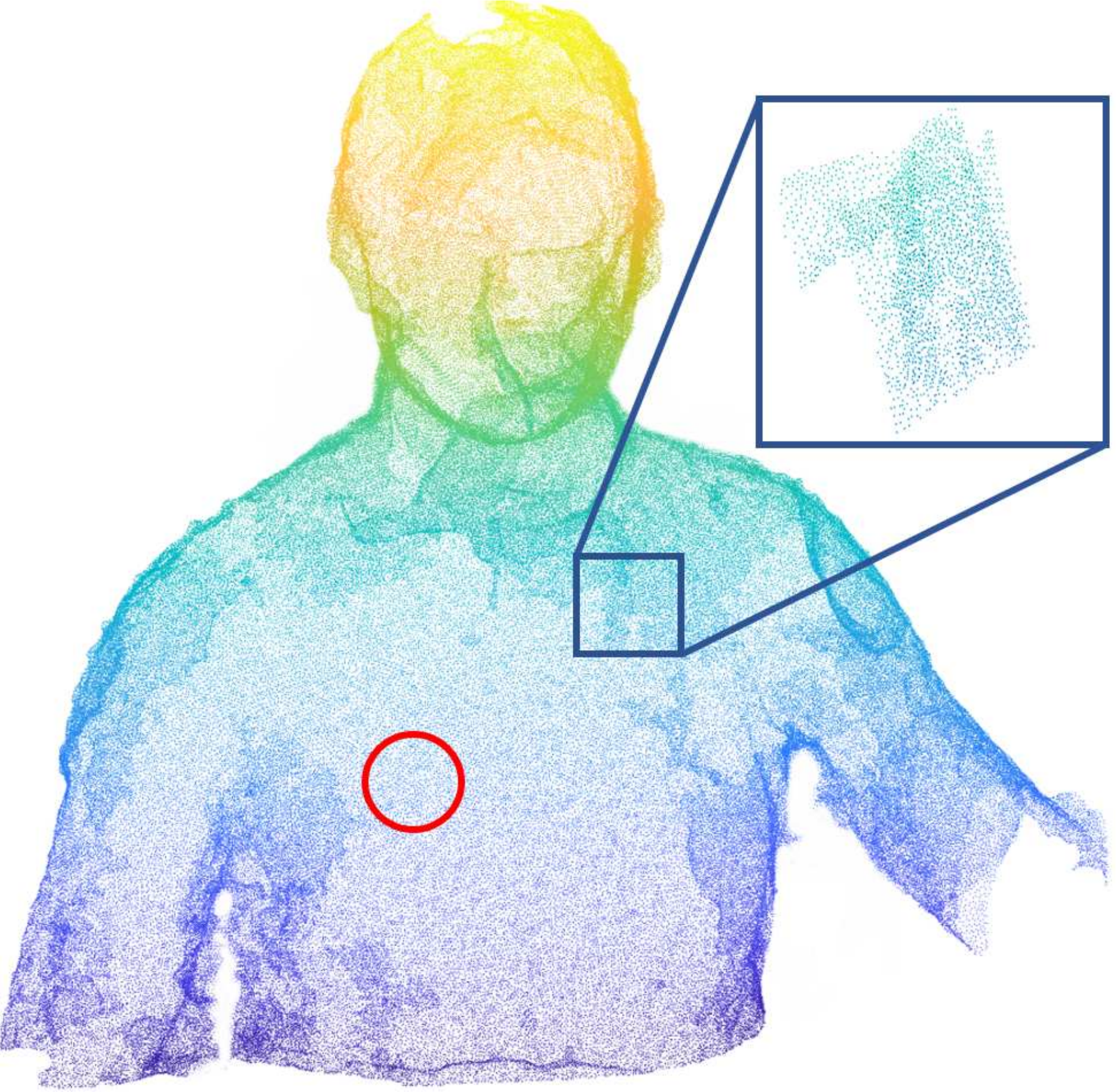} \\
			\vspace{0.01in}
			\includegraphics[width=\textwidth]{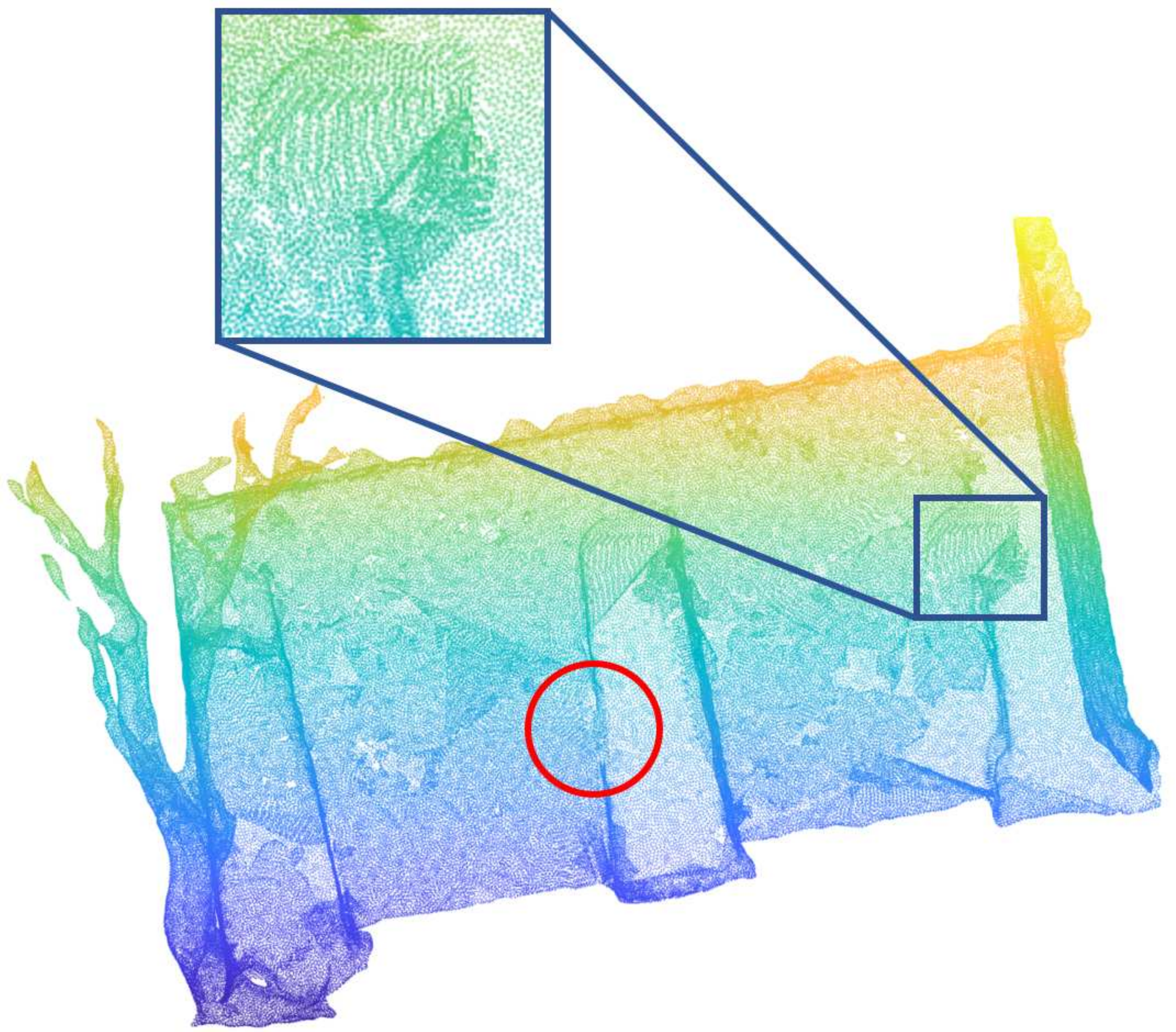}
		\end{minipage}
	}
	\subfigure[\cite{Lozes14}]{
		\begin{minipage}[b]{0.148\textwidth}
			\includegraphics[width=\textwidth]{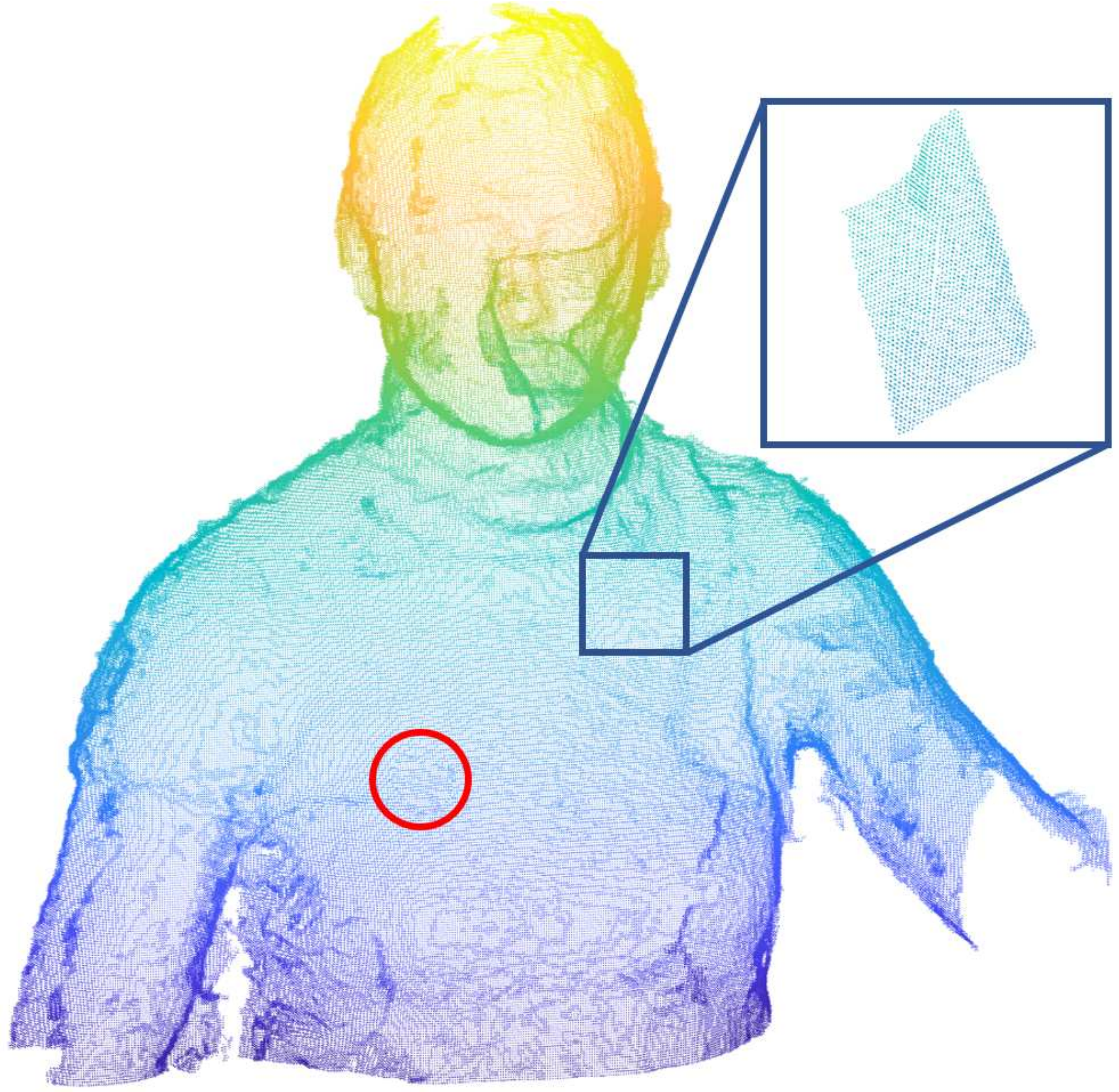} \\
			\vspace{0.01in}
			\includegraphics[width=\textwidth]{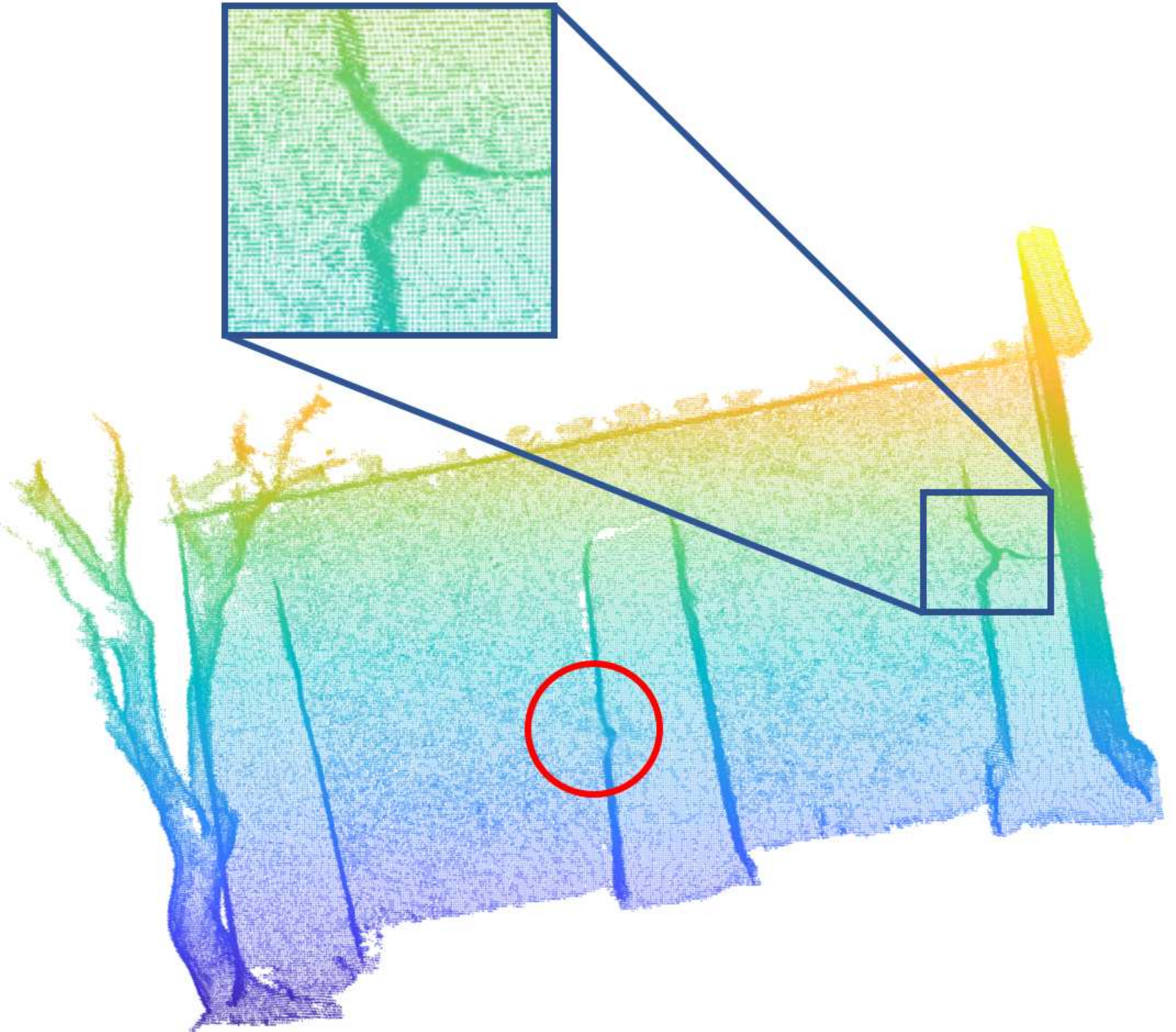}
		\end{minipage}
	}
	\subfigure[Proposed]{
		\begin{minipage}[b]{0.148\textwidth}
			\includegraphics[width=\textwidth]{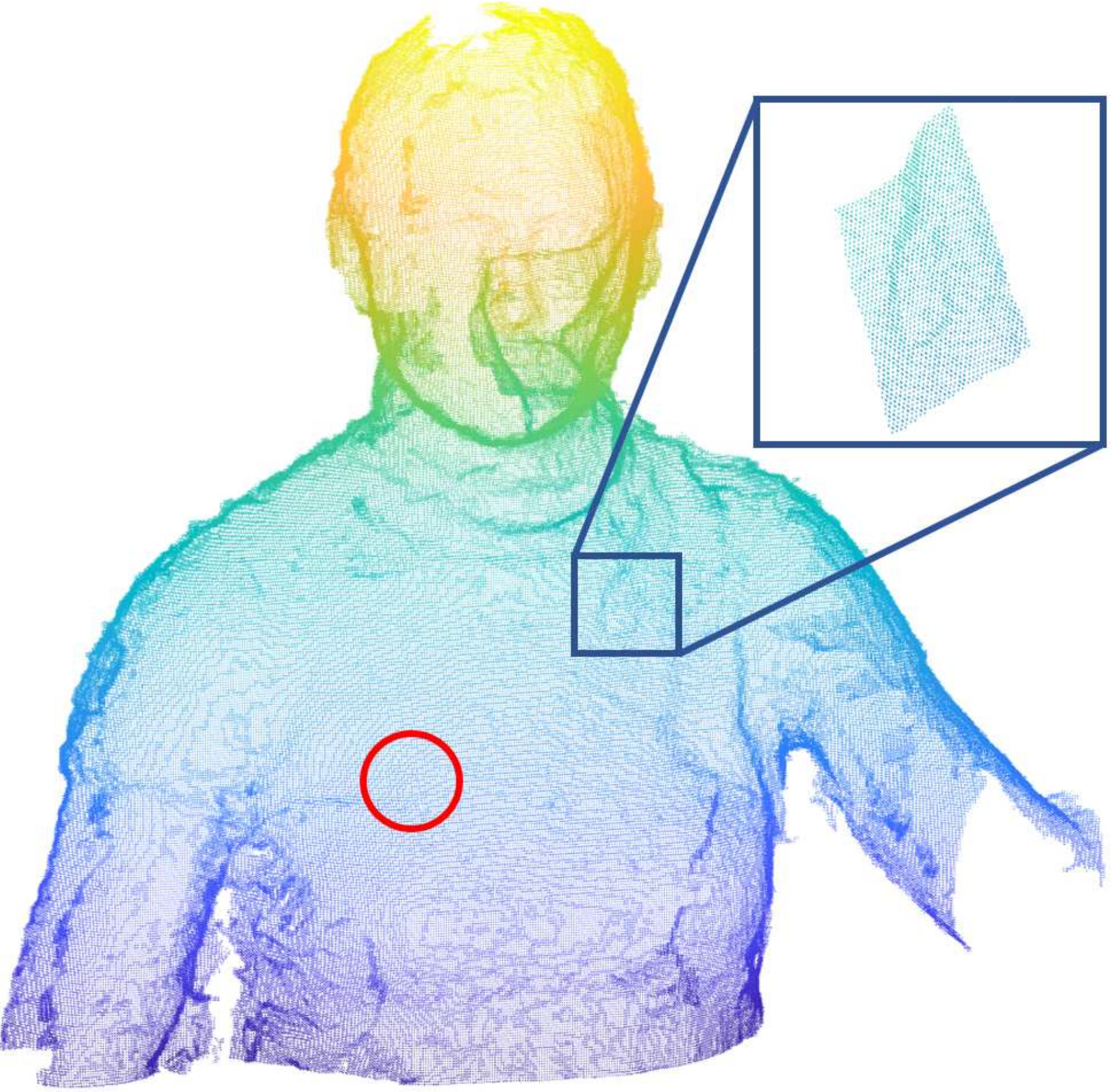} \\
			\vspace{0.01in}
			\includegraphics[width=\textwidth]{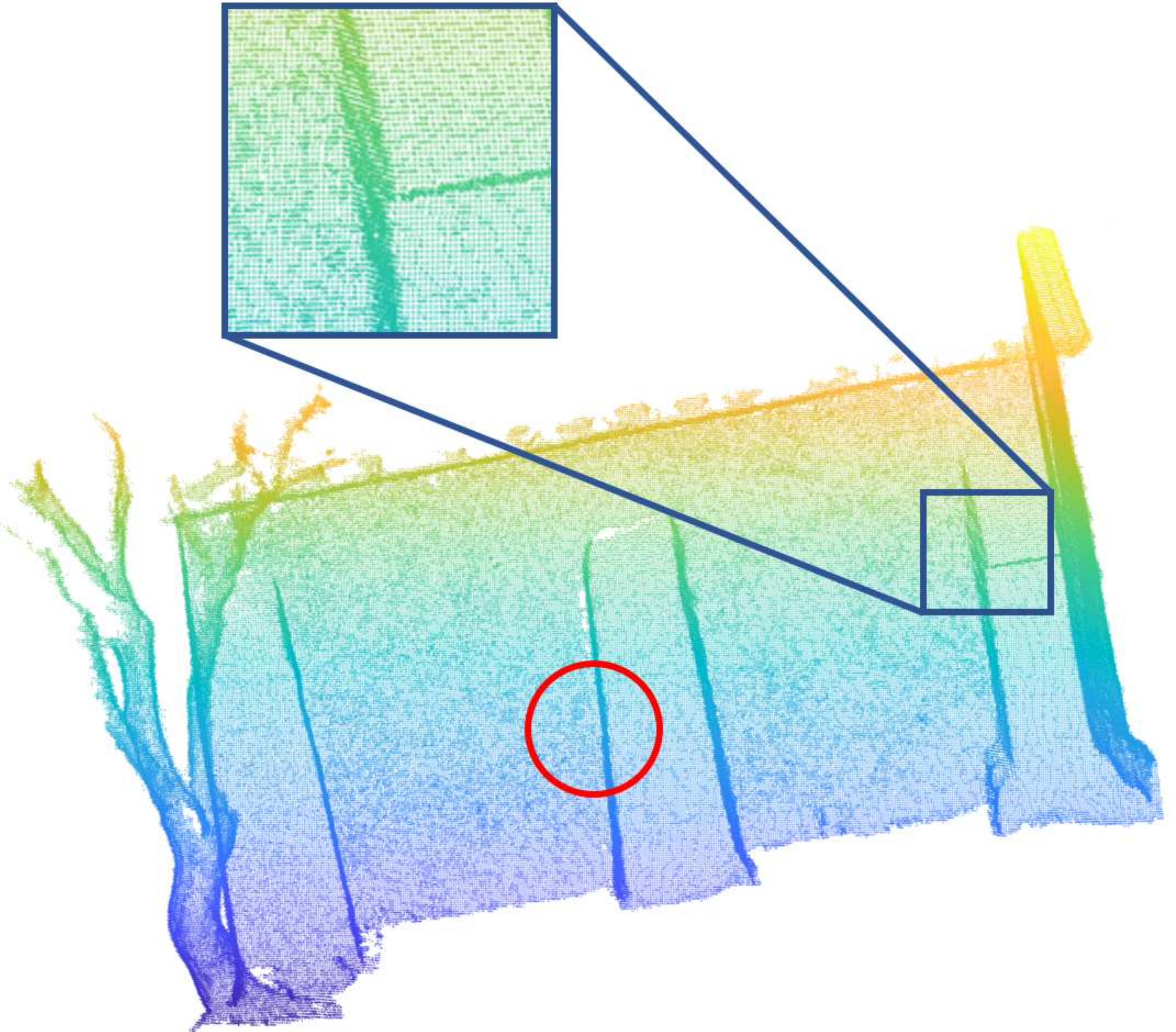}
		\end{minipage}
	}
	\subfigure[Ground Truth]{
		\begin{minipage}[b]{0.148\textwidth}
			\includegraphics[width=\textwidth]{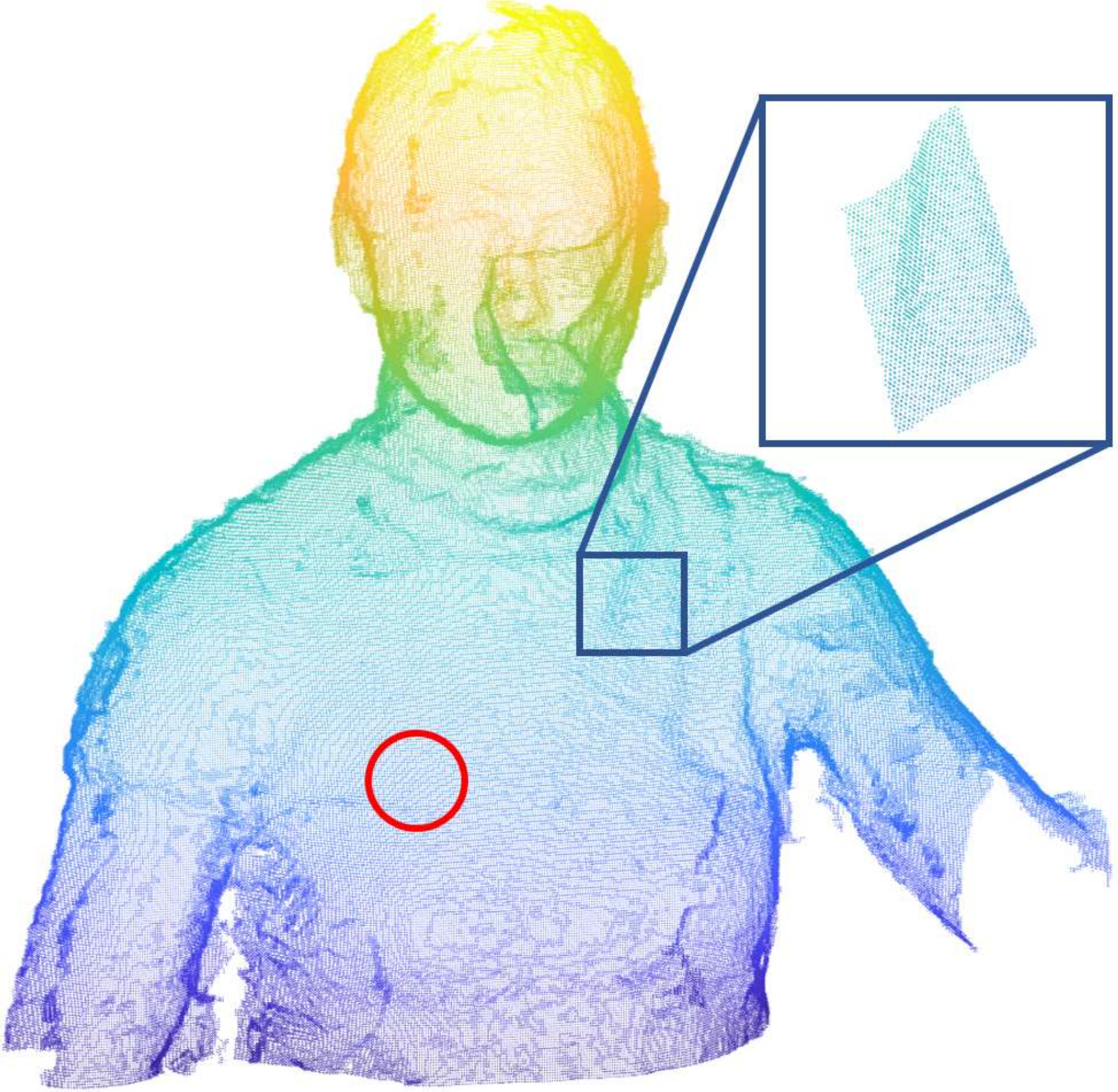} \\
			\vspace{0.01in}
			\includegraphics[width=\textwidth]{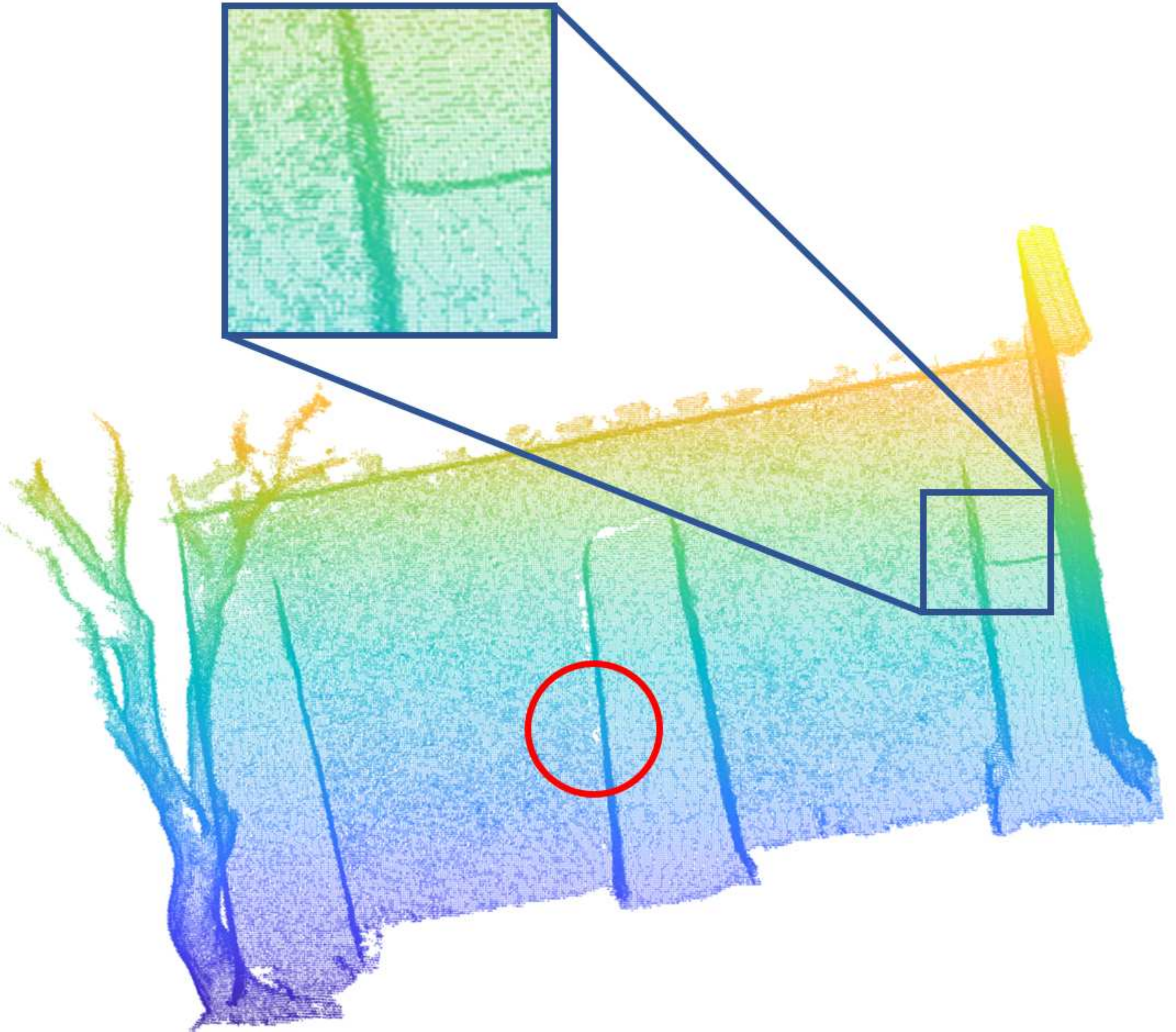}
		\end{minipage}
	}
	\caption{Inpainting from different methods for \textit{Sarah} (row 1), \textit{House} (row 2) with synthetic holes marked in red circles, with one representative cube magnified.}
	\label{fig:result2}
\end{figure*}

GPSNR measures the errors between point cloud $ \mathbf{A} $ and $ \mathbf{B} $ based on optimized PSNR:
\begin{equation}
	GPSNR_{\mathbf{A,B}} = 10 \log_{10} \frac{p^2} {e_{\mathbf{A,B}}},
	\label{eq:gpsnr}
\end{equation}
where $ p $ is the intrinsic resolution of the point clouds and $ e_{\mathbf{A,B}} $ is the point-to-plane distance. Thus the higher GPSNR is, the smaller the gap between $ \mathbf{A} $ and $ \mathbf{B} $ is. Note that GPSNR could be negative, depending on the assignment of the peak value~\cite{Tian17}.

NSHD (Normalized Symmetric Hausdorff Distance) is a normalized metric based on One-sided Hausdorff Distance~\cite{Huttenlocher93}:
\begin{equation}
	NSHD_{\mathbf{A,B}} = \frac{1}{V} \max \{ OHD_{\mathbf{A,B}}, OHD_{\mathbf{B,A}} \},
	\label{eq:nshd}
\end{equation}
where $ V $ is the volume of the smallest axes-aligned parallelepiped enclosing the given 3D point cloud, and $ OHD_{\mathbf{A,B}} $ is the One-sided Hausdorff Distance between point cloud $ \mathbf{A} $ and $ \mathbf{B} $:
\begin{equation}
	OHD_{\mathbf{A,B}} = \max\limits_{\vec{a}_i \in \mathbf{A}} \{ \min\limits_{\vec{b}_j \in \mathbf{B}} \norm{\vec{a}_i-\vec{b}_j}_2 \}.
	\label{eq:ohd}
\end{equation}
Thus the lower NSHD is, the smaller the difference between $ \mathbf{A} $ and $ \mathbf{B} $ is.

Table~\ref{tb:gpsnr} and Table~\ref{tb:nshd} show the objective results for synthetic holes in GPSNR and NSHD respectively. We see that our scheme outperforms all the competing methods in GPSNR and NSHD significantly. Specifically, in Table~\ref{tb:gpsnr} we achieve 49.14 dB gain in GPSNR on average over Meshfix, 31.96 dB over Meshlab, 27.63 dB over~\cite{Wang07}, and 11.58 dB over~\cite{Lozes14}. In Table~\ref{tb:nshd}, we produce much lower NSHD than the other methods, at least four times lower compared to the next best method,~\cite{Lozes14}. Note that, holes are synthesized so that we have the ground truth for the objective comparison.

Further, Fig.~\ref{fig:result1} and Fig.~\ref{fig:result2} demonstrate subjective inpainting results for real holes and synthetic holes respectively. The results of Meshfix and Meshlab exhibit non-uniform density, because they sometimes generate redundant mesh faces to fill wrong holes selected by their automatic hole detection algorithms. For the real holes in Fig.~\ref{fig:result1} (a), which are fragmentary, Meshfix and Meshlab are able to fill small holes with planar geometry structure, but it cannot fully inpaint large missing regions. Besides, \cite{Wang07} and \cite{Lozes14} fill most of the holes by connecting the boundary with simple structure and point density different from the original cloud. Our results shown in Fig.~\ref{fig:result1} (f) demonstrate that our method is able to inpaint both small holes and large holes with appropriate geometry structure and point density, even for holes with complicated geometry. This gives credits to the use of the non-local self-similarity and the graph-signal smoothness prior. 

\renewcommand\arraystretch{1.3}

\begin{table}[h]
	\caption{Performance Comparison in GPSNR (dB)}
	\label{tb:gpsnr}
	\vspace{-0.1in}
	\begin{center}
		\begin{tabular}{|p{1.03cm}<{\centering}|p{1.03cm}<{\centering}|p{1.03cm}<{\centering}|p{1.03cm}<{\centering}|p{1.03cm}<{\centering}|p{1.03cm}<{\centering}|}
			\hline
			{}			&  Meshfix	&  Meshlab	&  \cite{Wang07}	&  \cite{Lozes14}	&  Proposed\\
			\hline
			Andrew		&  3.5201	&  25.5331	&  12.1282	&  46.1208	&  \textbf{56.3579}\\
			\hline
			David		&  -10.7156	&  13.7398	&  27.0006	&  35.8101	&  \textbf{44.1576}\\
			\hline
			House		&  -5.2012	&  17.2414	&  24.5564	&  35.1124	&  \textbf{54.1287}\\
			\hline
			Longdress	&  9.2472	&  -0.8588	&  19.8245	&  37.1343	&  \textbf{50.9951}\\
			\hline
			Phili		&  0.9130	&  25.3133	&  31.5649	&  46.7532	&  \textbf{57.4127}\\
			\hline
			Ricardo		&  -1.4875	&  15.1558	&  20.2132	&  31.4574	&  \textbf{44.1145}\\
			\hline
			Sarah		&  -3.1453	&  21.3919	&  13.7822	&  35.8166	&  \textbf{46.1248}\\
			\hline
			Shiva		&  15.3988	&  28.4587	&  31.5300	&  40.8340	&  \textbf{48.3794}\\
			\hline
		\end{tabular}
	\end{center}
\end{table}

\begin{table}[h]
	\caption{Performance Comparison in NHSD ($ \times10^{-7} $)}
	\label{tb:nshd}
	\vspace{-0.1in}
	\begin{center}
		\begin{tabular}{|p{1.03cm}<{\centering}|p{1.03cm}<{\centering}|p{1.03cm}<{\centering}|p{1.03cm}<{\centering}|p{1.03cm}<{\centering}|p{1.03cm}<{\centering}|}
			\hline
			{}			&  Meshfix	&  Meshlab	&  \cite{Wang07}	&  \cite{Lozes14}	&  Proposed\\
			\hline
			Andrew		&  10.5931	&  16.7684	&  7.6424	&  2.7865	&  \textbf{0.6759}\\
			\hline
			David		&  11.4641	&  7.5213	&  3.5741	&  4.0126	&  \textbf{0.6285}\\
			\hline
			House		&  13.2186	&  19.2414	&  11.1300	&  3.1533	&  \textbf{0.6109}\\
			\hline
			Longdress	&  14.6943	&  18.1654	&  2.7877	&  2.3145	&  \textbf{0.3889}\\
			\hline
			Phili		&  18.4443	&  21.1001	&  13.4530	&  6.1444	&  \textbf{1.4678}\\
			\hline
			Ricardo		&  3.9157	&  4.5453	&  1.1547	&  1.0548	&  \textbf{0.2422}\\
			\hline
			Sarah		&  11.7554	&  6.9805	&  7.8014	&  3.7259	&  \textbf{0.4770}\\
			\hline
			Shiva		&  8.4811	&  13.4852	&  9.6078	&  1.6446	&  \textbf{0.2960}\\
			\hline
		\end{tabular}
	\end{center}
\end{table}

In Fig.~\ref{fig:result2}, we synthesize two holes both in \textit{Sarah} and \textit{House}, with simple and complex geometric structure respectively. We observe that Meshlab cannot fill the holes completely, while \cite{Wang07} introduces wrong geometry around the holes since it attempts to connect the boundary of the hole region using a simple plane. The results of \cite{Lozes14} reveal its drawbacks: it fills the hole by shrinking the hole circularly along the boundary of the hole. Thus the results always depend on the shape of the hole and the geometric trend around the hole. In comparison, our results shown in Fig.~\ref{fig:result2} (e) are inpainted from the nonlocal cubes with the most similar structure, which are almost the same as the ground truth in Fig.~\ref{fig:result2} (f).

\begin{figure}[h]
	\centering
	\subfigure[]{
		\includegraphics[width=0.23\textwidth]{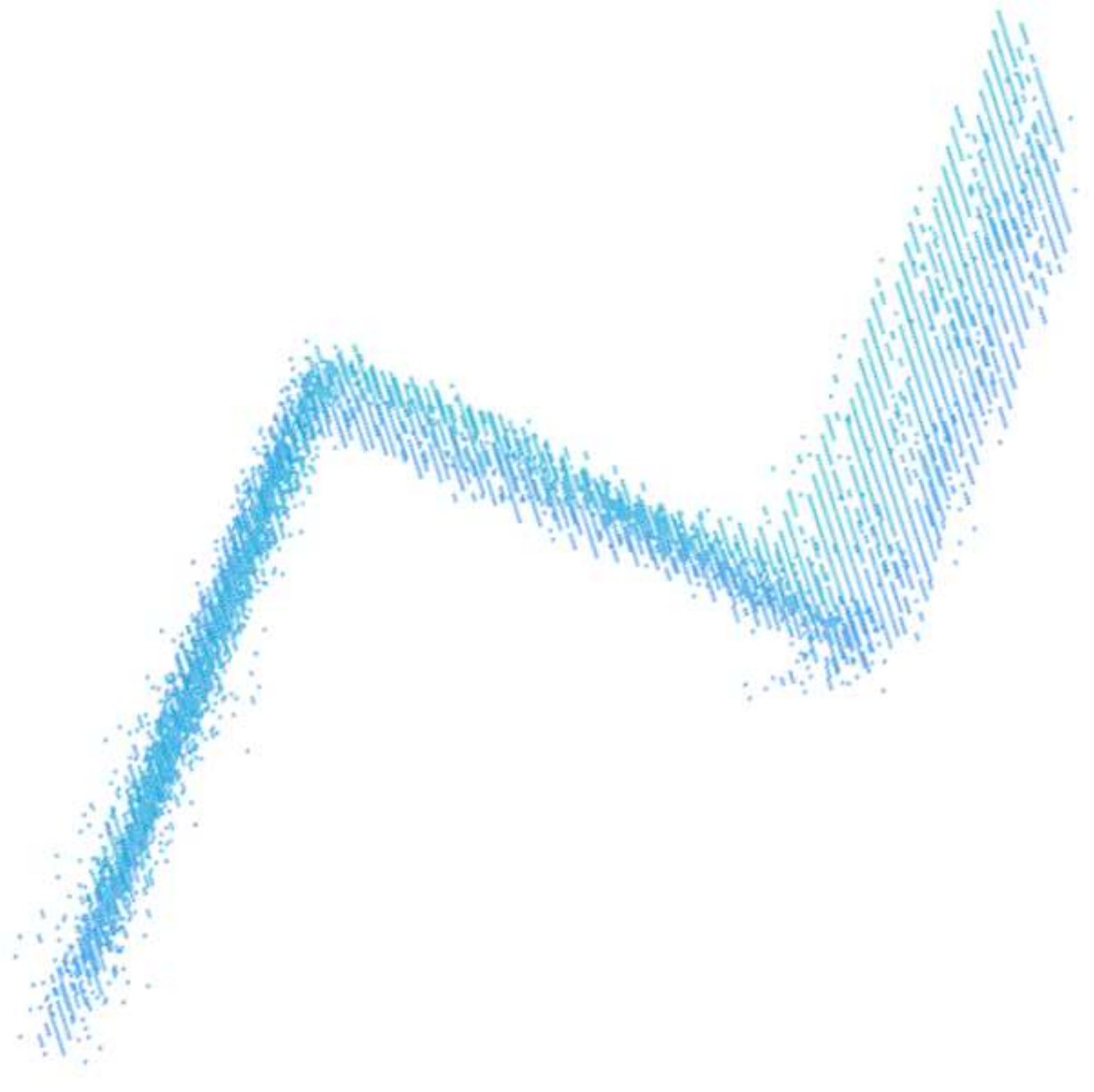}}
	\subfigure[]{
		\includegraphics[width=0.23\textwidth]{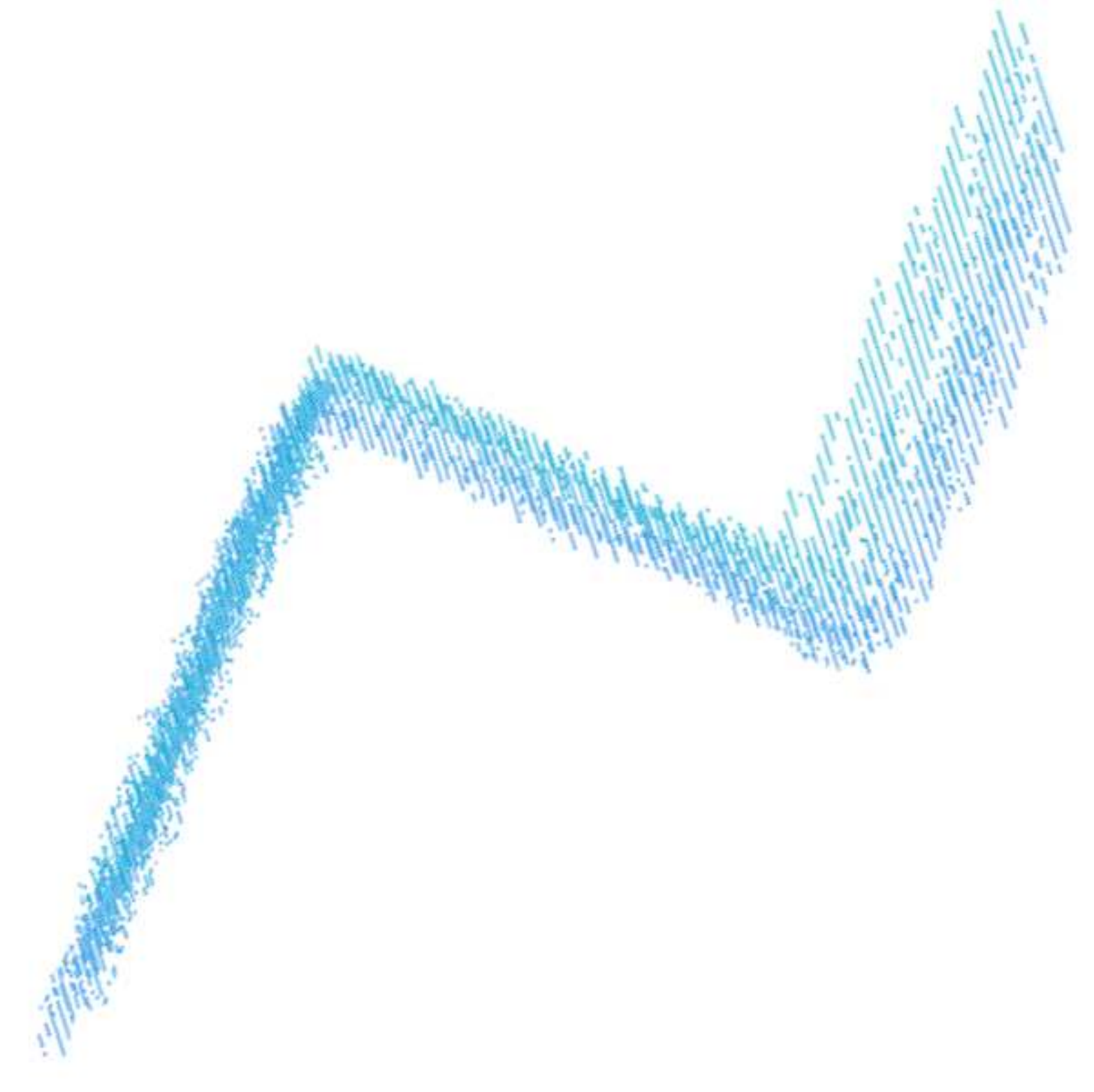}}
	\caption{Denosing effect of another synthetic hole in \textit{House}. (a) The ground truth. (b) The result obtained using the proposed method.}
	\label{fig:denoise}
\end{figure}

Further, we demonstrate the other hole in \textit{House} which is marked in the red circle in Fig.~\ref{fig:result2} from a different view, as in Fig.~\ref{fig:denoise}. We see that there remains some noise in the ground truth due to the acquisition equipment, while our inpainting result attenuates the noise obviously. This additional denoising effect comes from the property of the graph-signal smoothness prior.

%% file: conclude.tex
We exploit the local frequency interpretation and non-local self-similarity on graph for point cloud inpainting. Firstly, by defining frequency on graph as the number of zero crossings in graph nodal domains in particular, we prove the positive correlation between eigenvalues and the frequencies of eigenvectors for the graph Laplacian. Based on this frequency interpretation, we elaborate on the smoothing and denoising functionality of the graph-signal smoothness prior. Then we propose an efficient 3D point cloud inpainting approach leveraging on this prior, where the key observation is that point clouds exhibit non-local self-similarity in geometry. Specifically, we adopt fixed-sized cubes as the processing unit, and search for the most similar cube to the target cube which contains holes. The similarity metric is based on the direct component and the anisotropic graph total variation of normals in the cubes. We then cast the hole-filling problem as a quadratic programming problem, based on the selected most similar cube and regularized by the graph-signal smoothness prior, which also performs denoising along with inpainting. Besides, we propose voxelization and automatic hole detection for point clouds prior to inpainting. Experimental results show that our algorithm outperforms four competing methods significantly. In the future, we consider the extension to the inpainting of the color attribute of point clouds jointly with the geometry.